\documentclass{article}
\PassOptionsToPackage{numbers, compress}{natbib}
\usepackage[preprint]{neurips_data_2024}

\usepackage[utf8]{inputenc} 
\usepackage[T1]{fontenc}    
\usepackage{hyperref}       
\usepackage{url}            
\usepackage{booktabs}       
\usepackage{amsfonts}       
\usepackage{nicefrac}       
\usepackage{xcolor}         
\usepackage{multirow}
\usepackage{natbib}
\usepackage{makecell}
\usepackage{hhline}
\usepackage{graphicx}
\usepackage{amsmath}
\usepackage{bbding}
\usepackage{amssymb}
\usepackage{wrapfig}
\usepackage{tikz}
\usepackage{subcaption}
\newcommand{\myCheckMark}[0]{\textcolor{blue}{\Checkmark}}
\newcommand{\myCrossMark}[0]{\textcolor{red}{\XSolidBrush}}
\usepackage{arydshln}
\usepackage{graphics}
\usepackage{enumitem}

\usepackage{adjustbox}
\definecolor{Gray}{gray}{0.5}
\definecolor{LGray}{gray}{0.9}
\definecolor{darkblue}{RGB}{94,110,186}
\definecolor{darkGreen}{RGB}{92, 148, 110}
\definecolor{myblue}{RGB}{14, 121, 178}
\definecolor{myred}{RGB}{192, 0, 0}

\definecolor{RadarRed}{RGB}{219, 90, 109}
\definecolor{RadarBlue}{RGB}{0 ,79, 152}
\newcommand{\blue}[1]{\textcolor{blue}{#1}}

\definecolor{forestgreen}{rgb}{0.0, 0.50, 0.0}
\definecolor{goldenbrown}{rgb}{0.6, 0.4, 0.08}

\title{VideoVista: A Versatile Benchmark for Video Understanding and Reasoning}

\author{
   \textbf{Yunxin Li} \textsuperscript{1} \quad
   \textbf{Xinyu Chen}\textsuperscript{1}\quad
   \textbf{Baotian Hu}\textsuperscript{1}\quad
   \textbf{Longyue Wang}\quad
   \textbf{Haoyuan Shi}\textsuperscript{1} \quad
   \textbf{Min Zhang}\textsuperscript{1} \quad\\
   \textsuperscript{1} Harbin Institute of Technology, Shenzhen \\
   \texttt{liyunxin987@163.com, vincentwang0229@gmail.com}\\
   \texttt{\{hubaotian, zhangmin2021\}@hit.edu.cn}
}

\begin{document}

\maketitle

\begin{abstract}
Despite significant breakthroughs in video analysis driven by the rapid development of large multimodal models (LMMs), there remains a lack of a versatile evaluation benchmark to comprehensively assess these models' performance in video understanding and reasoning. 
To address this, we present {\em VideoVista}, a video QA benchmark that integrates challenges across diverse content categories, durations, and abilities. Specifically, VideoVista comprises 25,000 questions derived from 3,400 videos spanning 14 categories (e.g., Howto, Film, and Entertainment) with durations ranging from a few seconds to over 10 minutes. Besides, it encompasses 19 types of understanding tasks (e.g., anomaly detection, interaction understanding) and 8 reasoning tasks (e.g., logical reasoning, causal reasoning). To achieve this, we present an {\em automatic data construction framework}, leveraging powerful GPT-4o alongside advanced analysis tools (e.g., video splitting, object segmenting, and tracking). We also utilize this framework to construct {\em training data} to enhance the capabilities of video-related LMMs (Video-LMMs). Through a comprehensive and quantitative evaluation of cutting-edge models, we reveal that: 1) Video-LMMs face difficulties in fine-grained video tasks involving temporal location, object tracking, and anomaly detection; 2) Video-LMMs present inferior logical and relation reasoning abilities; 3) Open-source Video-LMMs' performance is significantly lower than GPT-4o and Gemini-1.5, lagging by 20 points. This highlights the crucial role VideoVista will play in advancing LMMs that can accurately understand videos and perform precise reasoning.

\end{abstract}

\section{Introduction}
With the burgeoning growth of online video platforms and the escalating volume of video content, the demand for proficient video analysis tools has intensified markedly. Recently, large language models (LLMs)~\cite{vicuna2023,chatgpt,touvron2023llama1} have shown remarkable capabilities in language and multimodal tasks, which are often used to build general large multimodal models (LMMs)~\cite{yin2023survey}, particularly promoting the development of open-source visual-language LMMs such as LLaVA~\cite{liu2023llava}, BLIP-2~\cite{li2023blip2}, and Mini-GPT4~\cite{zhu2023minigpt}. Building on these advancements, there is a growing trend toward incorporating LLMs into the field of video information processing~\cite{li2024llms}, and some researchers have introduced promising video-related LMMs (Video-LMMs), e.g., Video-LLaMA~\cite{zhang2023videollama}, VideoChat~\cite{li2023videochat}, and Video-LLaVA~\cite{lin2023video}. These video models aim to enhance user interface and video content through intelligent question-and-answer (QA) sessions, like ChatGPT~\cite{chatgpt} and Gemini~\cite{team2023gemini}. 
By leveraging the advanced language understanding capabilities of LLMs, these models can provide detailed explanations, context-aware responses, and interactive engagement with video content. 

A natural and significant research question is \textit{how to comprehensively assess these Video-LLMs' understanding and reasoning capability}, which will guide their future enhancements. 
However, current Video-LMMs are primarily tested on specific video QA tasks such as ActivityNet~\cite{yu2019activitynet}, WildQA~\cite{castro-etal-2022-in-the-wild}, and MSRVTT-QA~\cite{xu2016msr}. These benchmark datasets mainly consist of {\em short video clips} in {\em limited scenes} such as TV shows, cooking demonstrations, and simulated motion traces, focusing on understanding but {\em paying less attention to reasoning}. Although MVBench \cite{li2023mvbench} collects existing task-specific videos to build a diverse benchmark, it focuses on short video understanding with temporal reasoning and restricted video sources. These limitations hinder assessing Video-LMMs' performance on challenging long videos, diverse video categories, and various video reasoning tasks.

\begin{figure}[t]
 \begin{minipage}{0.35\textwidth}
 \small
 \renewcommand\tabcolsep{1.7pt} 
 \renewcommand\arraystretch{1.10} 
 \begin{tabular}{l r}

        \toprule
        \textbf{Category} & \textbf{Size} \\
        \midrule
        Task Classes & 27 \\
        ~- Understanding & 19 \\
        ~- Reasoning & 8 \\
        \hline
        Video Sources & 894 \\
        ~- Categories & 14 \\
        ~- Video Clips & 3,402 \\
        ~- Maximum Duration &  919s\\
        ~- Average Duration &  131s\\
        \midrule
        Maximum Question Length &  62\\ 
        Maximum Option Length & 142\\
        Maximum Choice Number & 5\\
        \hdashline\noalign{\vskip 0.5ex}
        Average Question Length &  13.55\\
        Average Option Length &  8.55\\
        Average Choice Number & 4\\
        \hdashline\noalign{\vskip 0.5ex}
        Total Samples & 24,906 \\
        Total Questions & 24,906 \\
        \bottomrule
    \end{tabular}
\end{minipage} 
\hfill
\begin{minipage}{0.60\textwidth}
 \small
 \vspace{-1mm}
\includegraphics[width=0.98\linewidth]{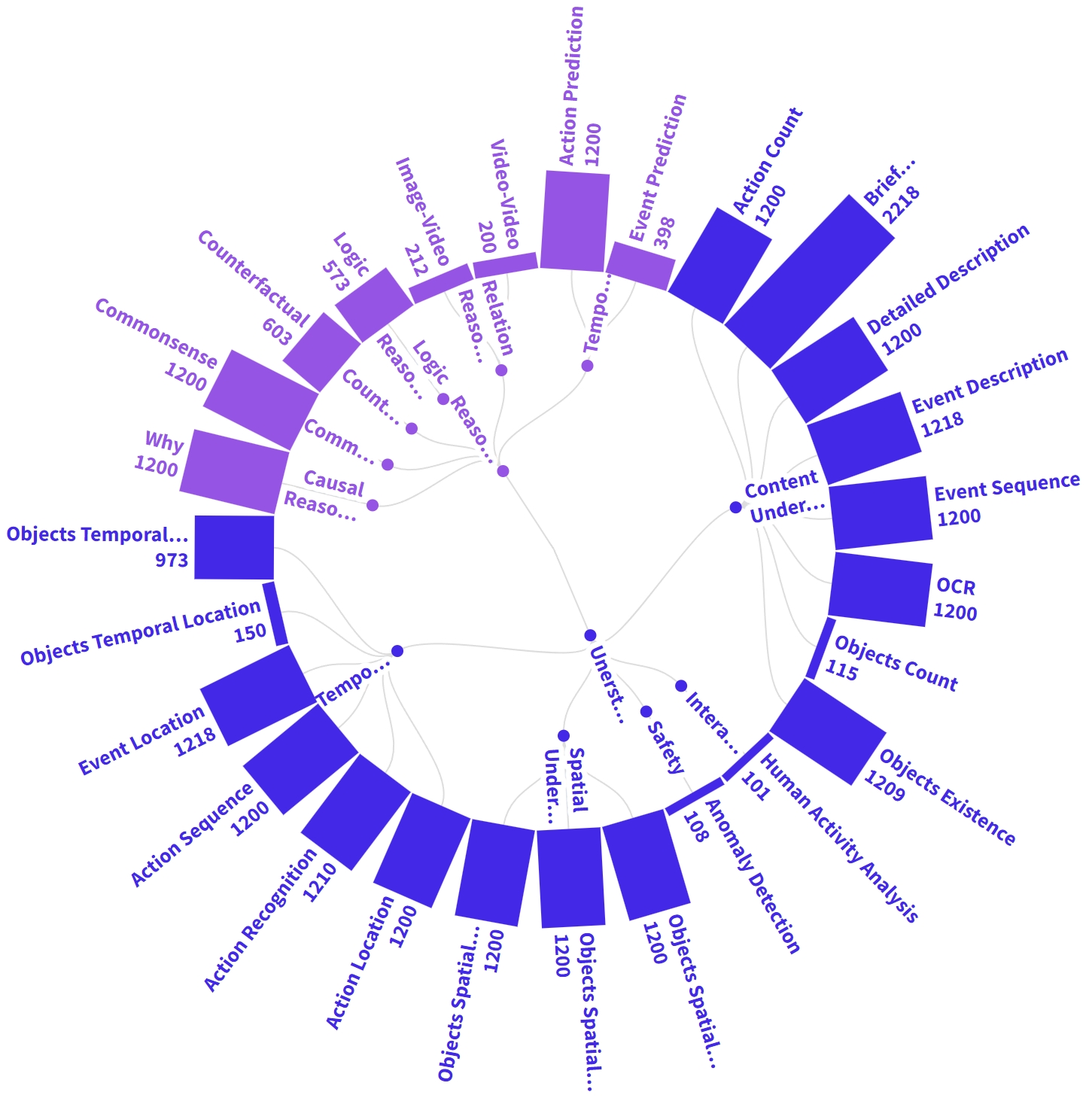}
\vspace{-1mm}
 \end{minipage}
\caption{Comprehensive statistics from different perspectives (left) and detailed sample counts for each {\em Task Class} (right) in the VideoVista dataset.}
\label{fig:comprehensive_stats}
\end{figure}


To address this issue, we present a comprehensive video QA benchmark dataset named {\em VideoVista}, encompassing challenges across diverse content categories, durations, and abilities (as shown in Figure \ref{fig:comprehensive_stats}). We propose a task taxonomy to guide the development of VideoVista: 1) We collect {\em 14 categories} of videos to check the general understanding capability of models, which covers Science and Technology, Sports, Entertainment etc.
2) We construct videos of {\em varying lengths}, ranging from a few seconds to over 10 minutes, to assess the accuracy of models in processing both short- and long-term videos; 3) We introduce {\em 19 types of understanding tasks} such as Event Location and Anomaly Detection etc.; and {\em 8 types of Reasoning tasks} such as Relation Reasoning (Image/Video-Video Relation) etc.
To achieve this, we download 894 complete videos from YouTube\footnote{\url{https://www.youtube.com}.} and split them into clips of diverse duration using a specialized video splitting technique. While formulating questions to test various abilities, we present an {automatic data construction framework} that leverages the powerful GPT-4 model alongside advanced video analysis methods including video splitting, object segmentation, and tracking. For challenging understanding and reasoning questions such as Object Count, Anomaly Detection, and Logical Reasoning, we manually verify the quality of the question-answer pairs and filter out incorrect examples to ensure the overall quality of the dataset. Totally, VideoVista comprises 3,402 videos, with about 25,000 questions to evaluate the total of 11 ability aspects (27 tasks) of Video-LMMs.

To better understand challenges posed by VideoVista, We conduct extensive evaluations and analyses on {\em 10 cutting-edge Video-LMMs}. Experimental results reveal that:
1) Video-LMMs face difficulties in handling long videos and some fine-grained video understanding tasks, e.g., temporal location and anomaly detection; 2) Video-LMMs present inferior logical and relation reasoning abilities, especially for Video-Video relations inference; 3) Open-source Video-LMMs' performance is significantly lower than GPT-4o and Gemini-1.5. There are three {\bf main contributions} in this work:
\begin{itemize}[leftmargin=*,topsep=0.1em,itemsep=0.1em,parsep=0.1em]
    \item We build and release a versatile video QA benchmark comprising 14 categories, 3$\sim$900s durations, and 27 types of tasks for thoroughly assessing the capabilities of Video-LMMs.\footnote{Dataset is available at \url{https://github.com/HITsz-TMG/UMOE-Scaling-Unified-Multimodal-LLMs/tree/master/VideoVista}.}
    \item We develop an automatic video annotation framework that facilitates the efficient creation of large-scale training and evaluation VideoQA datasets, as described in Sections \ref{videos_syn} and \ref{auto-qa-gen}.
    \item Our extensive analyses identify three principal shortcomings (understanding, reasoning, and comprehensive abilities) of the current Video-LMMs, highlighting areas for future enhancement.
\end{itemize}

\section{VideoVista Benchmark}
\label{dataset_construct}

\begin{figure}[t]
    \centering
    \includegraphics[width=0.95\textwidth]{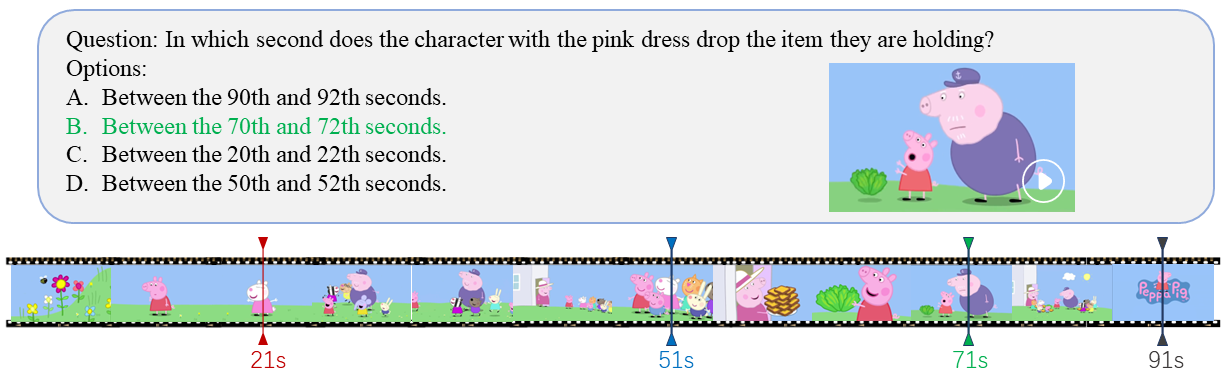}
    \caption{An example of {\em Action Location} task from the {\em Film \& Animation} category, demonstrating {\em Temporal Understanding Ability} in the VideoVista dataset. The ground-truth answer is highlighted in green. This video clip starts approximately at 57:55 of the original video (\url{https://www.youtu.be/dZr7oAB\_fc0}). Examples of all tasks are detailed in the Appendix~\S\ref{task_stat}.}
    \label{fig:example_case}
\end{figure}

\begin{figure}[t]
    \centering
    \includegraphics[width=0.95\textwidth]{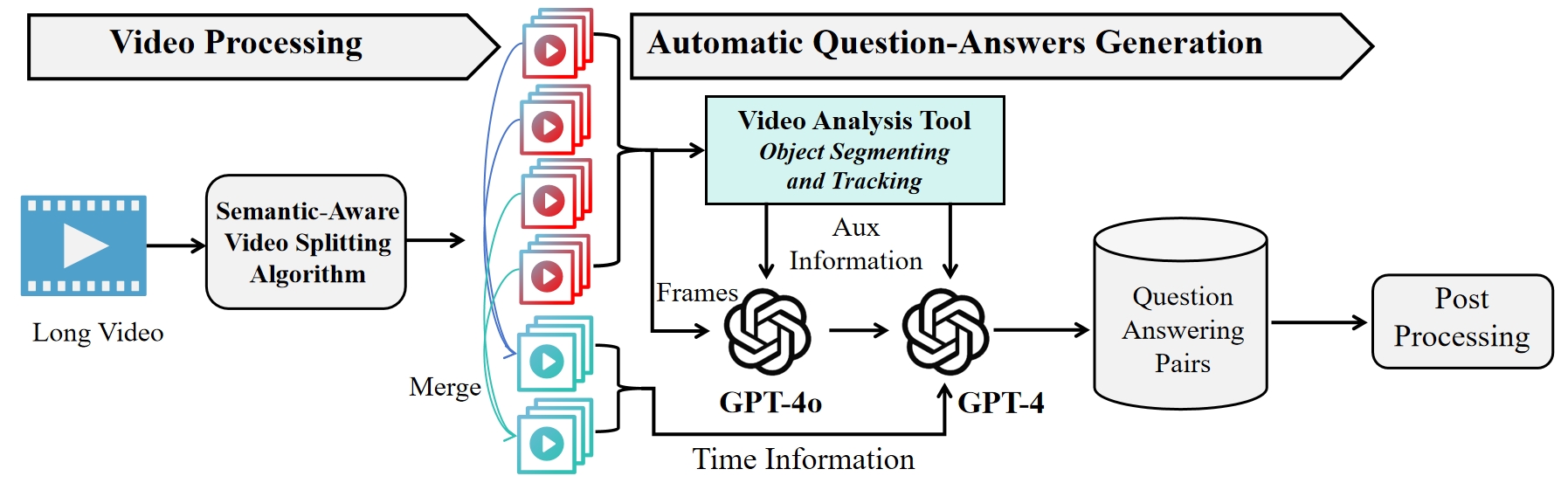}
    \caption{The overview of the construction process of VideoVista.}
    \label{fig:model_process}
\end{figure}

Figure~\ref{fig:example_case} shows an example from our VideoVista dataset. The construction process is shown in Figure~\ref{fig:model_process}. We split long videos into clips, merge adjacent clips into videos of varying lengths, annotate each clip using GPT-4o, and use the GPT-4 model to convert these annotations into question-answer pairs for the merged videos.
To construct a versatile benchmark, we follow these guidelines: 1) include diverse video categories; 2) ensure varying video durations, from a few seconds to over 10 minutes; 3) incorporate comprehensive video understanding and reasoning tasks.
Section \ref{videos_syn} provides a comprehensive walkthrough of the video splitting and merging procedures, while 
Section \ref{auto-qa-gen} delves into the meticulous process of annotating video clips and converting these annotations into the final question-answer pairs.
Section \ref{data-over} shows the statistics and companions of VideoVista.

\subsection{Video Processing}
\label{videos_syn}
Firstly, we randomly select 894 videos covering 14 categories from the test set of the Panda-70M dataset~\cite{chen2024panda70m}. Then we followed the Semantics-aware Video Splitting Algorithm presented in the Panda-70M dataset to split a whole video into several short clips with consistent semantics. However, some of these split video clips are still long (several minutes) with fixed lenses, which makes it difficult to annotate with video analysis tools. 

To address this issue, we use CLIP~\cite{xue2022clip} to split lengthy video clips. Specifically, we extract one frame per second and use CLIP to obtain frame features. Starting with the 0th frame, we calculate cosine similarity to identify the most deviating frame among the next 39 frames as the $i$th frame. This can create a sub-clip of less than 40 seconds from the 0th to the $(i-1)$th frame. We will continue this process from the $i$th frame until all sub-clips are shorter than 40 seconds.

\begin{figure}[t]
    \centering
    \begin{subfigure}[b]{0.49\textwidth}
        \centering
        \includegraphics[width=\textwidth]{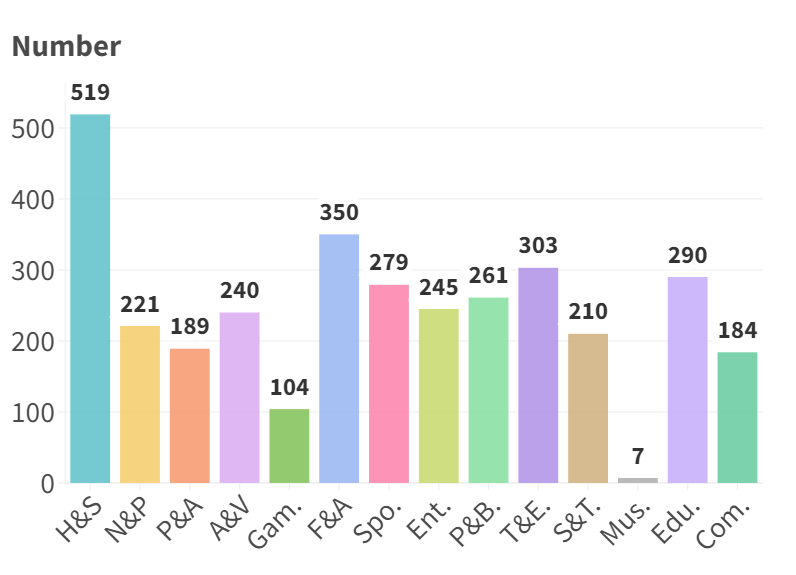}
        \caption{The statistics of 14 video categories}
        \label{fig:sub11}
    \end{subfigure}
    \hfill
    \begin{subfigure}[b]{0.49\textwidth}
        \centering
        \includegraphics[width=\textwidth]{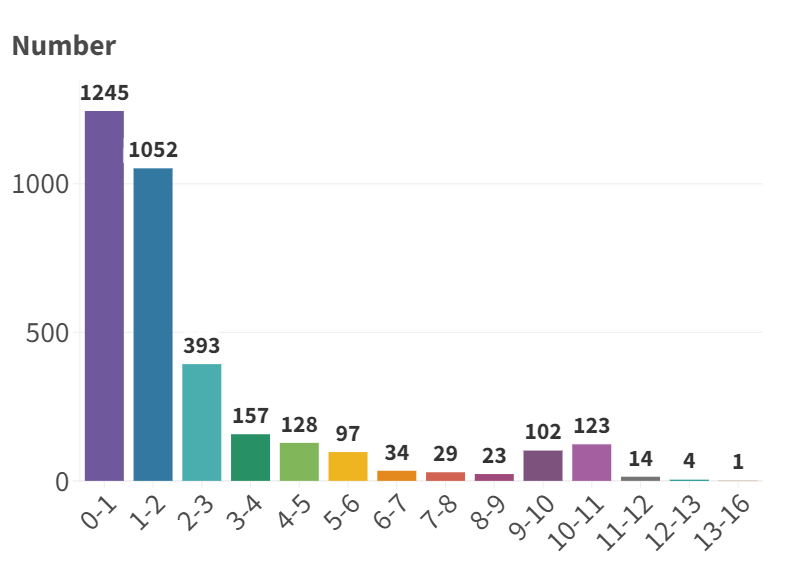}
        \caption{The distribution of video durations (minute)}
        \label{fig:sub22}
    \end{subfigure}
    \caption{An illustration of video categories and durations in VideoVista. (a) shows the number of videos each category contains and (b) presents the duration distribution of all videos. The category in (a) use abbreviations: Howto \& Style (\textbf{H\&S}), News \& Politics (\textbf{N\&P}), Pets \& Animals (\textbf{P\&A}), Autos \& Vehicles (\textbf{A\&E}), Gaming (\textbf{Gam.}), Film \& Animation (\textbf{F\&A}), Sports (\textbf{Spo.}), Entertainment (\textbf{ENT.}), People \& Blogs (\textbf{P\&B}), Travel \& Events (\textbf{T\&E}), Comedy (\textbf{Com.}), Science \& Technology (\textbf{S\&T}), Education (\textbf{Edu.}), Music (\textbf{Mus.})
    }
    \label{fig:durations_and_categories}
\end{figure}

Finally, we randomly merge these foundational video clips to form videos with varying lengths. We mainly merge short clips into long videos with four time intervals: less than 1 minute, 1-2 minutes, 2-5 minutes, 5-10 minutes, and over 10 minutes. During the merging process, we ensure that each video clip is merged only once within the same time interval to retain the diversity of video content. By doing so, we gain 3402 videos ranging from a few seconds to over 10 minutes. The detailed categories and durations of all videos are shown in Figure~\ref{fig:durations_and_categories}.

\subsection{Automatic QA Generation}
\label{auto-qa-gen}
%
With the guidance of task construction shown in Figure~\ref{fig:comprehensive_stats}, we annotate video clips and generate the video QA pairs from the following perspectives. 

\textbf{Action:} Firstly, we employ GPT-4o to annotate actions for each short clip, which receives video frames, titles, categories, and audio transcripts (extracted by Whisper~\cite{radford2023robust}). While inputting video frames, we create composite images from four frames each, with a maximum of ten images per video clip (covering clips up to 40 seconds). The output format of GPT-4o is triplets of \textit{(Time, Subject, Action)} and it can output multiple triplets possible per clip to capture different actions by different subjects at various times.

\textbf{Event:} Secondly, we use additional objects with bounding boxes and previously annotated action triplets alongside the above inputs (Action) to annotate video events. Objects with bounding boxes identified and detected by Segment Anything~\cite{kirillov2023segment} and Recognize Anything~\cite{zhang2023recognize}, provide appearing times and regions of objects in a video. We require GPT-4o to produce detailed event descriptions.

\textbf{Objects:} Thirdly, we leverage GPT-4 to directly generate question-and-answer pairs about objects in a video clip. The input for this stage comprises video frames, titles, categories, and objects with bounding boxes. Using this information, we instruct the model to select and generate three suitable types of QA pairs based on the inputs, from a predefined set of eight object-relevant tasks.

\textbf{Reasoning:} We instruct the GPT-4 model to generate reasoning QA pairs based on the input video clip from predefined reasoning task types, descriptions, and reference examples. The input information for this stage includes frames, video titles, video categories, and audio transcripts.

\textbf{Long Video Annotation:}
After annotating the foundational video clips, we create QA pairs for the merged videos. We first adjust time and transform annotations of short clips into merged video annotations. Based on these, GPT-4 generates Action and Event relevant QA pairs, specifically incorporating annotations from subsequent clips for prediction tasks. 
For the Objects and Reasoning tasks, we merge the QA pairs of short video clips to form new QA pairs of merged video clips. In particular, we provide GPT-4 with instructions to adjust time, merge similar questions, and add time intervals in answers. However, this process sometimes results in incorrect or biased answers because of GPT-4's instability, so we perform human filtering for task types with high error rates. Additionally, we also use a template-based construction method for Relation Reasoning and Anomaly Detection. For video-video relation task, the candidates contain four predefined relationships (before, after, in, and none) between two videos. When merging videos, we can determine the relationship between any two videos and effectively construct relation-inferring questions through templates with fixed questions and candidates. For anomaly detection, the candidates include four risk types (hate and fairness, sexual, violence, and self-harm) from the content safety of Azure OpenAI Service.

\begin{table}[t]
\centering
\caption{
The comparison of various benchmarks involves several key aspects: total number of videos (\textbf{\#Videos}), number of clips (\textbf{\#Clips}), average video duration (\textbf{Len.}), number of QA pairs (\textbf{\#QA Pairs}), annotation method (\textbf{Anno.}, where M/A indicates manual/automatic), whether the videos span multiple duration levels (\textbf{Multi.}), if the videos are sourced from diverse open domains (\textbf{Open.}), and whether subtitles are provided alongside audio information (\textbf{S.A.}). 
} 
\vspace{3pt}
\scalebox{0.8}{
\begin{tabular}{l rrrr c ccc}
\toprule
\textbf{Benchmarks} & \textbf{\#Videos} & \textbf{\#Clips} &\textbf{Len.(s)} & \textbf{\#QA Pairs} &  \textbf{Anno.} & \textbf{Multi.}  & \textbf{Open.} & \textbf{S.A.} \\
\midrule
MSRVTT-QA & 2,990 & 2,990 & 15.2 & 72,821 & A &  \myCrossMark & \myCheckMark & \myCrossMark\\
MSVD-QA & 504 & 504 & 9.8 & 13,157 & A &  \myCrossMark & \myCheckMark & \myCrossMark \\
TGIF-QA  \citep{jang2017tgif} & 9,575 & 9,575 & 3.0 & 8,506 & A\&M &  \myCrossMark & \myCheckMark & \myCrossMark \\
ActivityNet-QA \citep{yu2019activitynet} & 800 & 800 & 111.4 & 8,000 & M & \myCrossMark & \myCrossMark & \myCrossMark\\
TVQA \citep{lei2018tvqa} & 2,179 & 15,253 & 11.2 & 15,253 & M & \myCrossMark & \myCrossMark & \myCrossMark\\
How2QA & 1,166 & 2,852 & 15.3 & 2,852 & M & \myCrossMark& \myCheckMark & \myCrossMark \\
STAR & 914 & 7,098 & 11.9 & 7,098 & A & \myCrossMark & \myCheckMark & \myCrossMark\\
NExT-QA \citep{xiao2021next} & 1,000 & 1,000 & 39.5 & 8,564 & A & \myCrossMark & \myCheckMark & \myCrossMark \\
\midrule
MVBench \citep{li2023mvbench} & 3,641 & 3,641 & 16.0 & 4,000 & A & \myCrossMark & \myCheckMark & \myCrossMark\\
Video-Bench \citep{ning2023video} & 5,917 & 5,917 & 56.0 & 17,036 & A\&M & \myCrossMark & \myCheckMark & \myCrossMark\\
EgoSchema \citep{mangalam2024egoschema} & 5,063 & 5,063 & 180.0 & 5,063 & A\&M & \myCrossMark & \myCrossMark & \myCrossMark \\
AutoEval-Video \citep{chen2023autoeval} & 327 & 327 & 14.6 & 327 & M & \myCrossMark & \myCheckMark & \myCrossMark\\
TempCompass \citep{Liu2024TempCompassDV} & 410 & 500 & 11.4 & 7,540 & A\&M & \myCrossMark & \myCheckMark & \myCrossMark\\
Video-MME \citep{fu2024videomme} & 900 & 900 & 1024.0 & 2,700 & M &\myCheckMark & \myCheckMark & \myCheckMark\\
\midrule
\textbf{VideoVista} & 894 & 3,402 & 131.0 & 24,906 & A & \myCheckMark & \myCheckMark & \myCheckMark\\
\bottomrule
\end{tabular}}
\label{tab:comparison}
\vspace{-0.5cm}
\end{table}

\textbf{Human Filtering:} 
Upon reviewing QAs generated by GPT-4, we discovered obvious shortcomings in Object Count questions. Specifically, GPT-4o tended to focus solely on individuals in the centre of the frame, disregarding those in the background. Consequently, we implemented human filtering for the Object Count task. Additionally, errors were detected in the Objects Temporal Relation task (e.g., changes in the first appearance time of individuals for merged videos) and Human Activity Analysis task. Therefore, we also conducted human filtering for these two tasks.

\textbf{Option Generation:} 
To ensure fairness and accuracy during evaluation, we convert all open-ended QAs generated by GPT-4 into multiple-choice QA pairs. Given open-ended QAs and auxiliary information such as video titles and audio transcripts, we instruct GPT-4 to generate one correct answer option and three incorrect distractors for each input question. Ensuring the options have similar lengths helps mitigate the common issue where 'longer options tend to be correct'. \textit{All detailed instruction prompts are shown in the Appendix~\S\ref{guidlines_const}.}

\subsection{Dataset Statistics}
\label{data-over}
As shown in Figure \ref{fig:comprehensive_stats}, our VideoVista dataset encompasses 3,402 videos ranging from a few seconds to over 10 minutes in duration across 14 categories, and includes 24,906 questions across 27 task types.
We compare our benchmark's key characteristics with others in Table \ref{tab:comparison}. As seen, existing benchmarks focused on specific video categories or lacked videos of varying durations. Video-MME addressed these but had fewer task types and questions due to manual annotation. 







\begin{table}[t]
\centering
\caption{\textbf{Evaluation results on VideoVista dataset}. The language model used by Video-LLMs (\textbf{Language Model}), frames input (\textbf{Frames}), evaluation scores in the Video Understanding Task (\textbf{Unders.}), evaluation scores in the Video Reasoning Task (\textbf{Reason.}), overall evaluation scores (\textbf{Overall}). While 1 fps means the model samples one frame per second from the video.} 


\label{tab:VideoVista_Result} 
\vspace{3pt}
    \scalebox{0.83}{
    \begin{tabular}{l cc rrr}
        \toprule
        \textbf{Model}  & \bf Language Model & \bf Frames & \bf Unders. & \bf Reason. & \bf Overall \\
        \midrule
        {VideoChat+ChatGPT} \citep{li2023videochat} & gpt-3.5-turbo & 40 & 16.64 & 23.04 & 17.99\\
        {Video-LLaMA} \citep{zhang2023videollama} & Vicuna-7B & 16 &25.40 & 25.16 & 25.35 \\
        {Video-ChatGPT} \citep{Maaz2023VideoChatGPT} & Vicuna-7B & 100 & 36.09 & 38.73 & 36.65\\
        {IVA} \citep{li2024llms}  & Vicuna-7B & 200 & 37.38 & 48.38 & 39.70 \\
        {Video-LLaVA} \citep{lin2023video} & Vicuna-7B & 8 & 53.82 & 66.91 & 56.59 \\ 
        {LLaMA-VID} \citep{li2023llamavid} & Vicuna-7B & 1 fps & 54.00 & \textbf{67.61} & 56.87\\
        {LLaVA-NeXT-Video} \citep{liu2024llavanext} & Vicuna-7B & 16 & 54.12 & 66.14 & 56.66 \\
        {VideoChat2-Vicuna} \citep{li2023mvbench} & Vicuna-7B & 16 & 51.79 & 60.55 & 53.64\\
        {VideoChat2-Mistral} \citep{li2023mvbench} & Mistral-7B & 16 & \textbf{54.91} & 65.95 & \textbf{57.24}\\
        \midrule
        {Gemini-1.5-Flash} & Gemini & 1 fps & 74.73 & 82.30 & 76.39\\
        {GPT-4o} & GPT-4o & 100 & \textbf{75.15} & \textbf{87.97} & \textbf{78.26} \\
        \bottomrule
    \end{tabular}}
\end{table}

\section{Experiments}


In this section, we evaluate latest open-source Video-LLMs, including Video-LLaMA~\cite{zhang2023videollama}, Video-ChatGPT~\cite{Maaz2023VideoChatGPT}, IVA~\cite{li2024llms}, Video-LLaVA~\cite{lin2023video}, LLaMA-VID~\cite{li2023llamavid}, LLaVA-NeXT-Video~\cite{liu2024llavanext}, VideoChat2~\cite{li2023mvbench}, as well as commercial models GPT-4o~\citep{gpt4} and Gemini-1.5-Flash~\citep{team2023gemini}. \textit{Detailed introductions of these models are shown in the Appendix~\S\ref{model_descripton}.}

\subsection{Main Results}

\paragraph{Overall Performance.}
The overall evaluation results are shown in Table \ref{tab:VideoVista_Result}. We further present the evaluation results of the Video Understanding Task in Table~\ref{tab:task_understanding} and the Video Reasoning Task in Table~\ref{tab:task_reasoning}.
Inspired by MVBench~\cite{li2023mvbench}, we utilize answer prompts like "Best Option:" in task instructions to guide open-source Video-LLMs in generating answers in the desired formats. 
Table~\ref{tab:VideoVista_Result} demonstrates that commercial LMMs GPT-4o and Gemini significantly outperform the open-source Video-LMMs, exceeding their performance by approximately 20\%.

\begin{table}[t]
\centering
\caption{\textbf{Results on Video Understanding Tasks.} Objects Existence (\textbf{OE}), Objects Count (\textbf{OC}), Action Count (\textbf{AC}), Detailed Description (\textbf{DD}), Brief Description (\textbf{BD}), Event Description (\textbf{ED}), Event Sequence (\textbf{ES}), Optical Character Recognition (\textbf{OCR}), Action Recognition (\textbf{AR}), Action Sequence (\textbf{AS}), Action Location (\textbf{AL}), Event Location (\textbf{EL}), Objects Temporal Location (\textbf{OTL}), Objects Temporal Relation (\textbf{OTR}), Objects Spatial Location (\textbf{OSL}), Objects Spatial Relation (\textbf{OSR}), Objects Spatial Tracking (\textbf{OST}), Human Activity Analysis (\textbf{HAA}), Anomaly Detection (\textbf{AD}). 
}
\vspace{3pt}
\scalebox{0.55}{
\begin{tabular}{l r rrrrrrr rrrrrr rrr r r r}
\toprule
\multirow{2}{*}{\textbf{Model}} & \multirow{2}{*}{\textbf{Avg.}} & \multicolumn{8}{c}{\bf Content} & \multicolumn{6}{c}{\bf Temporal}  & \multicolumn{3}{c}{\bf Spatial} & \multirow{2}{*}{\textbf{HAA}} & \multirow{2}{*}{\textbf{AD}} \\
\cmidrule(lr){3-10} \cmidrule(lr){11-16} \cmidrule(lr){17-19}
&& \textbf{OE} & \textbf{OC} & \textbf{AC} & \textbf{DD} & \textbf{BD} & \textbf{ED} & \textbf{ES} & \textbf{OCR} & \textbf{AR} & \textbf{AS} & \textbf{AL} & \textbf{EL} & \textbf{OTL} & \textbf{OTR} & \textbf{OSL} & \textbf{OSR} & \textbf{OST}\\
\midrule
VideoChat+ChatGPT & 16.6 & 35.4 & 18.3 & 6.1 & 26.5 & 43.4 & 6.2 & 13.8 & 6.6 & 12.1 & 23.5 & 5.0 & 8.5 & 8.0 & 20.1 & 2.1 & 9.4 & 7.0 & 15.8 & 50.0 \\
{Video-LLaMA} & 25.4 & 26.9 & 30.4 & 24.6 & 26.3 & 25.3 & 23.9 & 24.8 & 24.7 & 25.2 & 25.4 & 24.5 & 24.7 & 24.0 & 27.6 & 25.7 & 27.8 & 25.2 & 27.7 & 11.1\\ 
{Video-ChatGPT} & 36.1 & 49.6 & 23.5 & 26.7 & 45.8 & 40.3 & 50.4 & 29.9 & 29.9 & 42.1 & 33.2 & 30.2 & 35.2 & 25.3 & 36.2 & 27.8 & 26.1 & 37.7 & 53.5 & 10.2\\
{IVA} & 37.4 & 37.9 & 21.7 & 23.4 & 44.9 & 58.9 & 50.6 & 32.3 & 29.3 & 44.6 & 33.8 & 31.1 & 35.0 & 26.0 & 34.6 & 25.2 & 27.7 & 35.8 & 54.5 & 16.7\\
{Video-LLaVA}  & 53.8 & 66.6 & 35.7 & 19.8 & 78.6 & 84.2 & 75.0 & 46.1 & 43.6 & 71.5 & 63.3 & 32.8 & 39.2 & 29.3 & 49.0 & 28.4 & 37.1 & 50.8 & 87.1 & 16.7\\
{LLaMA-VID} & 54.0 & 64.9 & 37.4 & 22.5 & 65.4 & 87.7 & 75.7 & 48.4 & 44.1 & 71.1 & 62.4 & 34.4 & 43.1 & 34.7 & 48.5 & 28.8 & 36.2 & 50.7 & 91.1 & 17.6\\
{LLaVA-NeXT-Video} & 54.1 & 68.3 & 32.2 & 23.1 & 65.8 & 86.3 & 70.6 & 45.1 & 51.9 & 71.8 & 59.9 & 35.3 & 35.1 & 43.3 & 53.5 & 34.4 & 37.3 & 50.3 & 81.2 & 16.7\\
{VideoChat2-Vicuna} & 51.8 & 73.7 & 20.9 & 21.3 & 79.7 & 85.1 & 67.2 & 39.0 & 46.1 & 70.3 & 57.9 & 28.0 & 34.2 & 28.7 & 45.6 & 31.8 & 34.4 & 39.6 & 85.1 & 9.0\\
{VideoChat2-Mistral} & \textbf{54.9} & 64.8 & 43.5 & 24.1 & 83.3 & 89.1 & 73.4 & 52.3 & 48.7 & 75.2 & 68.6 & 32.7 & 31.7 & 20.0 & 49.1 & 32.4 & 32.7 & 42.6 & 89.1 & 4.6\\


\midrule
{Gemini-1.5-Flash} & 74.7 & 80.9 & 61.8 & 41.8 & 95.4 & 98.5 & 96.3 & 77.9 & 85.7 & 93.7 & 84.8 & 58.6 & 69.1 & 67.4 & 59.9 & 47.4 & 55.2 & 55.2 & 95.6 & 30.0 \\
{GPT-4o} & \textbf{75.2} & 83.1 & 54.8 & 40.8 & 96.0 & 98.6 & 94.7 & 78.5 & 86.3 & 94.3 & 87.6 & 50.3 & 67.3 & 44.0 & 64.5 & 46.9 & 61.9 & 59.8 & 97.0 & 55.6\\

\bottomrule
\end{tabular}%
}
\label{tab:task_understanding} 
\end{table}

\begin{table}[t]
\centering
\caption{\textbf{Comparison of various models on Video Reasoning Tasks.} Relation Reasoning Image-Video (\textbf{I-V}), Relation Reasoning Video-Video (\textbf{V-V}), Temporal Reasoning Event Prediction (\textbf{EP}), Temporal Reasoning Action Prediction (\textbf{TP-R}), Causal Reasoning (\textbf{CAR}), Counterfactual Reasoning (\textbf{CFR}), Commonsense Reasoning (\textbf{CSR}), Logic Reasoning (\textbf{LR}). 
}
\vspace{3pt}
\scalebox{0.83}{
\begin{tabular}{l r rr rr rrrr}
\toprule
\multirow{2}{*}{\textbf{Model}} & \multirow{2}{*}{\textbf{Avg.}} & \multicolumn{2}{c}{\bf Relation} & \multicolumn{2}{c}{\bf Temporal}  & \multirow{2}{*}{\textbf{CAR}} & \multirow{2}{*}{\textbf{CFR}} & \multirow{2}{*}{\textbf{CSR}} & \multirow{2}{*}{\textbf{LR}} \\
\cmidrule(lr){3-4} \cmidrule(lr){5-6} 
&& \textbf{I-V} & \textbf{V-V} & \textbf{EP} & \textbf{AP} \\



\midrule
{VideoChat+ChatGPT} & 23.0 & - & - & 19.8 & 31.4 & 5.8 & 49.4 & 17.9 & 26.9\\
{Video-LLaMA} & 25.2 & - & - & 23.1 & 24.3 & 26.1 & 26.5 & 25.5 & 24.4\\ 
{Video-ChatGPT} & 38.7 & - & - & 34.9 & 33.0 & 52.4 & 38.1 & 36.8 & 29.3\\
{IVA} & 48.4 & - & - & 41.2 & 42.8 & 55.8 & 58.5 & 50.3 & 34.6\\
{Video-LLaVA} & 66.9 & - & - & 60.8 & 61.1 & 76.0 & 74.8 & 69.8 & 50.1\\
{LLaMA-VID} & \textbf{67.6} & - & - & 59.3 & 61.8 & 77.5 & 83.7 & 68.3 & 46.4\\
{LLaVA-NeXT-Video} & 66.1 & - & - & 51.3 & 58.4 & 73.8 & 86.9 & 68.2 & 50.6\\
{VideoChat2-Vicuna} & 60.6 & - & - & 50.0 & 53.7 & 70.6 & 65.5 & 64.8 & 47.3\\
{VideoChat2-Mistral} & 65.9 & - & - & 50.0 & 57.9 & 73.5 & 86.4 & 66.3 & 55.8\\
\midrule
{Gemini-1.5-Flash} & 82.3 & 85.8 & 30.2 & 66.8 & 73.5 & 95.2 & 96.8 & 93.0 & 65.2\\
{GPT-4o} & \textbf{88.0} & 94.8 & 56.0 & 83.7 & 85.3 & 93.1 & 97.8 & 91.3 & 77.0\\
\bottomrule
\end{tabular}}
\label{tab:task_reasoning} 
\end{table}


\paragraph{Video Understanding.}
As shown in Table~\ref{tab:task_understanding}, there is no significant performance gap between the GPT-4o and the Gemini-1.5-Flash in understanding. 
The performance of GPT-4o is worse on tasks that require answering a specific time, such as Action/Object/Temporal Location, especially on the first two tasks. This is likely because the GPT-4o model uses a uniform sampling of 100 frames as input when testing videos longer than 100 seconds, while the Gemini-1.5-Flash model uses a one-frame-per-second encoding method. This endows the Gemini-1.5-Flash with stronger temporal perception and detail capture capabilities.


Meanwhile, VideoChat2-Mistral leads open-source Video-LLMs with a performance of 54.9\%, highlighting its superior capability in video understanding. Recent open-source Video-LLMs also show commendable performance in macro-analysis tasks like Brief Description. However, these models struggle with tasks requiring detailed video information, such as Objects Temporal Location and Spatial Relation. Most of these open-source Video-LLMs use a simple fixed frame encoding method when processing videos, and they often extract only a small number of frames from the videos. This approach is very disadvantageous when it comes to answering tasks that require specific video details. Overall, there remains a notable gap in video understanding between current open-source Video-LLMs and mature commercial models like GPT-4o.

\paragraph{Video Reasoning.}
The results in the Table~\ref{tab:task_reasoning} indicate that GPT-4o achieves the highest results on the Video Reasoning Task, outperforming Gemini-1.5-Flash by 5.7\%. It is worth noting that the GPT-4o and Gemini-1.5-Flash both demonstrate subpar performance on the Relation Reasoning Video-Video task, struggling to effectively reason about relationships between input videos. This may be attributed to a lack of complex multi-video data during their training processes. 
In the realm of open-source Video-LLMs, LLaMA-VID achieves the highest evaluation scores in Video Reasoning tasks. The primary differences between these open-source models and commercial Video-LLMs are observed in the Causal Reasoning and Commonsense Reasoning tasks, possibly due to differences in the parameters and performance of the language models.
From the experimental results, it can be observed that the capabilities of open-source Video-LLMs in coarse-grained Event Prediction are already quite close to those of the Gemini-1.5-Flash model. However, there is still a significant gap in more fine-grained Action Prediction and Logic Reasoning tasks.

\subsection{Detailed Analysis}


%

\begin{figure}[t]
    \centering
    \begin{subfigure}[b]{0.42\textwidth}
        \centering
        \includegraphics[width=\textwidth]{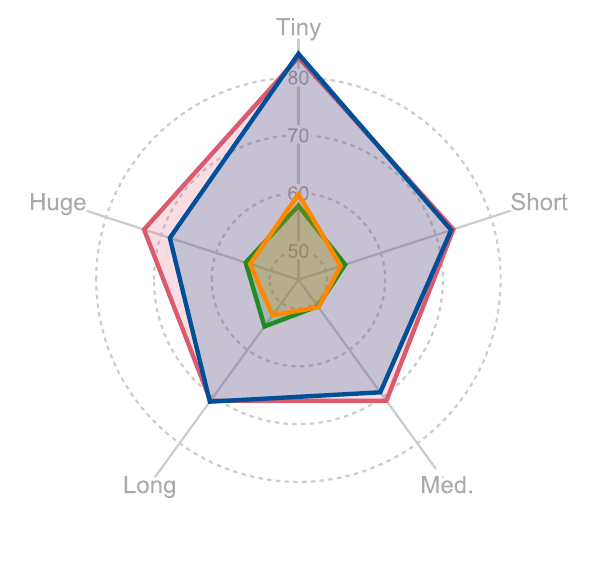}
        \caption{Evaluation results divided by Duration.}
        \label{fig:sub1}
    \end{subfigure}
    \hfill
    \begin{subfigure}[b]{0.42\textwidth}
        \centering
        \includegraphics[width=\textwidth]{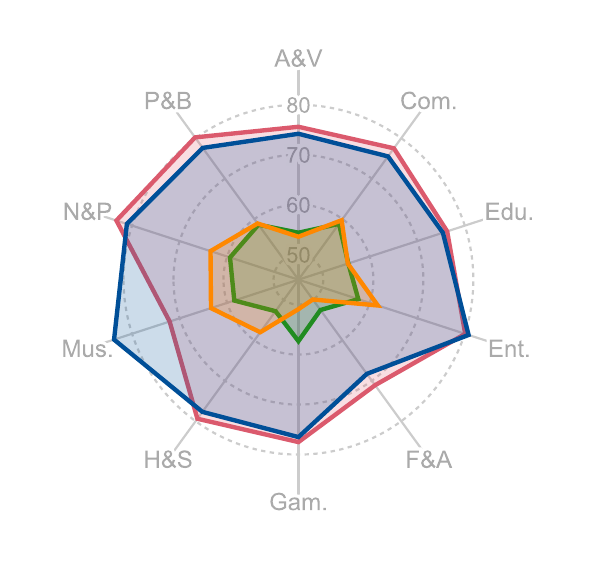}
        \caption{Evaluation results divided by Category.}
        \label{fig:sub2}
    \end{subfigure}
    \caption{Detailed analysis of Video-LLMs: {\color{RadarRed}GPT-4o}, {\color{RadarBlue}Gemini-1.5-Flash}, \textcolor{forestgreen}{LLaMA-VID}, {\color{orange}VideoChat2-Mistral}. (a) shows the video understanding performance divided by Duration and (b) presents the overall performance divided by Category. The Duration in (a): 0-60s (\textbf{Tiny}), 60-120s (\textbf{Short}), 120-300s (\textbf{Med.}), 300-600s (\textbf{Long}), 600s+ (\textbf{Huge}).}
    \label{fig:results_durations_and_categories}
\end{figure}

\paragraph{Duration.}
As shown in (a) of Figure~\ref{fig:results_durations_and_categories}, we can see that once the video length exceeds 1 minute, the performance of almost all models on the Video Understanding tasks declines significantly. This decline also occurs in Gemini-1.5-Flash and LLaMA-VID, which use a one-frame-per-second encoding method. It indicates that, besides the video frames encoding method, current Video-LLMs struggle with understanding long contextual information when processing long videos.

\paragraph{Category.} Based on the experimental results in (b) of Figure~\ref{fig:results_durations_and_categories}, we find that Video-LMMs exhibit some variability in performance across different categories. Gemini-1.5-Flash surpasses GPT-4o in the Music category, possibly due to its capability to process video audios. Similarly, open-source Video-LLMs like Videochat2-mistral also struggle in the Music category due to their inability to incorporate audio information from videos.

\begin{wraptable}{r}{0.6\textwidth}
\vspace{-10pt}
\centering
\footnotesize
\caption{\textbf{Evaluation results on other dataset after model trained on VideoVista.} ${\clubsuit}$ indicates the model is trained with same-size Video-ChatGPT-100K. SEED\textsuperscript{PU}: SEED-Bench Proceduce Understanding.}
\begin{tabular}{l c c c c}
\toprule
\textbf{Model}  & \textbf{SIQ-2} & \textbf{LQA} & \textbf{WildQA} & \textbf{SEED\textsuperscript{PU}}\\
\midrule
Video-LLaMA & 55.8 & 35.8 & - & -\\
Video-ChatGPT$^{\clubsuit}$ & 57.5 & 33.9 & - & 21.1\\
IVA$^{\clubsuit}$ & 54.0 & 46.5 & 51.2 & 27.5\\
IVA-VideoVista & \textbf{61.9} & \textbf{47.3} & \textbf{55.1} & \textbf{38.0}\\ 
\bottomrule
\end{tabular}
\label{tab:result_datavalue} 
\end{wraptable}

\paragraph{Data Value.} We retrain the IVA model using all 93K constructed data without manual filtering and evaluate it on dataset Social-IQ2 ~\cite{siq2}, LifeQA~\cite{castro-etal-2020-lifeqa}, WildQA~\cite{castro-etal-2022-in-the-wild} and SEED-Bench~\cite{li2023seed}. Table~\ref{tab:result_datavalue} shows that the model trained on VideoVista achieves significant improvements compared to those trained on the same-size video dataset, i.e., Video-ChatGPT-100K dataset. Moreover, the comparative results suggest the effectiveness of our automatic annotation method of videos.


\section{Related Work}

\paragraph{Video-LMMs: LLMs for Video Understanding}
Recent progress in LLMs has led to revolutionary achievements in in-context learning \cite{zhang2023dnagpt} and the modelling of extensive contexts~\cite{lyu2023gpt}. This technological evolution has opened new avenues for merging LLMs with computer vision technologies, as exemplified by projects such as Visual-ChatGPT~\cite{wu2023visual}, LMEye~\cite{li2023lmeye}, and InstructBLIP~\cite{dai2023instructblip}. These models surpass conventional limitations by integrating vision model APIs~\cite{qin2023tool}, thereby solving intricate problems within the field of computer vision. Fusing language models with video understanding technologies~\cite{maaz2023video,zhang2023llama,li2023videochat,xu2023retrieval,song2023moviechat,li2024llms,pan2023retrieving,wang2024videotree} has significantly improved multimodal interpretation, allowing for sophisticated analyses of the relationships between visuals and textual elements. Additionally, applying LLMs to video understanding tasks~\cite{tang2023video_survey} represents an important advancement in utilizing their capacity to analyze and reason about visual information effectively.


\paragraph{Video QA Datasets}
VideoQA tasks have led to the creation and adaptation of several datasets, which are designed to address different aspects of video comprehension. MSVD-QA~\cite{xu2017video}, built on the MSVD dataset, focuses on content-specific questions, shifting from mere caption generation to interactive video understanding. MSRVTT-QA~\cite{xu2016msr} utilizes narrative-style video descriptions to provide a controlled environment for examining QA mechanisms. TGIF-QA~\cite{li2016tgif} introduces a unique angle by emphasizing temporal reasoning and the understanding of repetitive actions. ActivityNet-QA~\cite{yu2019activitynet} encourages comprehension of prolonged and complex activities. PororoQA~\cite{kim2017deepstory} leverages animated children's stories to explore story-based video understanding, while TVQA~\cite{lei2018tvqa} utilizes long-form TV show episodes to create a multimodal challenge that involves intricate plots and character interactions.  Recently, with the rapid development of large models that take natural language question-answer as the interaction scope, diverse multimodal tasks are becoming unified into a unified benchmark~\cite{li2023comprehensive,ma2023vista,yue2023mmmu}. MVBench~\cite{li2023mvbench} is a typical diverse video understanding benchmark including 20 tasks. 

\section{Discussion and Future Work}

The evaluation results on VideoVista demonstrate that current open-source Video-LLMs still have considerable room for improvement and reveal some potential directions for enhancement.

\textbf{Improving the encoding method for long videos.} The assessment on VideoVista indicates that open-source Video-LLMs need significant improvements, suggesting two main enhancements: 1) Enhancing long-context processing by increasing context data during training to better manage long videos, and 2) Optimizing video frame downsampling by minimizing the representation tokens per frame while preserving the video's original semantics, as current encoding methods with a fixed frame count lead to suboptimal performance in detailed video analysis and temporal tasks.



\textbf{Introducing information from more modalities.} Our experimental results indicate a significant improvement in model performance on certain categories with the addition of audio information. Additionally, audio information is indispensable for the model to correctly understand video content. With the rise of Mixture of Experts (MoE) technology, integrating information from different modalities through different modal experts has become a potential solution.


\section{Limitations}
The proposed method has several limitations: 1) The VideoVista dataset’s maximum video length of 919 seconds does not cater adequately to real-world applications involving longer content like movies. There is a pressing need to expand the dataset to encompass longer videos. 2) Utilizing GPT-4o for video annotation can lead to errors due to 'hallucinations' caused by insufficient data. Developing efficient strategies to minimize these errors is crucial and remains a significant challenge.

\section{Conclusion}

We have developed a comprehensive video evaluation benchmark for Video-LLMs that covers both video understanding and reasoning across 27 tasks. This benchmark includes standard tasks such as action recognition and video description, as well as more complex tasks like Action Location and Relation Reasoning. Additionally, we introduce an effective automatic data construction framework that employs tools like GPT-4o for efficient annotation of long videos. Our approach provides a robust framework for assessing and enhancing the capabilities of Video-LLMs.

\begin{ack}
We would like to acknowledge the efforts of the reviewers and editors for checking our work. We also thank all paper authors for contributing to data collection and quality control, appreciating the support of funding (xxx). 
\end{ack}

\bibliographystyle{plain}
\bibliography{main}


\newpage
\appendix

\section{Dataset Document}

\subsection{Usage}
\begin{itemize}
    \item VideoVista can be reviewed at \textcolor{blue}{\url{https://huggingface.co/datasets/VideoVista/VideoVista}}
    
    \item License: VideoVista is under apache-2.0 license.
\end{itemize}

\subsection{Data Sources}
\begin{figure}[h]
\vspace{0.15cm}
\centering
\includegraphics[width=0.7\textwidth]{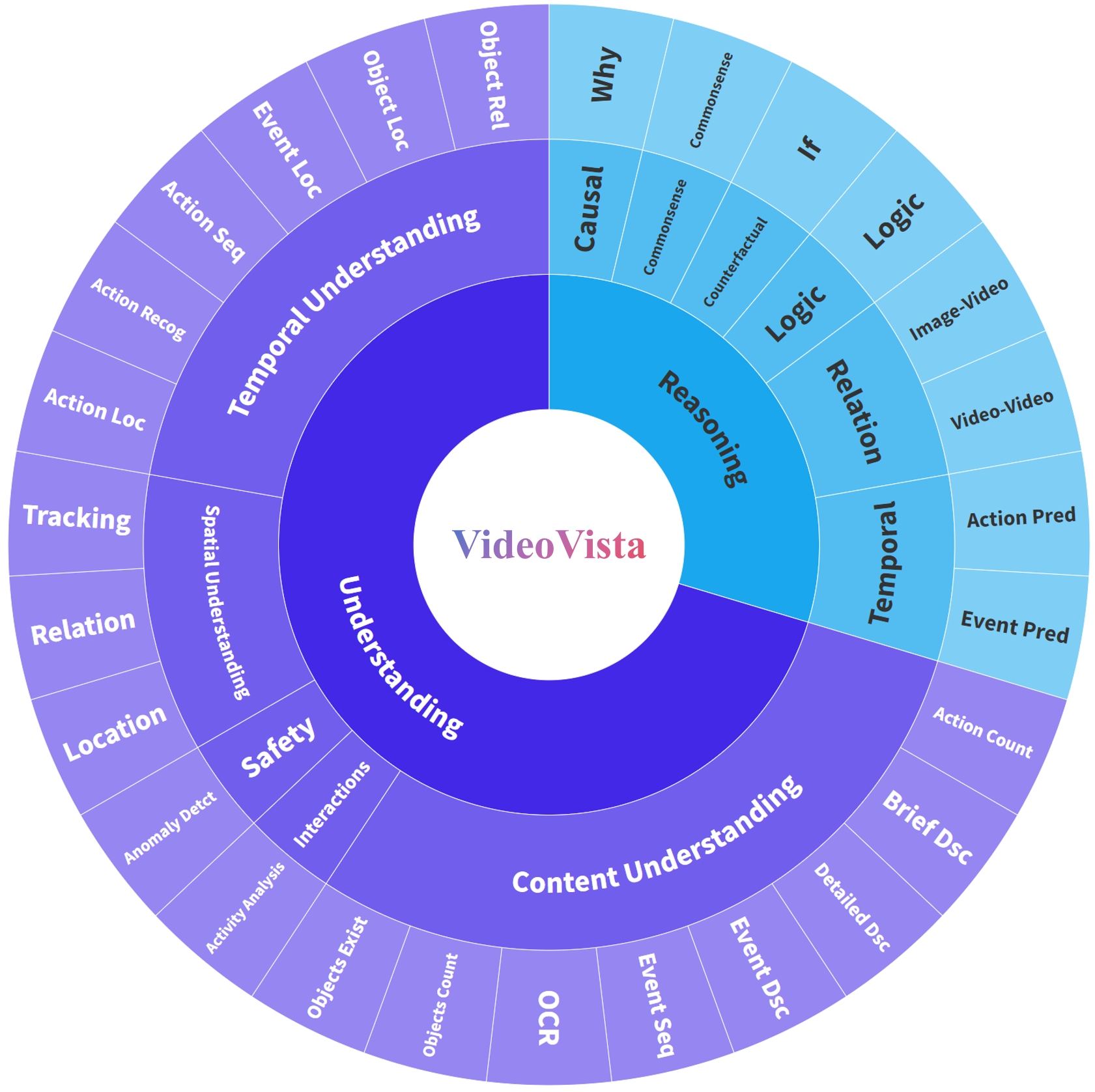}
\vspace{0.1cm}
\caption{All 27 task types across VideoVista.}
\label{fig:task_ability}
\vspace{-0.5cm}
\end{figure}

The dataset includes 24,906 questions and their corresponding answers, all of which were annotated and generated by the generative large language model GPT-4. The 894 video sources used in the dataset all come from YouTube and we randomly selected them from the test set of Panda-70M. As shown in Fugure~\ref{fig:task_ability}, we introduce {\em 19 types of understanding tasks} such as Description (Short-Long), Event (Location, Sequence, and Description), Actions (Location, Recognition, Sequence, and Count), and Objects (Existence, Count, Location (Temporal-Spatial), Objects Relation (Temporal-Spatial), and Tracking), Optical Character Recognition, Human Activity Analysis, and Anomaly Detection; and {\em 8 types of Reasoning tasks}: Causal Reasoning, Temporal Reasoning (Actions and Events), Logical Reasoning, Counterfactual Reasoning, Commonsense Reasoning, and Relation Reasoning (Image/Video-Video Relation).

The specific example of each ability is given in Table~\ref{tab:task_dimension}.


\begin{table}[tp]
    \centering
    \caption{
    \textbf{Task examples of VideoVista for every task category}.
}
\vspace{3pt}
    \label{tab:task_dimension}
    \scalebox{0.7}{
        \begin{tabular}{c c c p{11cm}}
        \toprule
        \textbf{Ability} & \textbf{Task Type} & \textbf{Video Source (Secs)} & \textbf{Example} \\
        \midrule
        \multirow{16}{*}{\textbf{\makecell[c]{Content\\ Understanding}}} & Object & \multirow{2}{*}{\makecell[c]{People \& Bog \\(148)}} & \textit{\blue{Are there any stuffed animals visible in the video?}} \\
         & Existence & ~ & (A) No stuffed animals. (B) \dots (C) \dots (D) \dots \\
        \cdashline{2-4}\noalign{\vskip 0.5ex}
        ~ & Object & \multirow{2}{*}{\makecell[c]{Howto \& Style \\ (60)}} &    \textit{\blue{How many pink shoes appear in the video from 0 to 14 seconds?}} \\
        ~ & Count & ~ &   (A) Four pink shoes. (B) \dots (C) \dots (D) \dots \\
        \cdashline{2-4}\noalign{\vskip 0.5ex}
        ~ & Action & \multirow{2}{*}{\makecell[c]{Comedy \\(71)}} &   \textit{\blue{How many different actions does it take for the man with \dots hit by a broom?}} \\
        ~ & Count & ~ &   (A) four. (B) \dots (C) \dots (D) \dots \\
        \cdashline{2-4}\noalign{\vskip 0.5ex}
        ~ & \multirow{2}{*}{\makecell[c]{Detailed \\ Description}} & \multirow{2}{*}{\makecell[c]{People \& Blogs\\ (758)}} &   \textit{\blue{What are the detailed steps taken in the video to prepare the dish?}} \\
        ~ & ~ & ~ &  (A) Heating oil in a purple pan, sautéing vegetables (B) \dots (C) \dots (D) \dots \\
        \cdashline{2-4}\noalign{\vskip 0.5ex}
        ~ & \multirow{2}{*}{\makecell[c]{Brief \\ Description}} & \multirow{2}{*}{\makecell[c]{Pets \& Animals \\ (35) }} &    \textit{\blue{What happens at the video?}}  \\
        ~ & ~ & ~ &   (A) A person extends their hand towards two Siberian Huskies. (B) \dots (D) \dots \\
        \cdashline{2-4}\noalign{\vskip 0.5ex}
        ~ & \multirow{2}{*}{\makecell[c]{Event \\Description}} & \multirow{2}{*}{\makecell[c]{Education \\(105)} } &   \textit{\blue{How does the speaker describe the transformation brought by the women \dots?}}  \\
        ~ & ~ & ~ &   (A) They saw themselves as victims and didn't take action. (B) \dots (D) \dots \\
        \cdashline{2-4}\noalign{\vskip 0.5ex}
        ~ & \multirow{2}{*}{\makecell[c]{Event\\Sequence}} & \multirow{2}{*}{\makecell[c]{Film \& Animation \\(172)} } &    \textit{\blue{Can you list the events in the order they occur in the video?}}  \\
        ~ & ~ & ~ &   (A) A woman in a white dress searches for 'Daniel' [......] (B) \dots (D) \dots \\
        \cdashline{2-4}\noalign{\vskip 0.5ex}
        ~ & \multirow{2}{*}{OCR} & \multirow{2}{*}{\makecell[c]{News \& Politics \\(10)} } &    \textit{\blue{What is written on the screen at the start of the video?}} \\
        ~ & ~ & ~ &   (A) ALBINO COBRA (B) \dots (C) \dots (D) \dots \\
        \midrule
        \multirow{12}{*}{\textbf{\makecell[c]{Temporal\\Understanding}}} & \multirow{2}{*}{\makecell[c]{Action\\ Recognition}} & \multirow{2}{*}{\makecell[c]{People \& Blogs \\ (35)}} &  \textit{\blue{What action does the person with \dots perform in the early part of the video?}} \\
        ~ & ~ & ~ & (A) painting daisies on a canvas (B) \dots (C) \dots (D) \dots\\
        \cdashline{2-4}\noalign{\vskip 0.5ex}
        ~ & \multirow{2}{*}{\makecell[c]{Action \\ Sequence}} & \multirow{2}{*}{\makecell[c]{Gaming \\ (90)}} &    \textit{\blue{Describe the sequence of actions performed from the beginning to the end \dots}}\\
         ~ & ~ & ~ & (A) Green-haired man watches, groups march, blue [......] (B) \dots (D) \dots\\
        \cdashline{2-4}\noalign{\vskip 0.5ex}
        ~ & \multirow{2}{*}{\makecell[c]{Action \\ Location}} & \multirow{2}{*}{\makecell[c]{Science \& Technology\\ (28)}} &    \textit{\blue{At what time \dots person's hands start using pliers to work with a metal wire?}} \\
         ~ & ~ & ~ & (A) from 6 to 8 seconds (B) \dots (C) \dots (D) \dots \\
        \cdashline{2-4}\noalign{\vskip 0.5ex}
        ~ & \multirow{2}{*}{\makecell[c]{Event \\ Location}} & \multirow{2}{*}{\makecell[c]{Pets \& Animals \\ (110)}} &   \textit{\blue{When does the individual in the blue shirt hold a \dots puppy inside the vehicle?}} \\
         ~ & ~ & ~ & (A) Around 75-85 seconds (B) \dots (C) \dots (D) \dots \\
        \cdashline{2-4}\noalign{\vskip 0.5ex}
        ~ & \multirow{2}{*}{\makecell[c]{Object \\Temporal Location}} & \multirow{2}{*}{\makecell[c]{Travel \& Events \\(10)}} &    \textit{\blue{Which second does the woman start placing squash on the plate?}} \\
        ~ & ~ & ~ &   (A) 10th second (B) \dots (C) \dots (D) \dots\\
        \cdashline{2-4}\noalign{\vskip 0.5ex}
        ~ & \multirow{2}{*}{\makecell[c]{Object \\Temporal Relation}} &  \multirow{2}{*}{\makecell[c]{Film \& Animation \\ (98)}} &   \textit{\blue{How does the brick building visually change throughout the video?}} \\
         ~ & ~ & ~ & (A) It changes color and texture. (B) \dots (C) \dots (D) \dots\\
        \midrule
        \multirow{6}{*}{\textbf{\makecell[c]{Spatial\\ Understanding}}} & \multirow{2}{*}{\makecell[c]{Object \\Spatial Location}} & \multirow{2}{*}{\makecell[c]{Education \\ (61) }} &    \textit{\blue{What is the bounding box coordinates of the tree at 14 second of the video?}} \\
        ~ &  ~ & ~ &  (A) [0.3, 0.0, 0.71, 1.0] (B) \dots (C) \dots (D) \dots\\
        \cdashline{2-4}\noalign{\vskip 0.5ex}
        ~ & \multirow{2}{*}{\makecell[c]{Objects \\Spatial Relation}} & \multirow{2}{*}{\makecell[c]{Film \& Animation\\(284)}} &    \textit{\blue{How does the pink pig character with a red dress typically hold the football?}} \\
        ~ & ~ & ~ &   (A) On her head, balancing it. (B) \dots (C) \dots (D) \dots \\
        \cdashline{2-4}\noalign{\vskip 0.5ex}
        ~ & \multirow{2}{*}{\makecell[c]{Object \\Spatial Tracking}} & \multirow{2}{*}{\makecell[c]{Film \& Animation \\ (127)} } &    \textit{\blue{What direction is the car moving in the video?}} \\
        ~ & ~ & ~ &  (A) towards the right (B) \dots (C) \dots (D) \dots \\
        \midrule
        \multirow{2}{*}{\makecell[c]{\textbf{Understanding}}} & \multirow{2}{*}{\makecell[c]{Human Activity\\ Analysis}} & \multirow{2}{*}{\makecell[c]{Travel \& Events\\ (100)}} &    \textit{\blue{Why are various people gathered in the video?}} \\
        ~ & ~ & ~ &   (A) To visit a historical monument. (B) \dots (C) \dots (D) \dots \\
        \midrule
         \multirow{2}{*}{\makecell[c]{\textbf{Understanding}}} & \multirow{2}{*}{\makecell[c]{Anomaly \\ Detection}} & \multirow{2}{*}{\makecell[c]{Travel \& Events\\ (336)}} &    \textit{\blue{Which of the following violations might be present in the video?}} \\
        ~ & ~ & ~ &   (A) Sexual. (B) \dots (C) \dots (D) \dots (E) \dots\\
        \midrule
        \multirow{4}{*}{\textbf{\makecell[c]{Relation\\ Reasoning}}} & \multirow{2}{*}{Image-Video} &  \multirow{2}{*}{\makecell[c]{News \& Politics\\(12)}} &    \textit{\blue{What role does the person in the image portray in the video?}} \\
        ~ & ~ & ~& (A) A teacher (B) \dots (C) \dots (D) \dots\\
        \cdashline{2-4}\noalign{\vskip 0.5ex}
        ~ & \multirow{2}{*}{Video-Video} &  \multirow{2}{*}{\makecell[c]{Sports \\ (7 and 10)}} &    \textit{\blue{What's the relationship between video 1 and video 2?}}  \\
        ~ & ~ & ~& (A) video 1 is before video 2 (B) \dots (C) \dots (D) \dots\\
        \midrule
        \multirow{4}{*}{\textbf{\makecell[c]{Temporal\\ Reasoning}}} & \multirow{2}{*}{\makecell[c]{Event \\Prediction}} & \multirow{2}{*}{\makecell[c]{Howto \& Style \\(73)}} &  \textit{\blue{What will the person do next after concluding their exercise on the yoga mat?}} \\
        ~ & ~ & ~& (A) The person will start doing jumping jacks beside the mat. (B) \dots (D) \dots \\
        \cdashline{2-4}\noalign{\vskip 0.5ex}
        ~ & \multirow{2}{*}{\makecell[c]{Action\\ Prediction}} &  \multirow{2}{*}{\makecell[c]{Film \& Animation \\ (107)}} &    \textit{\blue{What action is expected to occur immediately after the video concludes?}}  \\
        ~ & ~ & ~& (A) The drawing man in blue and the skunk will walk together. (B) \dots (D) \dots\\
        \midrule
        \multirow{2}{*}{\textbf{\makecell[c]{Causal\\ Reasoning}}} & \multirow{2}{*}{Why} &  \multirow{2}{*}{\makecell[c]{Comedy \\(593)}} &    \textit{\blue{Why do the people in the last scene seem excited?}} \\
        ~ & ~ & ~ & (A) They received unexpected gifts. (B) \dots (C) \dots (D) \dots \\
        \midrule
        \multirow{2}{*}{\textbf{\makecell[c]{Counterfactual\\ Reasoning}}} & \multirow{2}{*}{Counterfactual} &  \multirow{2}{*}{\makecell[c]{Sports\\ (63)}} &    \textit{\blue{Imagine if the SUV \dots had proper winter tires instead of \dots, what is the result?}} \\
        ~ & ~ & ~& (A) It has a smoother ride on the snowy surface. (B) \dots (C) \dots (D) \dots\\
        \midrule
        \multirow{2}{*}{\textbf{\makecell[c]{Commonsense\\ Reasoning}}} & \multirow{2}{*}{Commonsense} &  \multirow{2}{*}{\makecell[c]{Entertainment \\(64)}} &    \textit{\blue{What type of fruit is being shown at the market in the video?}} \\
        ~ & ~ & ~& (A) Mangosteen (B) \dots (C) \dots (D) \dots\\
        \midrule
        \multirow{2}{*}{\textbf{\makecell[c]{Logical\\ Reasoning}}} & \multirow{2}{*}{Logic} &  \multirow{2}{*}{\makecell[c]{Science \& Technology\\ (70)}} &    \textit{\blue{What will happen right after this video comes to an end?}} \\
        ~ & ~ & ~& (A) A woman will be presenting a new armor design on a stage. (B) \dots (D) \dots\\
        \bottomrule
        \end{tabular}
    }

\end{table}

\subsection{Data Processing Guidelines}
\label{guidlines_const}
We present the overall data construction process from video downloading, preprocessing, and question-answers generation.

\paragraph{Video Downloading}
The videos utilized in VideoVista are sampled from the test set of the Panda-70M dataset. Every video in the Panda-70M can be identified by a unique YouTube video ID. To access a specific video, prepend the ID with 'https://www.youtube.com/watch?v='. For downloading these videos, we employ Python library \textit{youtube-utils}, which enables direct downloading and local saving of videos using their video IDs. Additionally, through the \textit{extract-info} method of the \textit{YoutubeDL} class, we can also obtain auxiliary information such as the title, category, tags, and description of the corresponding YouTube video.

\paragraph{Video Processing}

The processing of the videos consists of video splitting, merging, and pre-annotation. We present the details of each processing step in the following paragraph.

\textbf{Video splitting.}
For video splitting, the first step is Semantics-aware Video Splitting. 
We follow the method outlined in Panda-70M\cite{chen2024panda70m} for semantics-aware video splitting. This two-stage algorithm divides a long video into semantically coherent clips: 1) Shot Boundary Detection: In the first stage, we split the video based on shot boundary detection, as semantic content often changes with the start of a new scene. 2) Stitching Adjacent Clips: In the second stage, adjacent clips that were incorrectly separated in the first stage are merged to avoid overly short segments. This is achieved by using ImageBind\cite{girdhar2023imagebind} to extract frame embeddings and merge clips with similar embeddings. Different from Panda-70M, we did not filter out videos with complex transitions or redundant clips to maintain continuity for later merging.

The second for video splitting is to use CLIP\cite{xue2022clip} to split videos longer than 40 seconds, so that the video clips can be annotated using GPT series tools. Specifically, we extract one frame per second and use CLIP to obtain frame features. Starting with the 0th frame, we calculate cosine similarity to identify the most deviating frame among the next 39 frames as the $i$th frame. This can create a sub-clip of less than 40 seconds from the 0th to the $(i-1)$th frame. We will continue this process from the $i$th frame until all sub-clips are shorter than 40 seconds. Using this method, all the video clips we obtain are within 40 seconds. The clip model we used here is CLIP-ViT-L-patch14.

\textbf{Video merging.}
For video merging, we combined videos specifically for the four-time ranges of the 60s-120s, 120s-300s, 300s-600s, and 600s+. While merging videos into a specific time range, we ensured that at least one clip in the merged video had an action annotation that was 'none'. We also ensure that each merged video has no same video clips within the same range of time intervals. With the help of this strategy, we successfully merged video clips into videos of different lengths.

\textbf{Video pre-annotation.}
First, we use Whisper~\cite{radford2023robust} to extract the audio information from the video. The model we use is from \url{https://github.com/m-bain/whisperX}. With the help of Whisper, we can efficiently extract audio information from video clips and convert it into transcripts, which can then serve as auxiliary information for subsequent annotation. Next, we use the Segment Anything~\cite{kirillov2023segment} to perform object detection and labelling on the previously extracted frame images. We use the model from \url{https://github.com/IDEA-Research/Grounded-Segment-Anything}, specifically employing the Grounded-SAM with RAM pipeline. With this pipeline, we can obtain labels and bounding boxes corresponding to all objects in the image. After filtering out objects that occupy too much (usually the background) or too little space in the image, we use them as auxiliary information during annotation. In addition, we also download the video title and category, which users write, as the topic information to help GPT-4o understand the topics of video clips.

\begin{figure}
    \includegraphics[width=1.1\textwidth]{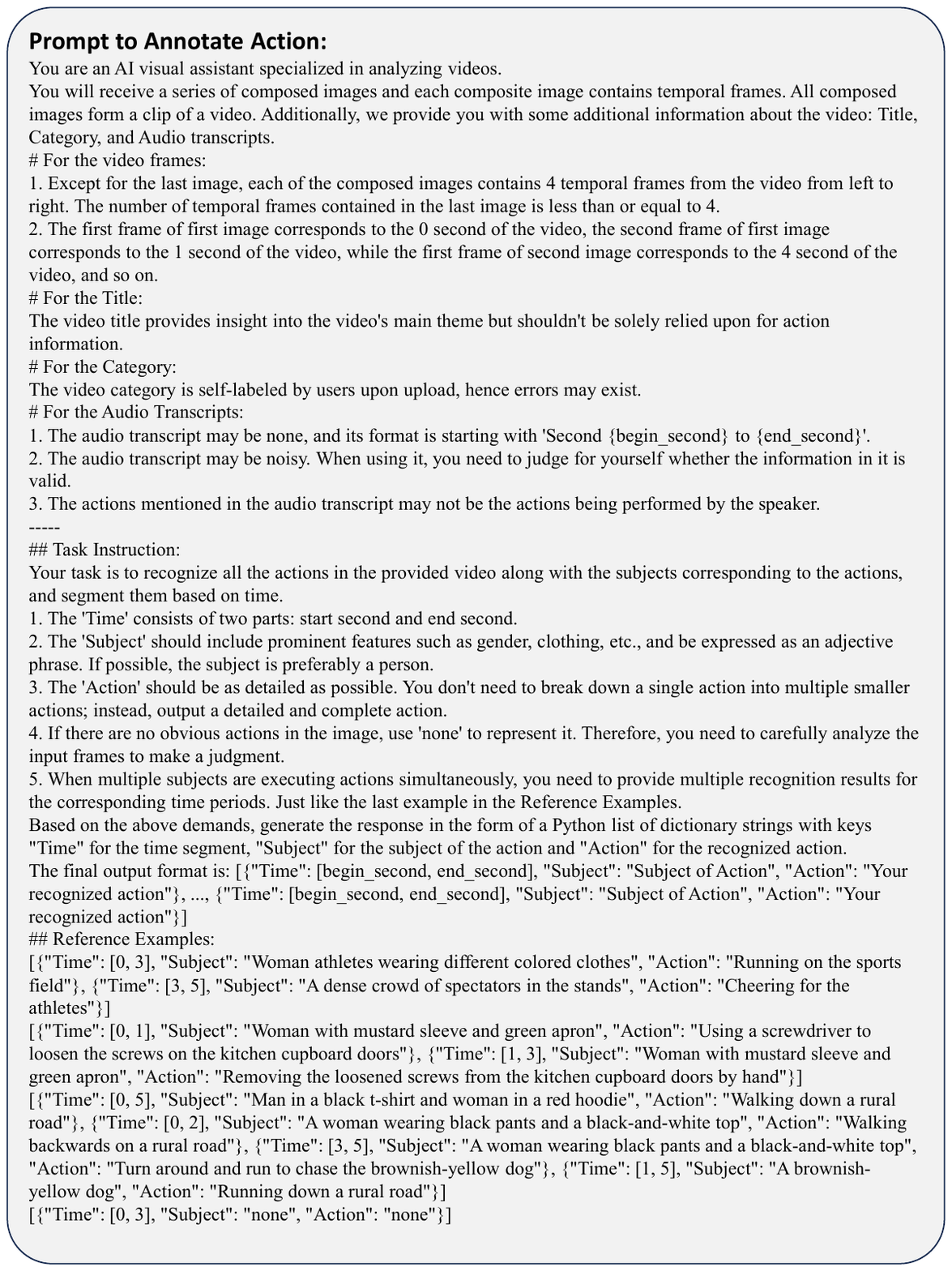}
    \caption{Prompts of annotating action of video clips.}
    \label{fig:prompt_action_annotate}
\end{figure}

\begin{figure}
    \includegraphics[width=1.1\textwidth]{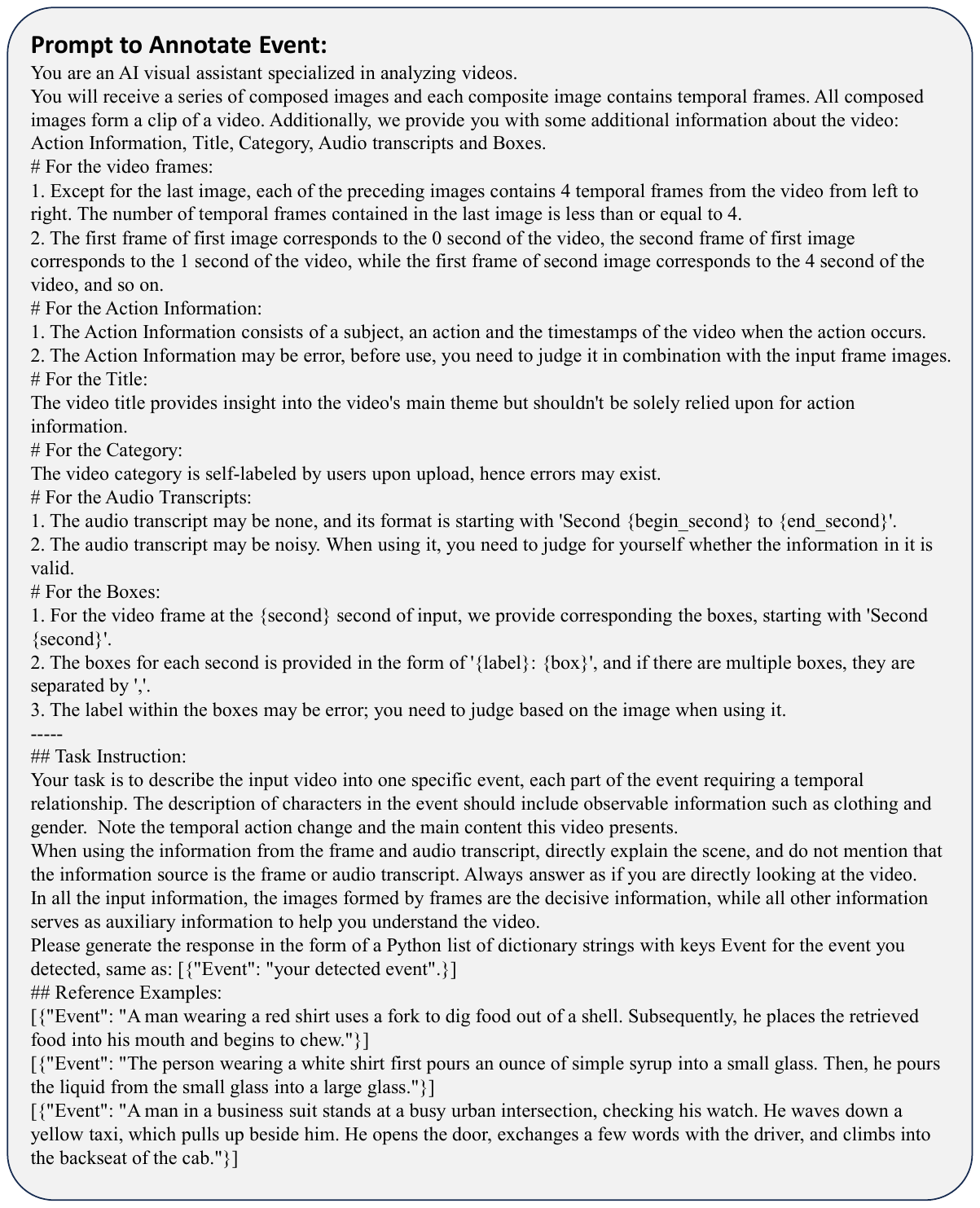}
    \caption{Prompts about the annotating event of video clips.}
    \label{fig:prompt_event_annotate}
\end{figure}

\begin{figure}
    \includegraphics[width=1.1\textwidth]{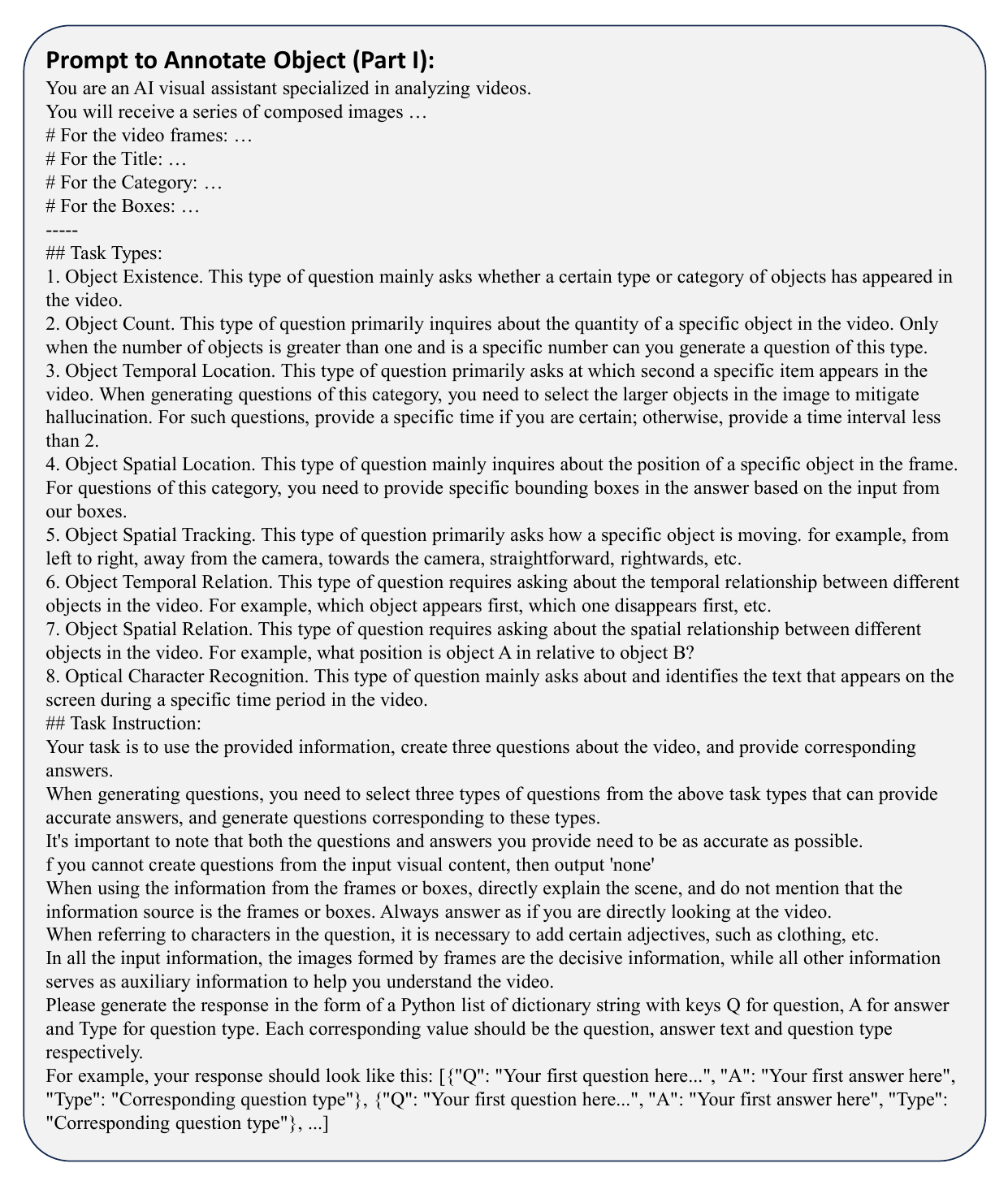}
    \caption{Prompts about the annotating objects of video clips. The description of the input information is partially omitted here because it is the same as the previous prompt.}
    \label{fig:prompt_object_annotate}
\end{figure}

\begin{figure}
    \includegraphics[width=1.1\textwidth]{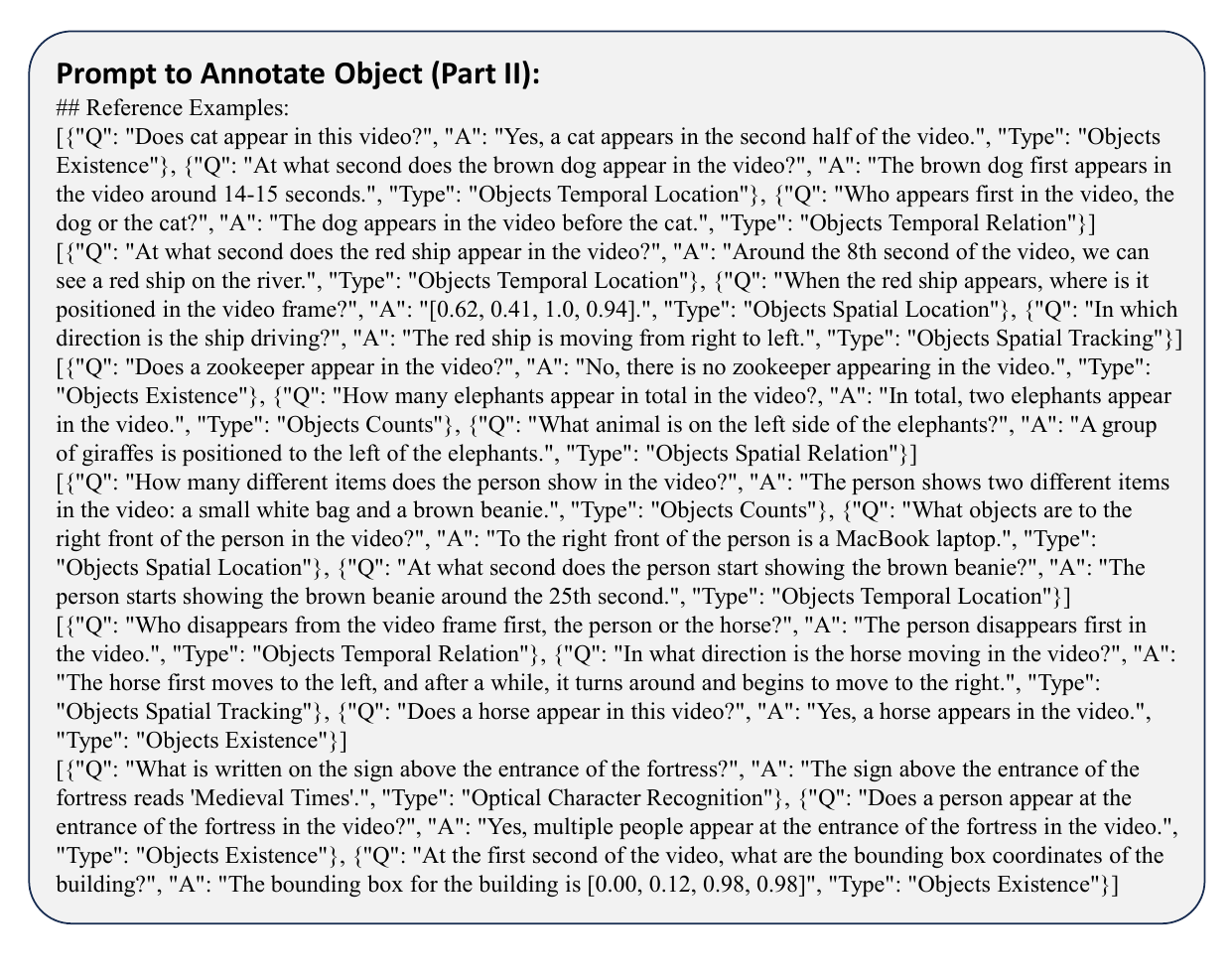}
    \caption{Prompts about the annotating objects of video clips.(continued)}
    \label{fig:prompt_object_annotate_continued}
\end{figure}

\begin{figure}
    \includegraphics[width=1.1\textwidth]{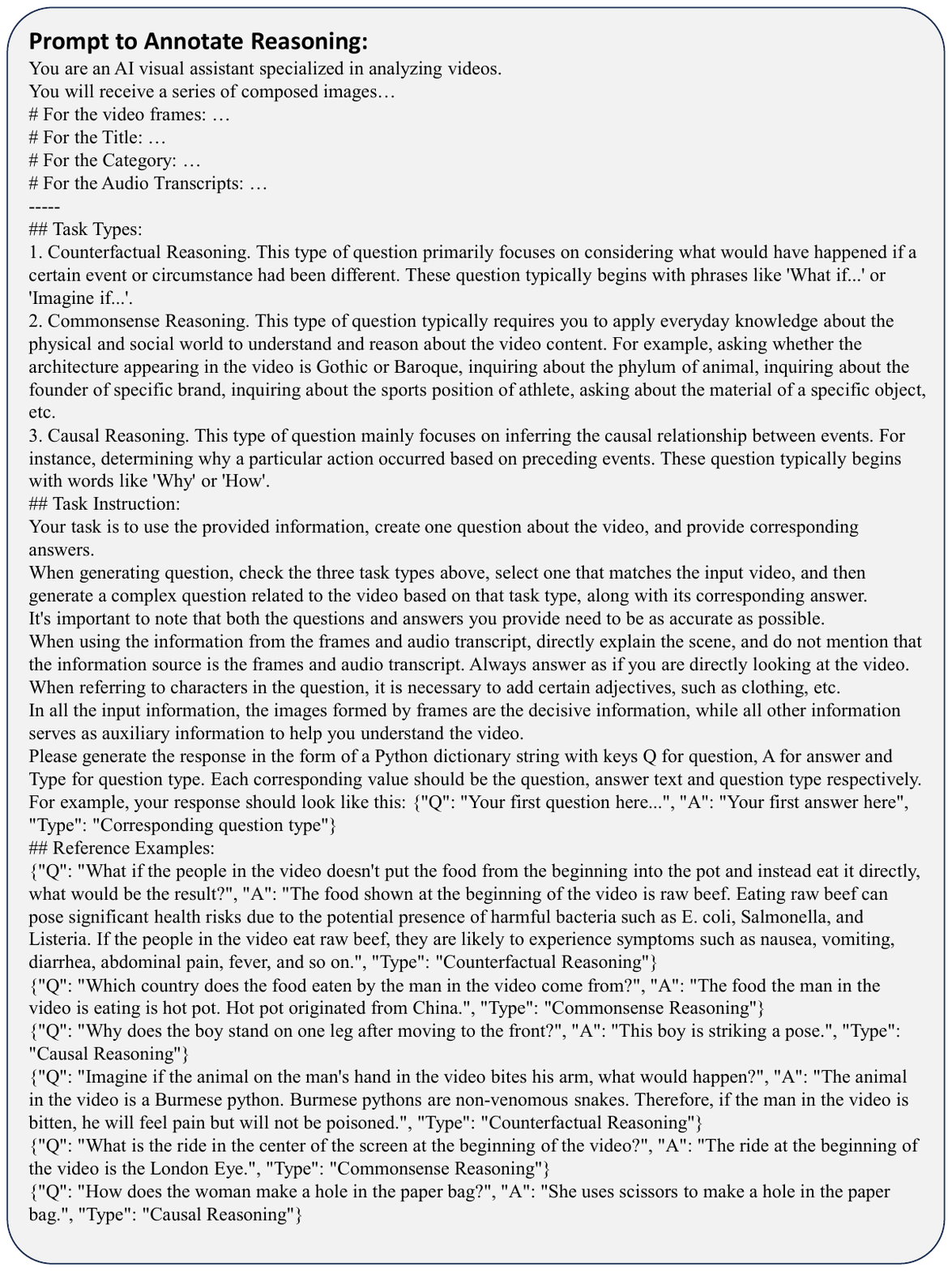}
    \caption{Prompts of annotating reasoning of video clips. The description of the input information is partially omitted here because it is the same as the previous prompt.}
    \label{fig:prompt_reasoning_annotate}
\end{figure}

\begin{figure}
    \includegraphics[width=1.1\textwidth]{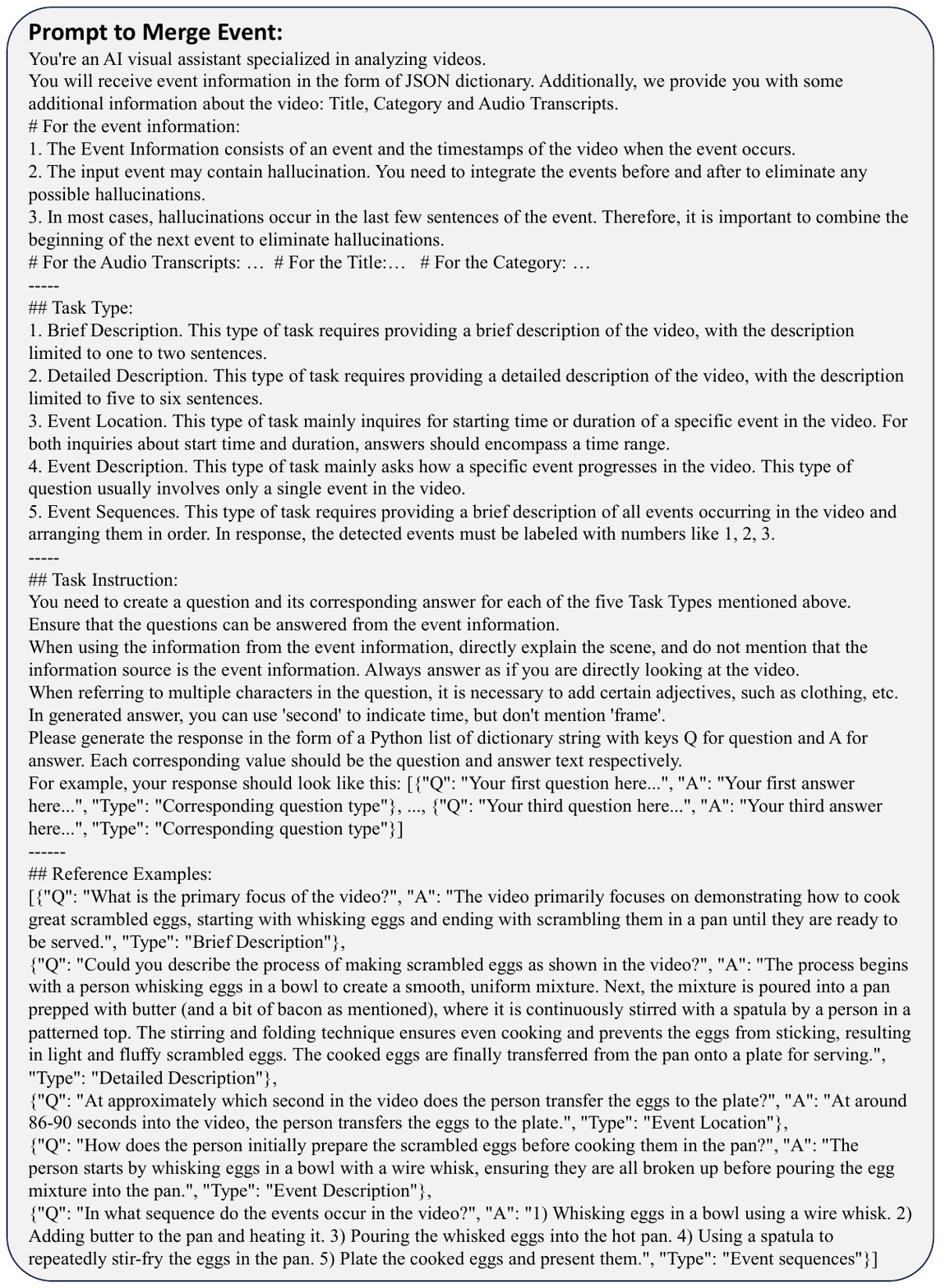}
    \caption{Prompts of merging event annotation of video clips.}
    \label{fig:prompt_event_merge}
\end{figure}

\begin{figure}
    \includegraphics[width=1.1\textwidth]{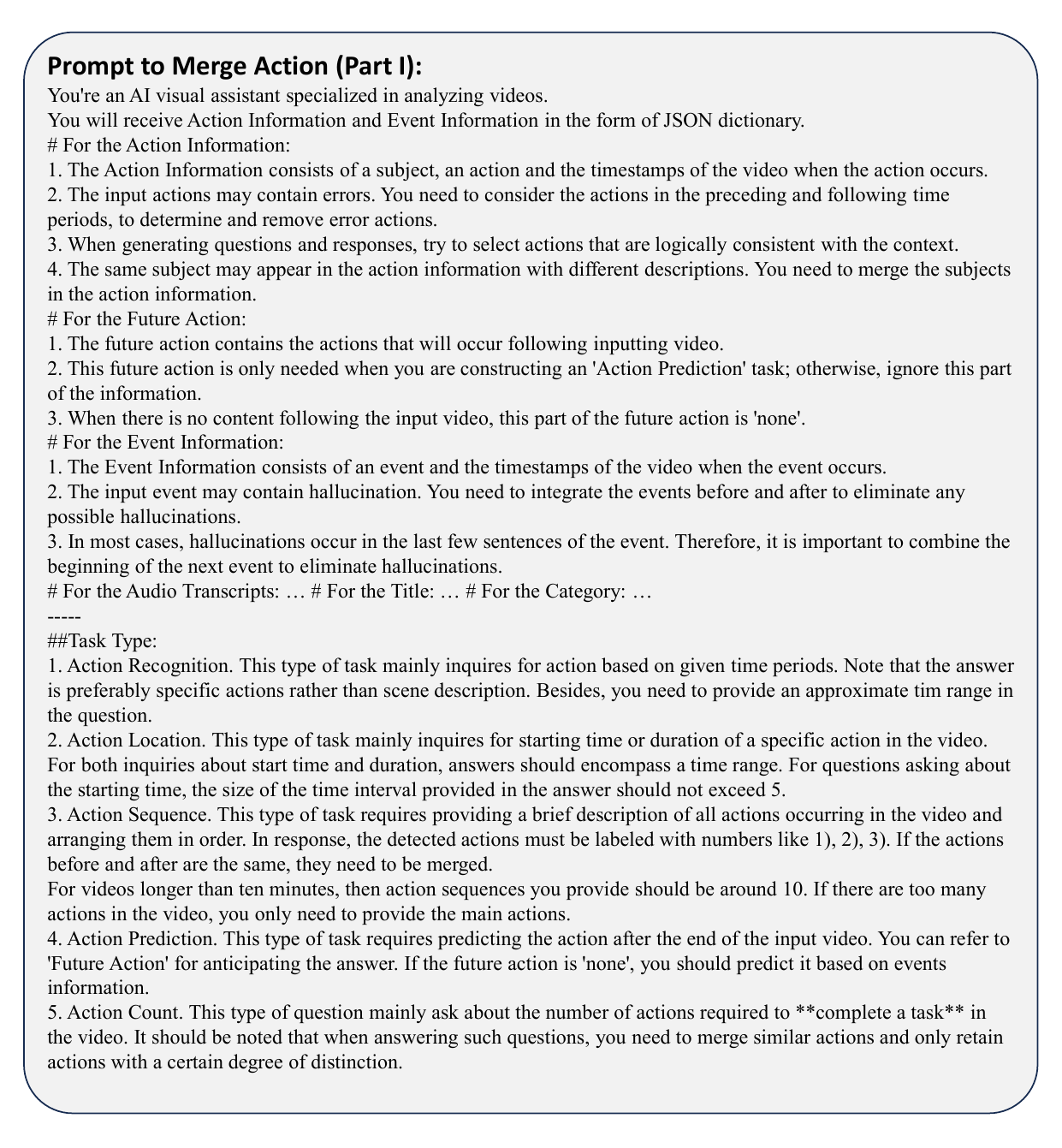}
    \caption{Prompts of merging action annotation of video clips.}
    \label{fig:prompt_action_merge}
\end{figure}

\begin{figure}
    \includegraphics[width=1.1\textwidth]{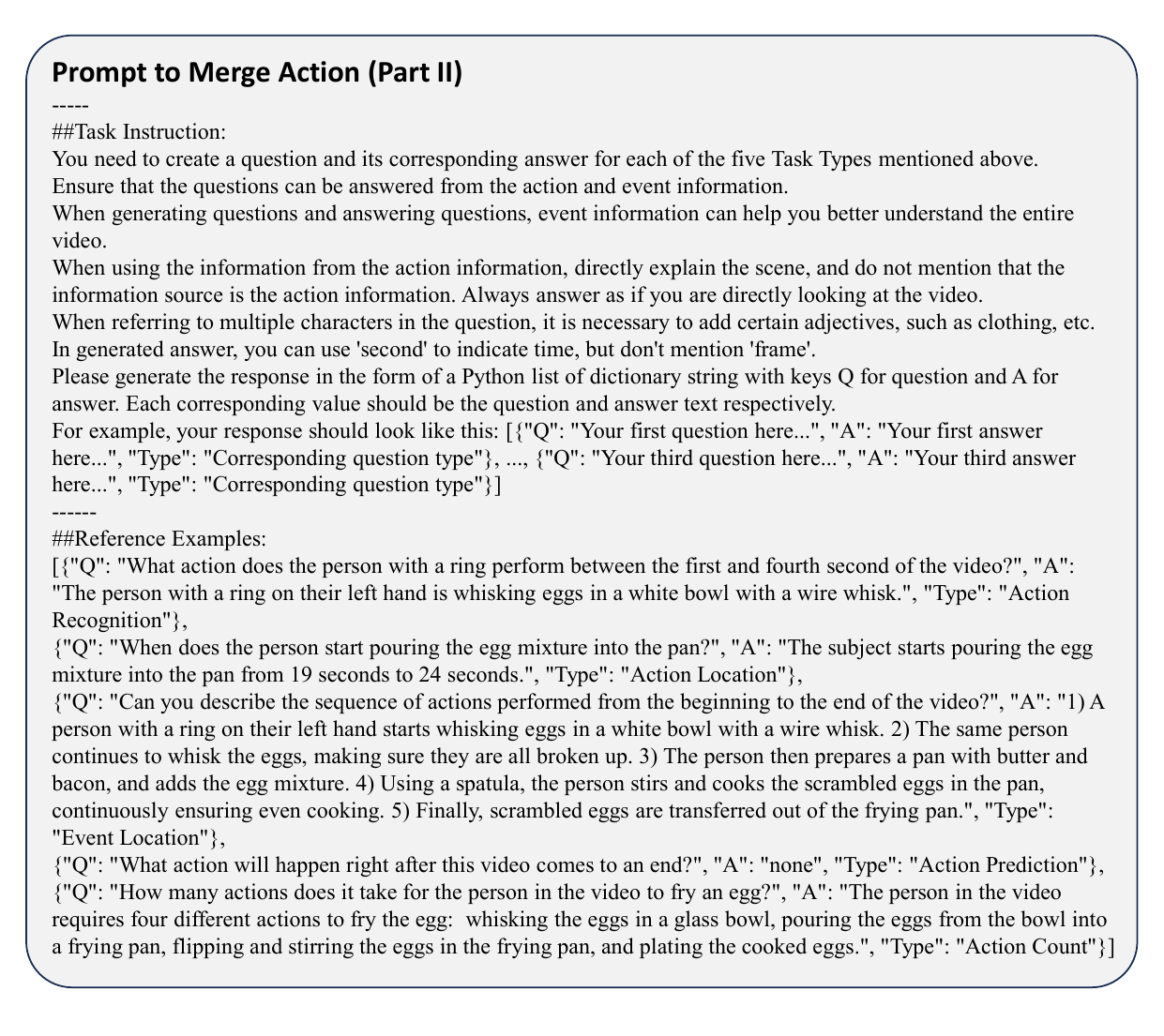}
    \caption{Prompts of merging action annotation of video clips(continued).}
    \label{fig:prompt_action_merge_continued}
\end{figure}

\begin{figure}
    \includegraphics[width=1.1\textwidth]{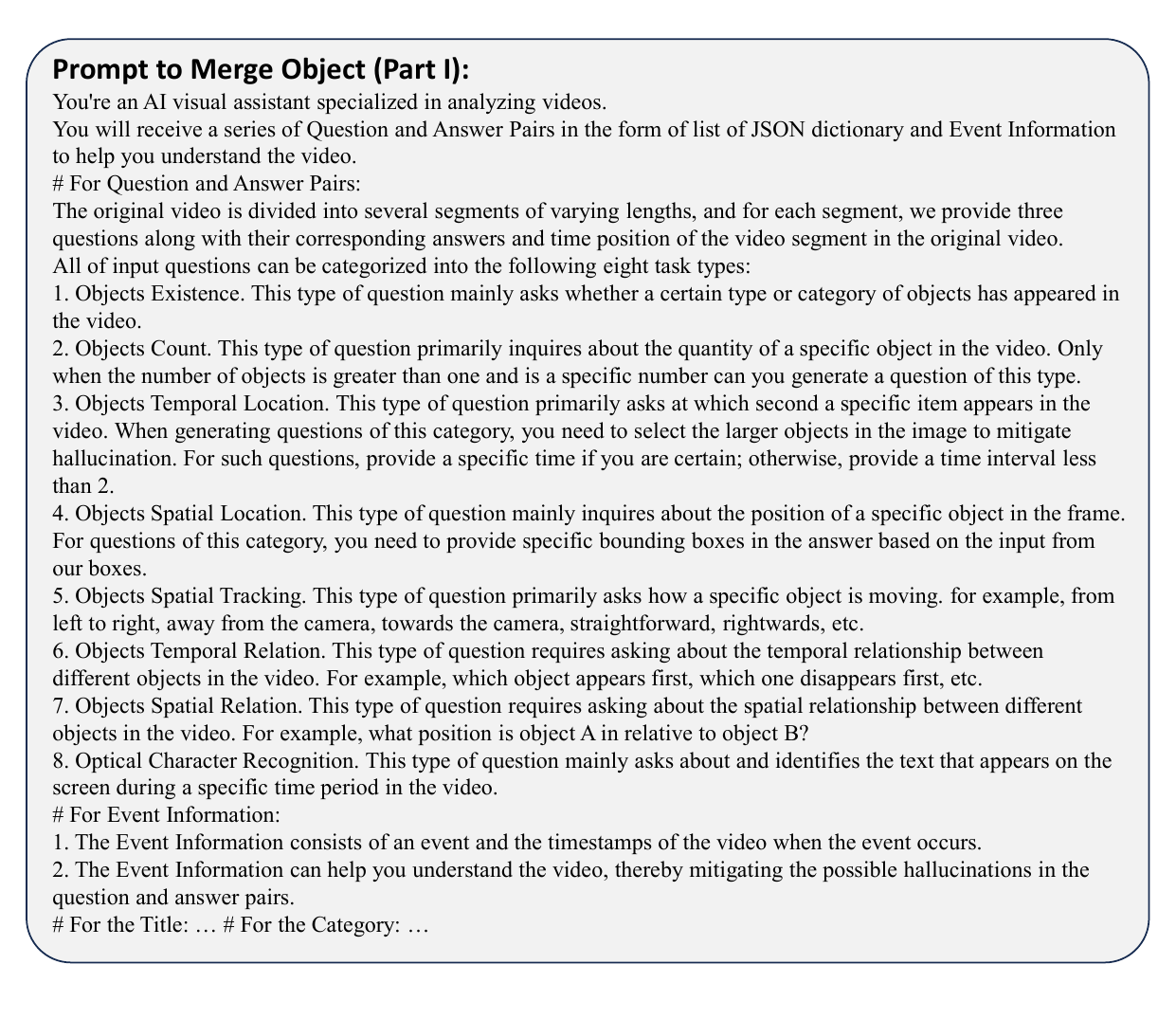}
    \caption{Prompts of merging object annotation of video clips.}
    \label{fig:prompt_object_merge}
\end{figure}

\begin{figure}
    \includegraphics[width=1.1\textwidth]{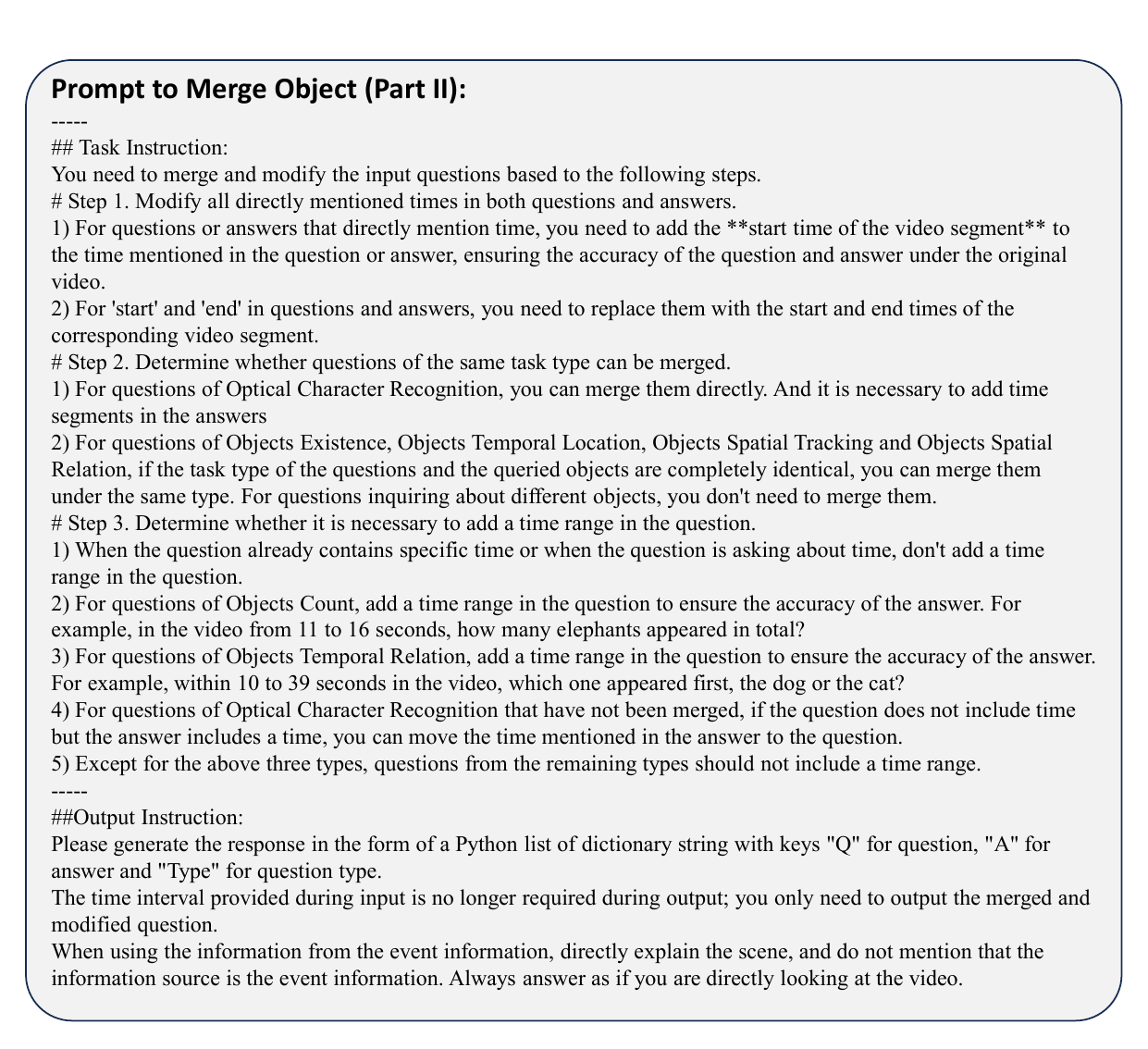}
    \caption{Prompts of merging object annotation of video clips.}
    \label{fig:prompt_object_merge_continued}
\end{figure}

\begin{figure}
    \includegraphics[width=1.1\textwidth]{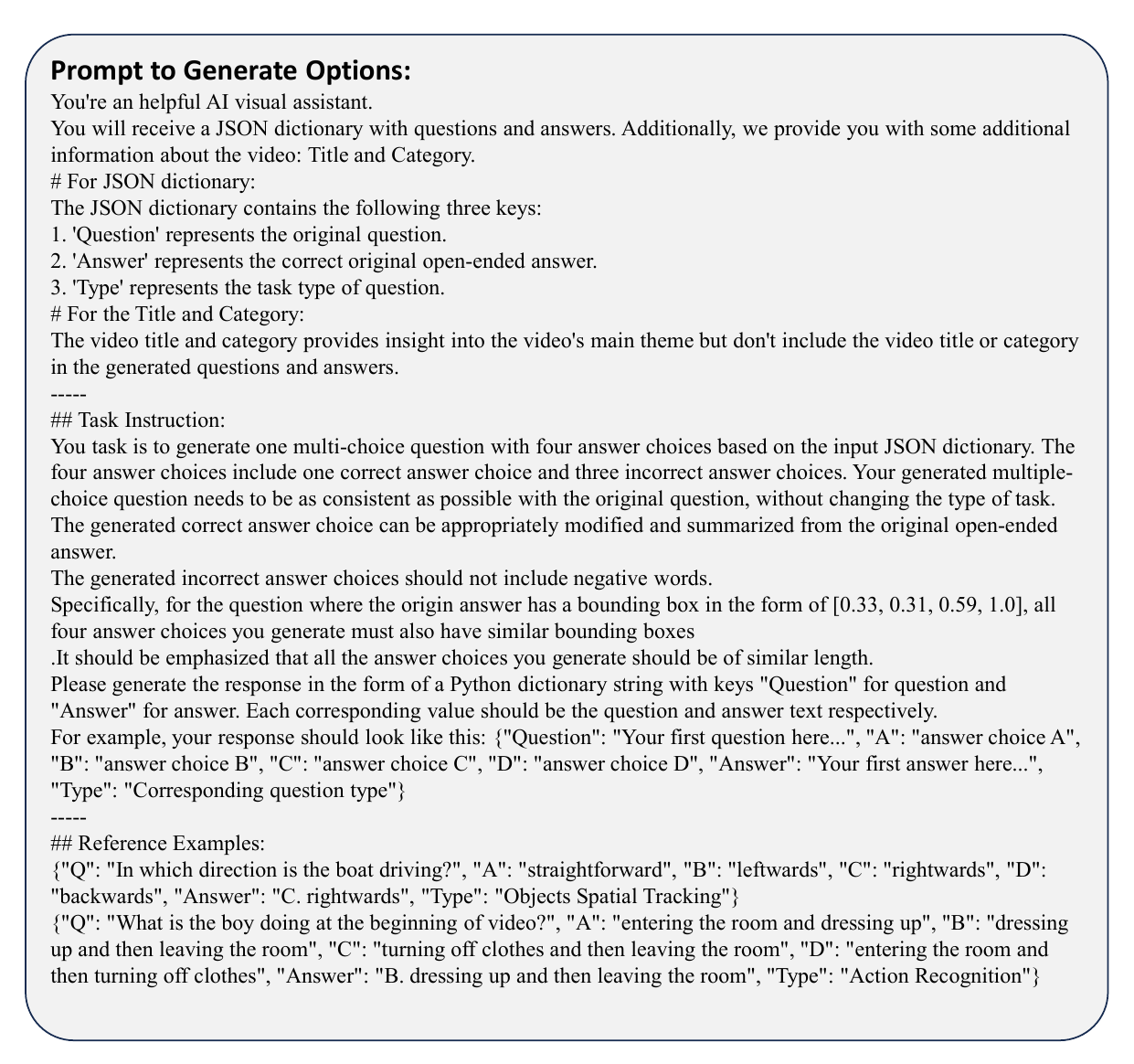}
    \caption{Prompts of generating multi-choices question answer pairs for open-ended question answer pairs.}
    \label{fig:prompt_option_generation}
\end{figure}

\paragraph{Detailed Prompts}
In this section, we present all the prompts we used to construct question-answer pairs based on video clips. We first annotate the video clips from different aspects in the order of Figure~\ref{fig:prompt_action_annotate}, ~\ref{fig:prompt_event_annotate}, ~\ref{fig:prompt_object_annotate}, \ref{fig:prompt_object_annotate_continued}, and~\ref{fig:prompt_reasoning_annotate}. Regarding the specific input, we convert the concatenated image into base64 format, while for textual information, it is input in the form of "Video Title: \{title\}$\backslash$nVideo Category: \{category\}$\backslash$n," where \{title\} and \{category\} represent the actual title and category of the video, respectively. Then, we use Figure~\ref{fig:prompt_event_merge},~\ref{fig:prompt_action_merge}, \ref{fig:prompt_action_merge_continued} ~\ref{fig:prompt_object_merge}, and \ref{fig:prompt_object_merge_continued} to transform the annotations of the video clips into annotations of longer merged videos and construct the corresponding question-answer pairs. Finally, we use Figure~\ref{fig:prompt_option_generation} to convert the previously constructed open-ended question-answer pairs into multiple-choice question-answer pairs.
\subsection{Evaluation Description}
\label{model_descripton}

To streamline the evaluation of models like GPT-4o and Gemini-1.5-Flash, which necessitate API calls, a video is submitted alongside multiple questions, with a cap of eight questions at once. For GPT-4o, a specific methodology is applied to process videos for assessment. This involves sampling one frame per second for videos less than 100 seconds long, and for videos exceeding this duration, a uniform sampling of 100 frames is employed. When inputting frame images into the GPT4-o, we use the same method of composed images as during annotation. We concatenate four adjacent frame images horizontally into one image, thus controlling the maximum number of images input into GPT-4o to be within 25.
In contrast, for the Gemini-1.5-Flash model, the original, full-length video is utilized directly in evaluations, accommodating its capability to handle extensive multi-modal contexts.

Notably, the evaluation of the Relation Reasoning task is somewhat special, as this task involves an additional image input or video input besides the regular video input. Most existing open-source Video-LLMs do not support video-image joint input or multi-video input. Therefore, we only conducted related evaluations on GPT-4o and Gemini-1.5-Flash. When evaluating the Image-Video task of GPT-4o, we place the single image at the front of all composed images. For the Video-Video evaluation, we specify in the system message up to which frame image corresponds to video 1. The API interface of Gemini already supports mixed image and video input or multiple video inputs. We have pre-implemented declarations of the input information in the text.

\subsection{Result Reproduction}

\begin{table}[t]
\centering
\caption{\textbf{Evaluation results on VideoVista sorted by Category.} These categories come from YouTube's classification of videos. Autos \& Vehicles (\textbf{A\&E}), Comedy (\textbf{Com.}), Education (\textbf{Edu.}), Entertainment (\textbf{ENT.}), Film \& Animation (\textbf{F\&A}), Gaming (\textbf{Gam.}), Howto \& Style (\textbf{H\&S}), Music (\textbf{Mus.}), News \& Politics (\textbf{N\&P}), People \& Blogs (\textbf{P\&B}), Pets \& Animals (\textbf{P\&A}), Science \& Technology (\textbf{S\&T}), Sports (\textbf{Spo.}), Travel \& Events (\textbf{T\&E}).}
\vspace{3pt}
\scalebox{0.7}{
\begin{tabular}{l rrrrrrrrrrrrrrr}
\toprule
\textbf{Model} & \textbf{A\&V} & \textbf{Com.} & \textbf{Edu.} & \textbf{Ent.} & \textbf{F\&A} & \textbf{Gam.} & \textbf{H\&S} & \textbf{Mus.} & \textbf{N\&P} & \textbf{P\&B} & \textbf{P\&A} & \textbf{S\&T} & \textbf{Spo.} & \textbf{T\&E} \\
\midrule
Video-LLaMA & 26.7 & 26.0 & 24.9 & 24.2 & 25.0 & 23.9 & 24.3 & 26.8 & 24.8 & 26.3 & 26.1 & 26.0 & 25.4 & 26.4  \\
Video-ChatGPT & 36.5 & 34.8 & 38.4 & 36.1 & 34.4 & 36.0 & 36.1 & 48.8 & 38.4 & 37.7 & 35.6 & 39.3 & 38.3 & 36.9\\
IVA & 38.7 & 38.4 & 39.7 & 40.9 & 35.0 & 41.4 & 37.0 & 46.3 & 41.6 & 40.3 & 41.2 & 39.5 & 46.0 & 45.2 \\
Video-LLaVA & 52.9 & 57.5 & 54.9 & 58.9 & 52.5 & 53.8 & 53.5 & 56.1 & 59.4 & 56.6 & 62.7 & 58.8 & 60.6 & 63.6 \\
 LLaMA-VID & 54.4 & 58.8 & 55.5 & 57.6 & 52.5 & 57.3 & 52.8 & 58.5 & 59.4 & 58.6 & 60.3 & 58.4 & 61.6 & 63.8\\
LLaVA-NeXT-Video & 55.6 & 58.3 & 54.3 & 58.5 & 54.0 & 59.5 & 53.0 & 51.2 & 61.3 & 56.1 & 60.9 & 56.5 & 60.7 & 60.4 \\
VideoChat2-Vicuna & 50.5 & 53.8 & 52.7 & 57.4 & 48.7 & 49.4 & 53.1 & 53.7 & 57.3 & 53.5 & 55.7 & 57.8 & 56.1 & 58.0 \\
VideoChat2-Mistral & 53.6 & 59.7 & 55.4 & 61.7 & 49.9 & 50.9 & 58.0 & 63.4 & 63.5 & 58.9 & 63.7 & 60.6 & 60.4 & 59.2 \\
\midrule
Gemini-1.5-Flash & 74.2 & 75.5 & 75.4 & \textbf{80.8} & 68.3 & 76.5 & 77.7 & \textbf{83.8} & 81.1 & 77.6 & 79.2 & 78.1 & 79.3 & 79.8 \\
GPT-4o & \textbf{75.6} & \textbf{77.5} & \textbf{76.3} & 80.1 & \textbf{71.1} & \textbf{77.5} & \textbf{79.4} & 72.1 & \textbf{83.3} & \textbf{80.2} & \textbf{80.1} & \textbf{79.0} & \textbf{81.4} & \textbf{81.8} \\
\bottomrule
\end{tabular}
}
\label{tab:video_result_category}
\end{table}

\begin{table}[t]
\centering
\caption{\textbf{Evaluation results on VideoVista sorted by Duration.} The videos were divided into five duration ranges: 0-60s (\textbf{Tiny}), 60-120s (\textbf{Short}), 120-300s (\textbf{Med.}), 300-600s (\textbf{Long}), and 600s+ (\textbf{\textbf{Huge}}). The evaluation scores for the Video Understanding tasks were then calculated for videos within each duration range.}
\vspace{3pt}
\scalebox{0.83}{
\begin{tabular}{l r r r r r}
\toprule
\textbf{Model}  & \textbf{Tiny} & \textbf{Short} & \textbf{Med.} & \textbf{Long} & \textbf{Huge} \\
\midrule
Video-LLaMA & 25.1 & 25.9 & 26.2 & 23.3 & 21.7\\
Video-ChatGPT & 35.6 & 36.4 & 35.6 & 38.6 & 35.4\\
IVA  & 37.5 & 36.6 & 36.3 & 40.5 & 45.6\\
Video-LLaVA & 58.0 & 52.9 & 51.0 & 53.9 & 54.3\\ 
LLaMA-VID & 57.8 & 53.5 & 50.7 & 55.0 & 54.6 \\
LLaVA-NeXT-Video & 59.8 & 52.7 & 50.9 & 52.6 & 53.8\\
VideoChat2-Vicuna & 57.1 & 51.5 & 47.5 & 50.6 & 51.2\\
VideoChat2-Mistral & 58.0 & 52.9 & 51.0 & 53.9 & 54.3 \\
\midrule
Gemini-1.5-Flash & 84.1 & 72.8 & 69.1 & 71.1 & 68.4 \\
GPT-4o & 83.5 & 73.2 & 70.9 & 70.9 & 73.1 \\
\bottomrule
\label{tab:video_result_duration} 
\end{tabular}}
\end{table}

\subsubsection{Comparative Models}

\textbf{Video-LLaMA}~\cite{zhang2023videollama} leverages Vision Encoder and Audio Encoder to process video independently. The Vision Encoder uses 16 video frames as input.

\textbf{Video-ChatGPT}~\cite{Maaz2023VideoChatGPT} utilize average pooling to generate temporal and spatial features. It uniformly samples 100 frames from the video and obtains frame-level embeddings. These embeddings undergo average pooling to generate temporal and spatial features. The features are concatenated to form video-level features, which are then input into the LLM.

\textbf{Video-LLaVA}~\cite{lin2023video} binds visual signals to the language feature space, unifying visual representations, and proposes a solution to align before projection.  When processing input videos, it uses 8 frames uniformly sampled from the video.

\textbf{LLaMA-VID}~\cite{li2023llamavid} representing each frame with two distinct tokens, namely context token and content token. The context token encodes the overall image context based on user input, whereas the content token encapsulates visual cues in each frame. As for video input, LLaMA-VID extracts frames at a speed of 1 FPS.

\textbf{IVA}~\cite{li2024llms} present an interactive visual adapter within LLMs, designed to enhance interaction with fine-grained visual elements. The IVA model uses a high-frequency sampling method of five frames per second for shorter videos, while it samples 200 frames from the video for longer videos.

\textbf{LLaVA-NeXT-Video}~\cite{liu2024llavanext} employs the same model architecture as LLaVA\cite{liu2023visual}, effectively reduces the feature dimension of a single image representation through spatial pooling.

\textbf{VideoChat2}~\cite{li2023mvbench} effectively compresses the visual features using the UMT-L visual encoder and Qformer.
It also samples 16 frames from each video as input.

\textbf{Gemini-1.5-Flash} is the newest Gemini model published by Google. It has a default context window of up to one million tokens and has demonstrated strong performance in long videos.

\textbf{GPT-4o} is the latest GPT model published by OpenAI. It achieves GPT-4 Turbo-level performance on text, reasoning, and coding intelligence while setting new high watermarks on multilingual, audio, and vision capabilities.

\subsubsection{Detailed Experimental results}

We further present the tables of experimental results about model performances divided by Duration and Category in Tables~\ref{tab:video_result_category} and \ref{tab:video_result_duration}.

\subsection{Case Data}
\label{task_stat}
\begin{figure}
    \includegraphics[width=1\textwidth]{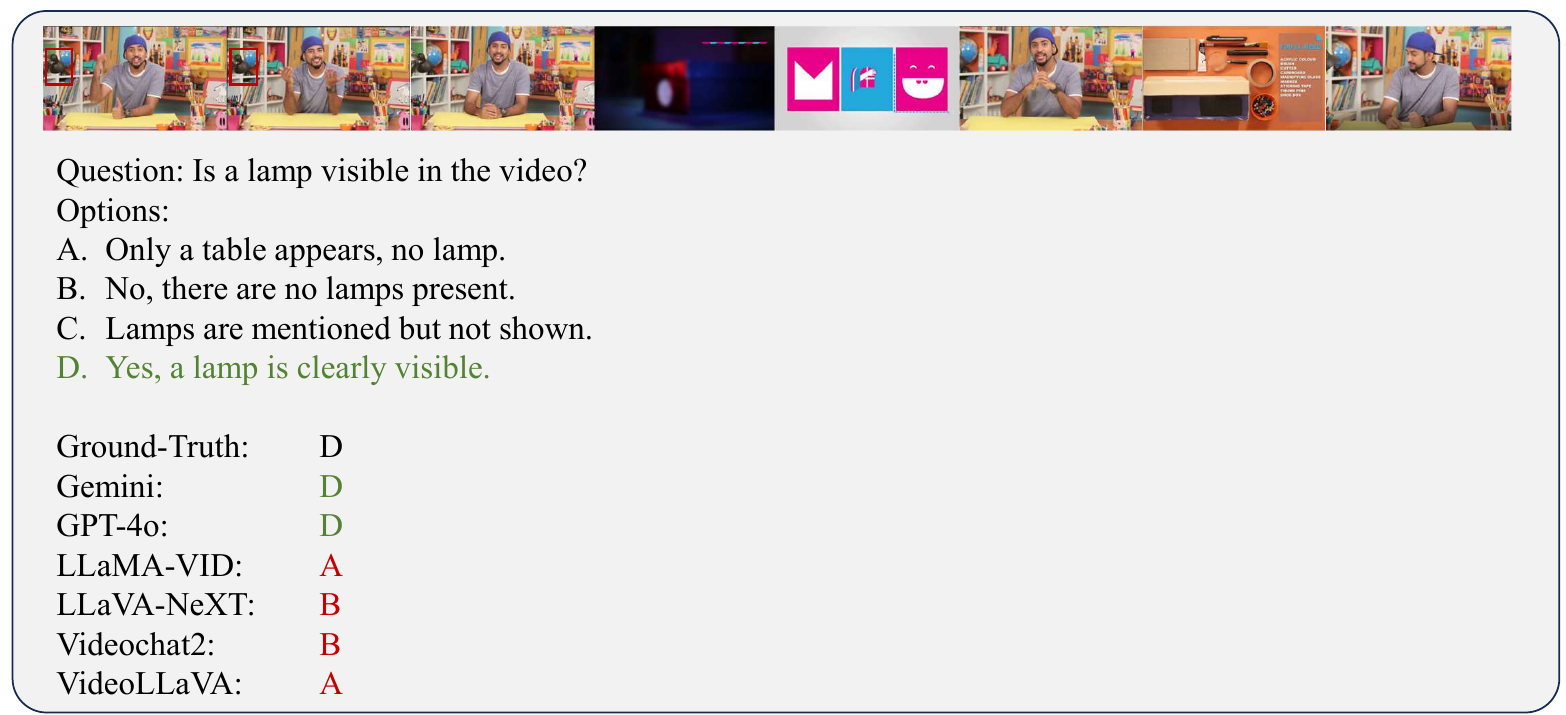}
    \caption{An Example of Video-LLMs responses and evaluation results of the Object Existence. We use \textcolor{forestgreen}{Green} to indicate correct and \textcolor{red}{Red} to indicate incorrect.}
    \label{fig:example_oe}
\end{figure}

\begin{figure}
    \includegraphics[width=1\textwidth]{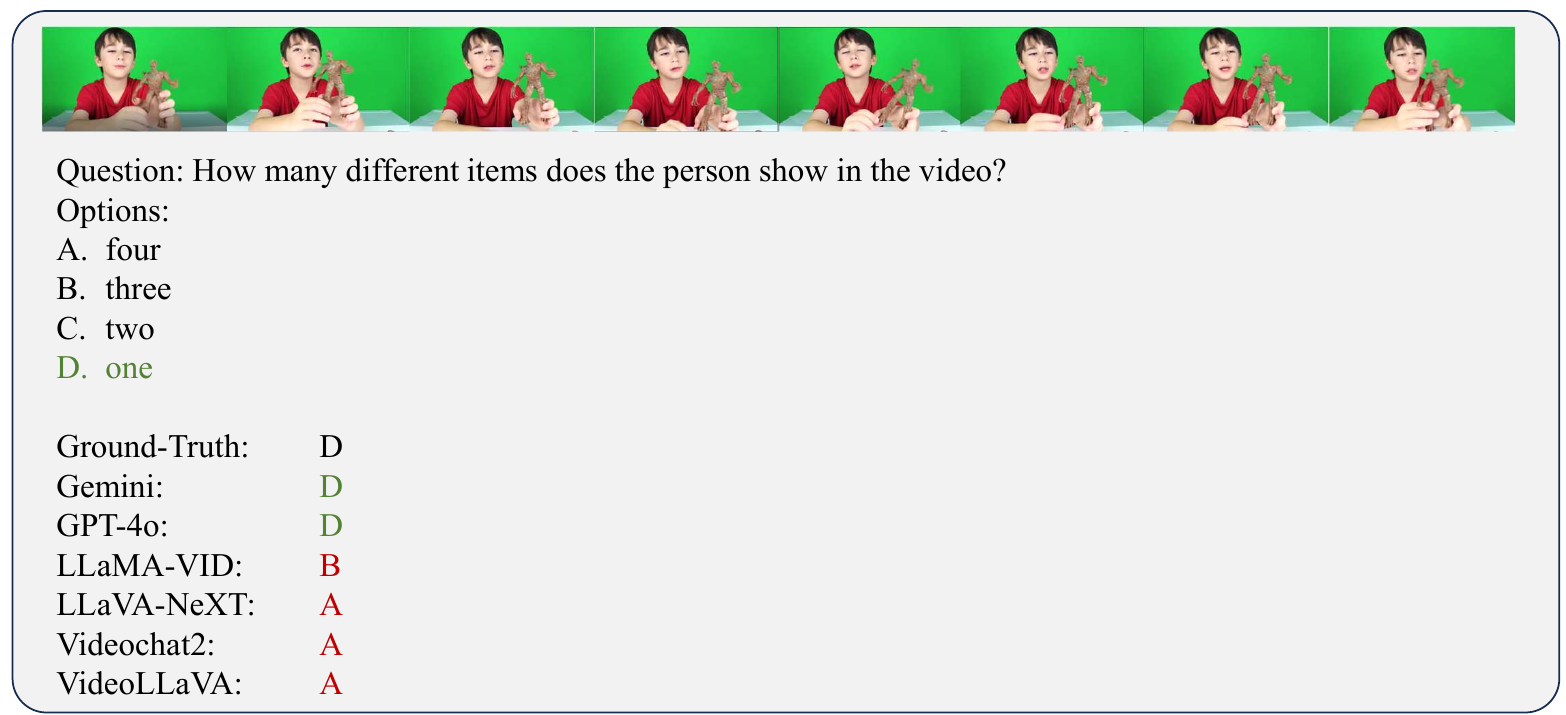}
    \caption{An Example of Video-LLMs responses and evaluation results of the Object Count. We use \textcolor{forestgreen}{Green} to indicate correct and \textcolor{red}{Red} to indicate incorrect.}
    \label{fig:example_oc}
\end{figure}

\begin{figure}
    \includegraphics[width=1\textwidth]{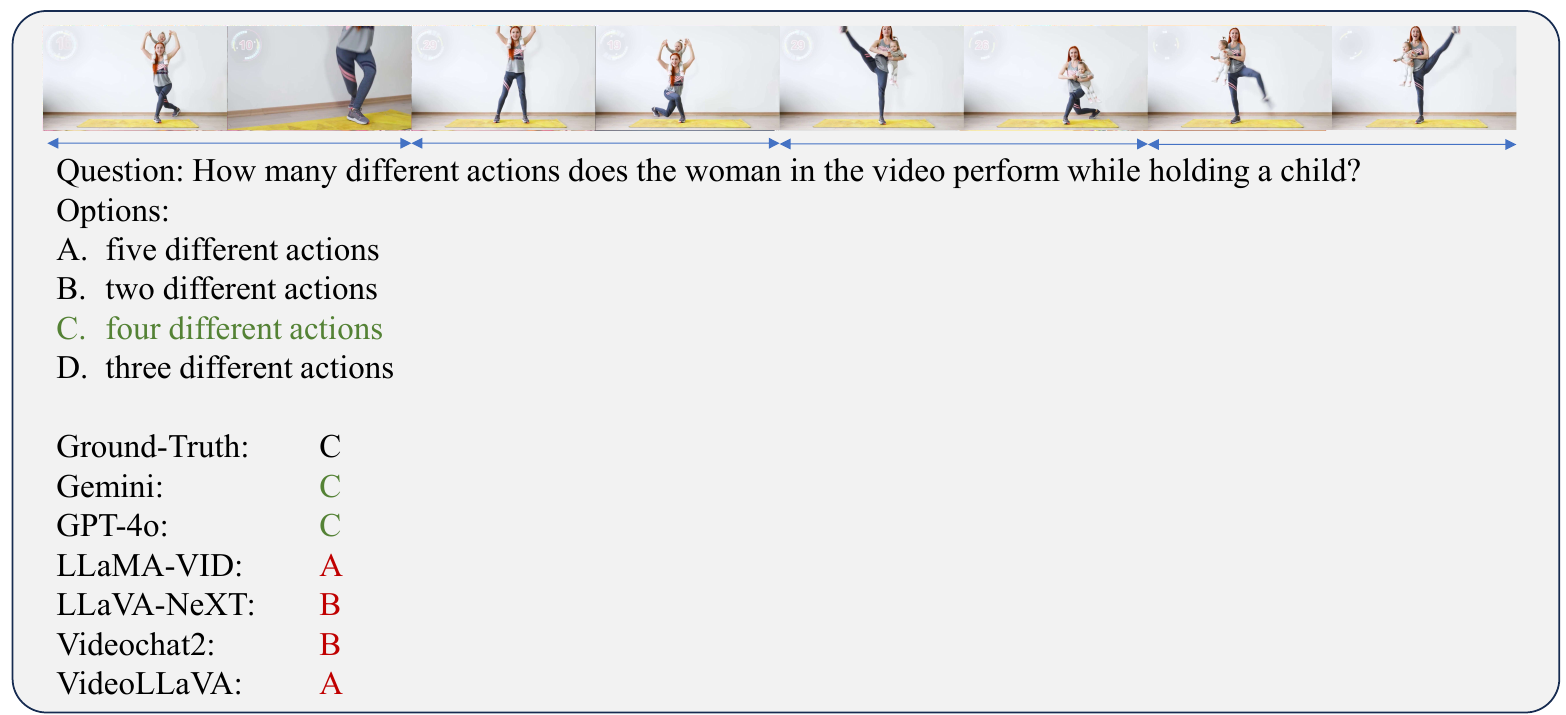}
    \caption{An Example of Video-LLMs responses and evaluation results of the Action Count. We use \textcolor{forestgreen}{Green} to indicate correct and \textcolor{red}{Red} to indicate incorrect.}
    \label{fig:example_ac}
\end{figure}

\begin{figure}
    \includegraphics[width=1\textwidth]{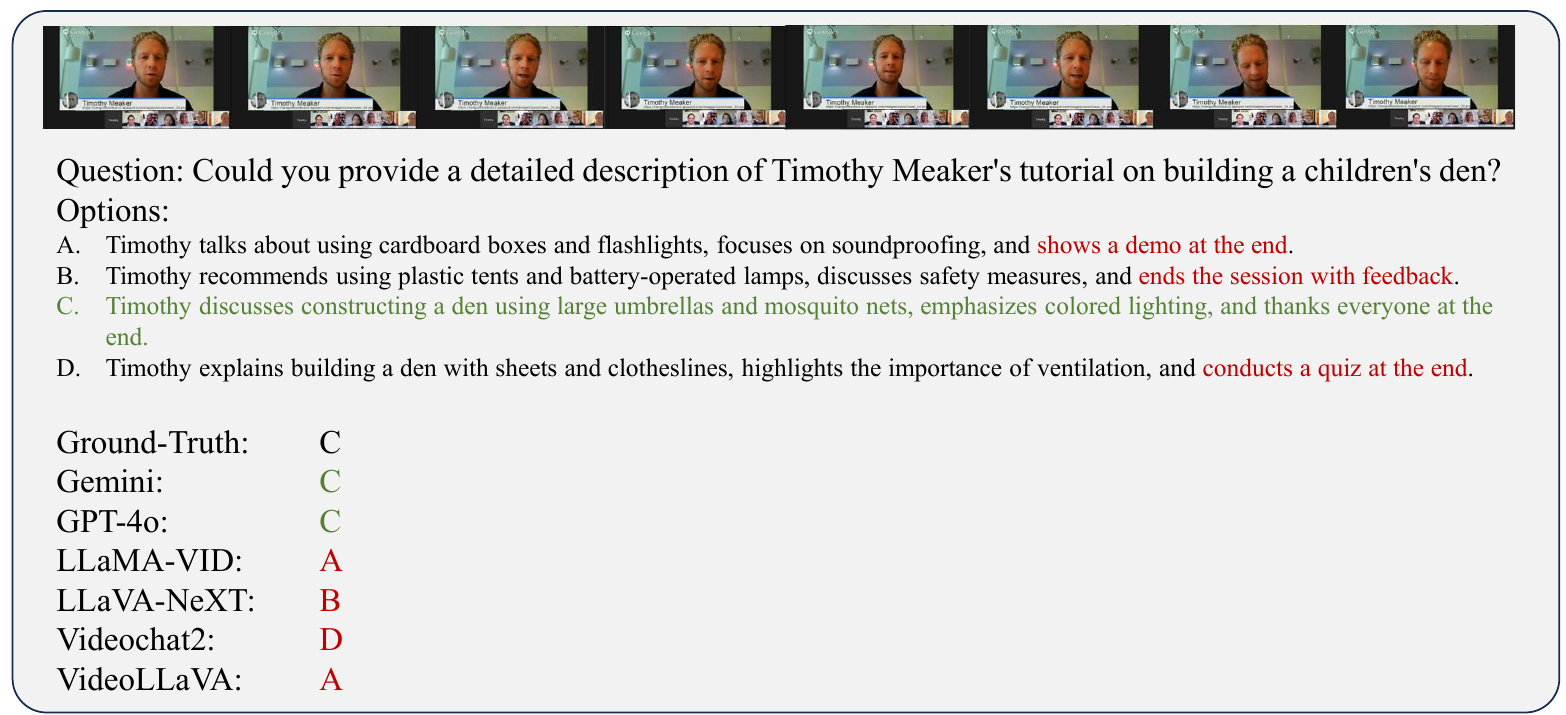}
    \caption{An Example of Video-LLMs responses and evaluation results of the Detailed Description. We use \textcolor{forestgreen}{Green} to indicate correct and \textcolor{red}{Red} to indicate incorrect.}
    \label{fig:example_dd}
\end{figure}

\begin{figure}
    \includegraphics[width=1\textwidth]{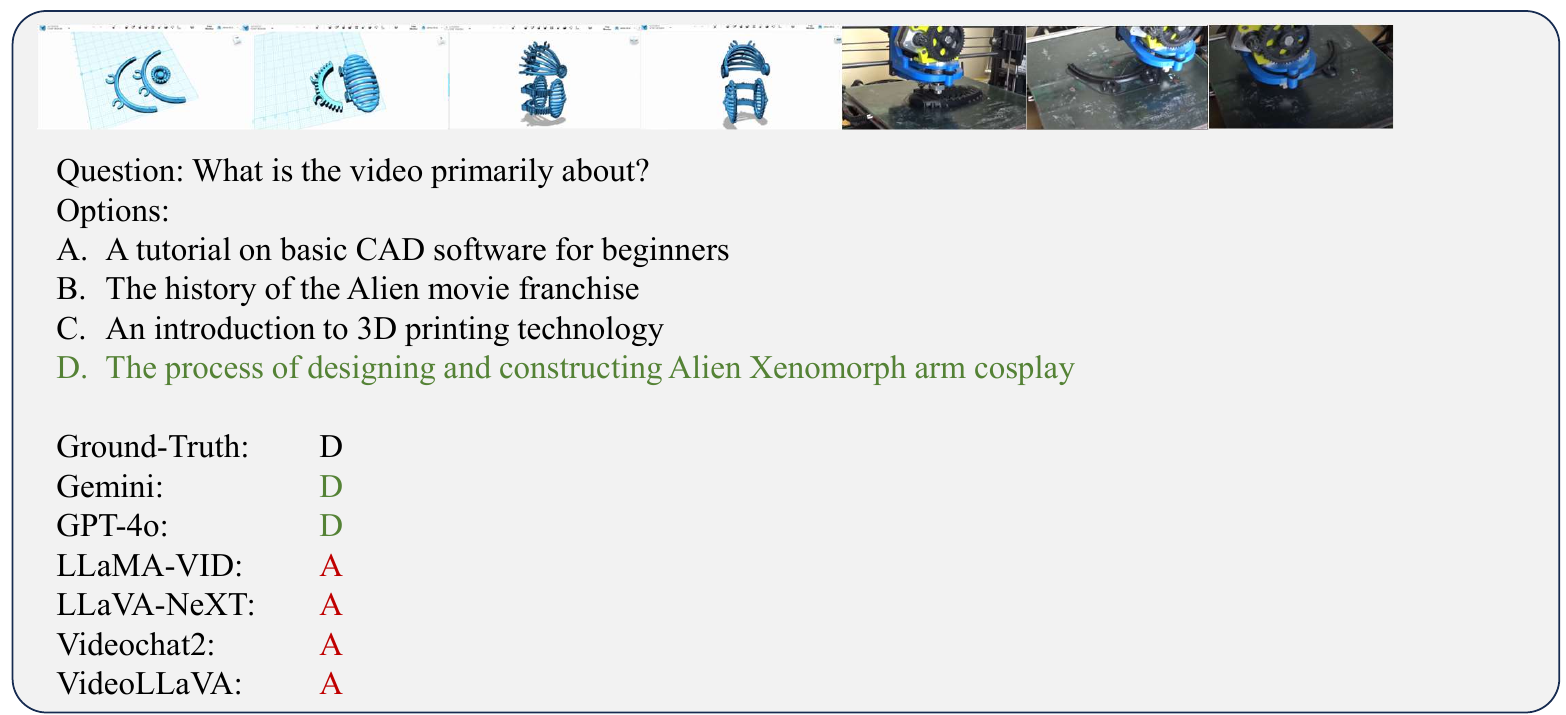}
    \caption{An Example of Video-LLMs responses and evaluation results of the Brief Description. We use \textcolor{forestgreen}{Green} to indicate correct and \textcolor{red}{Red} to indicate incorrect.}
    \label{fig:example_bd}
\end{figure}

\begin{figure}
    \includegraphics[width=1\textwidth]{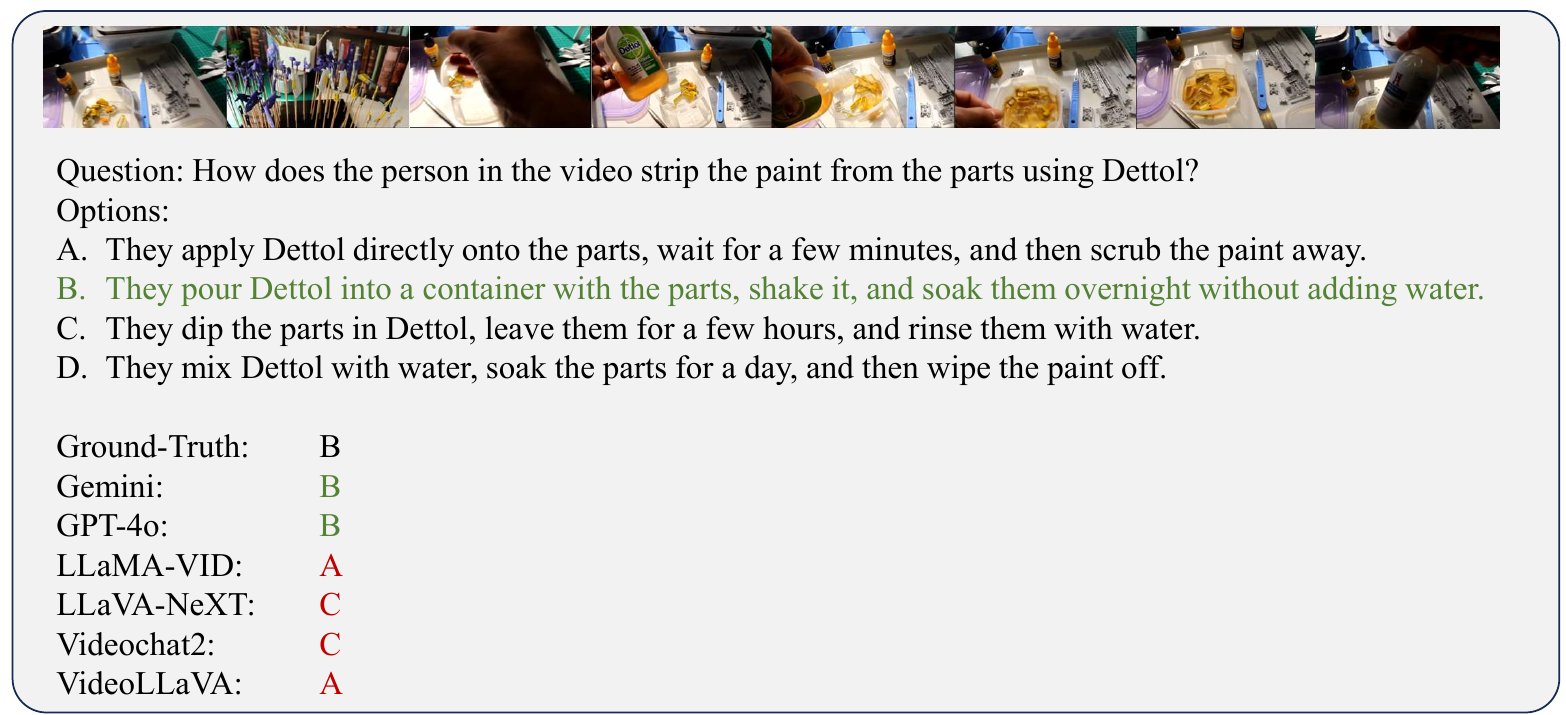}
    \caption{An Example of Video-LLMs responses and evaluation results of the Event Description. We use \textcolor{forestgreen}{Green} to indicate correct and \textcolor{red}{Red} to indicate incorrect.}
    \label{fig:example_ed}
\end{figure}

\begin{figure}
    \includegraphics[width=1\textwidth]{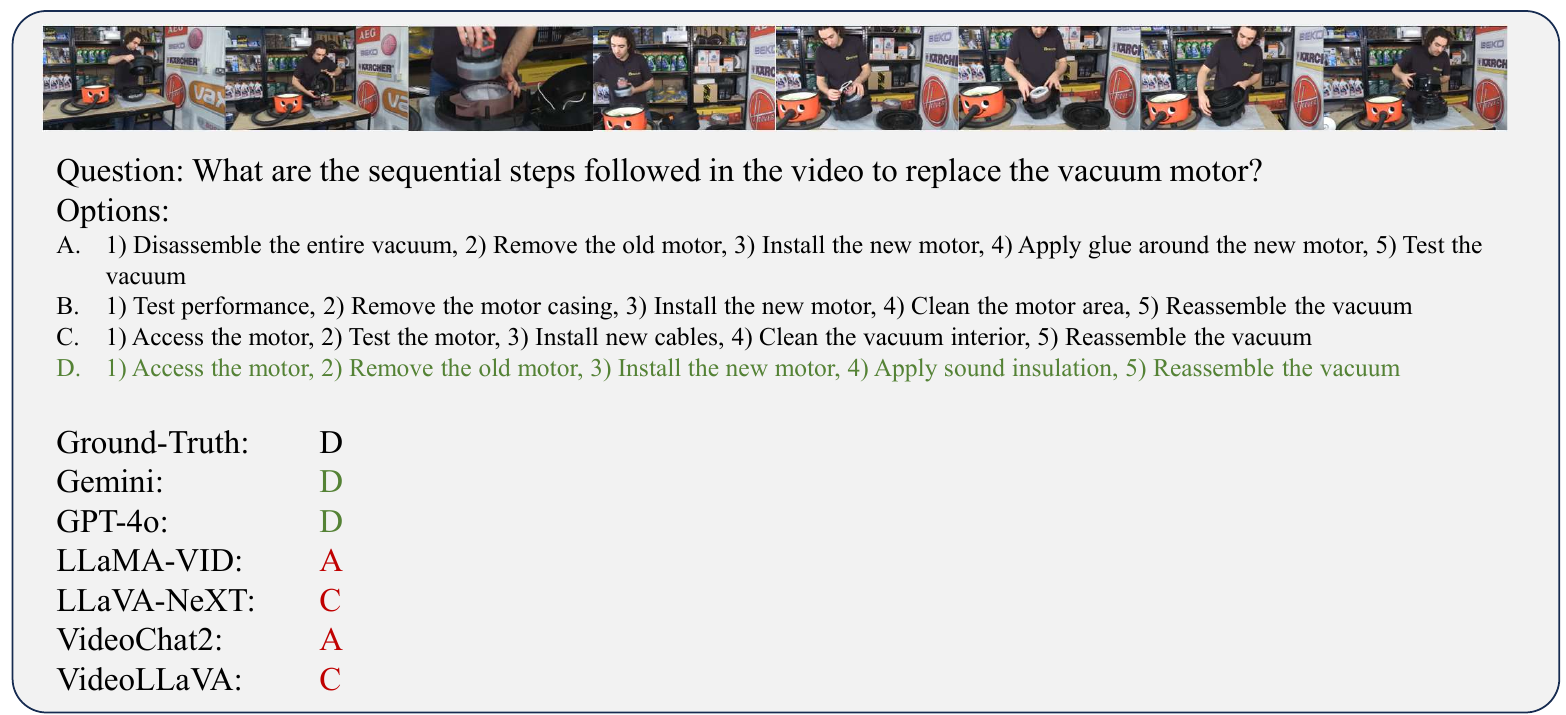}
    \caption{An Example of Video-LLMs responses and evaluation results of the Event Sequence. We use \textcolor{forestgreen}{Green} to indicate correct and \textcolor{red}{Red} to indicate incorrect.}
    \label{fig:example_es}
\end{figure}

\begin{figure}
    \includegraphics[width=1\textwidth]{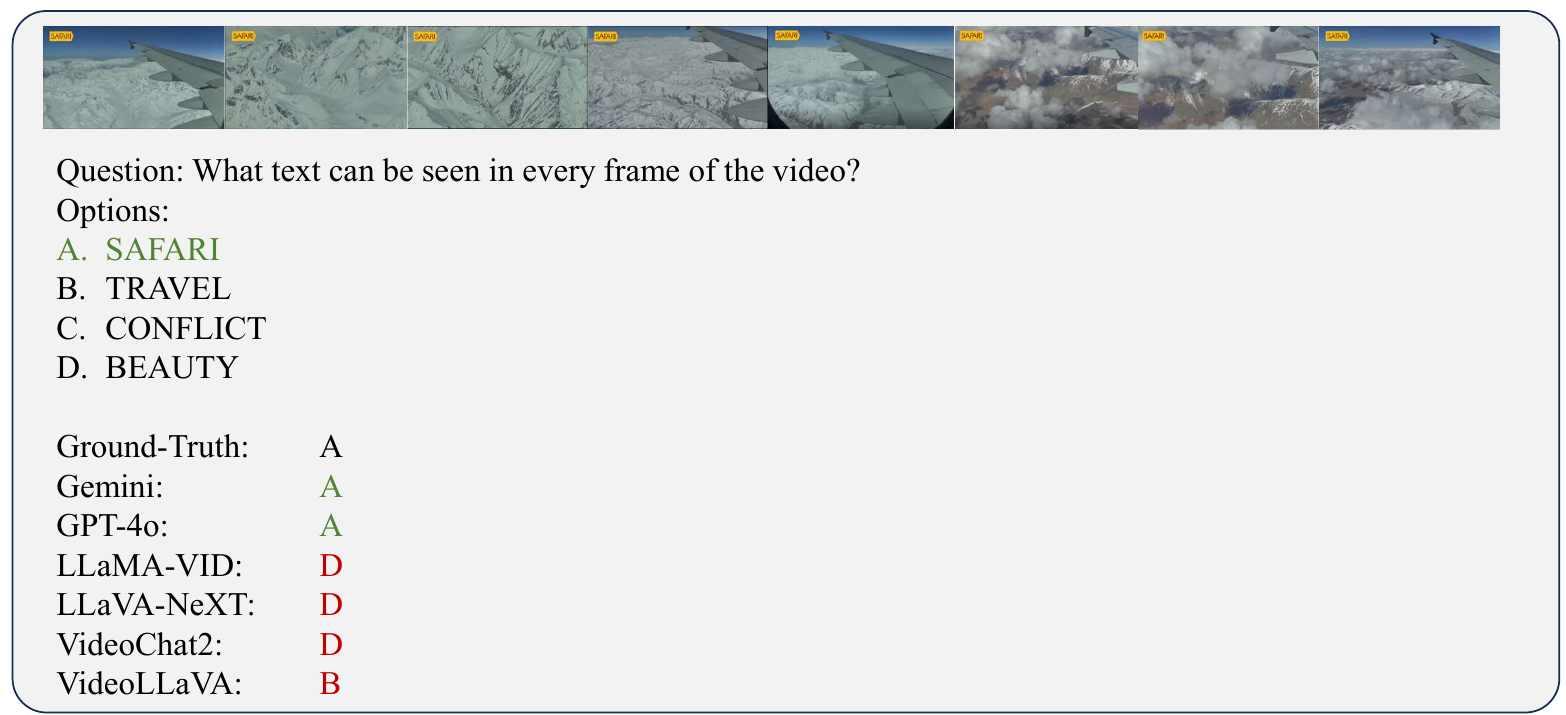}
    \caption{An Example of Video-LLMs responses and evaluation results of the OCR. We use \textcolor{forestgreen}{Green} to indicate correct and \textcolor{red}{Red} to indicate incorrect.}
    \label{fig:example_ocr}
\end{figure}

\begin{figure}
    \includegraphics[width=1\textwidth]{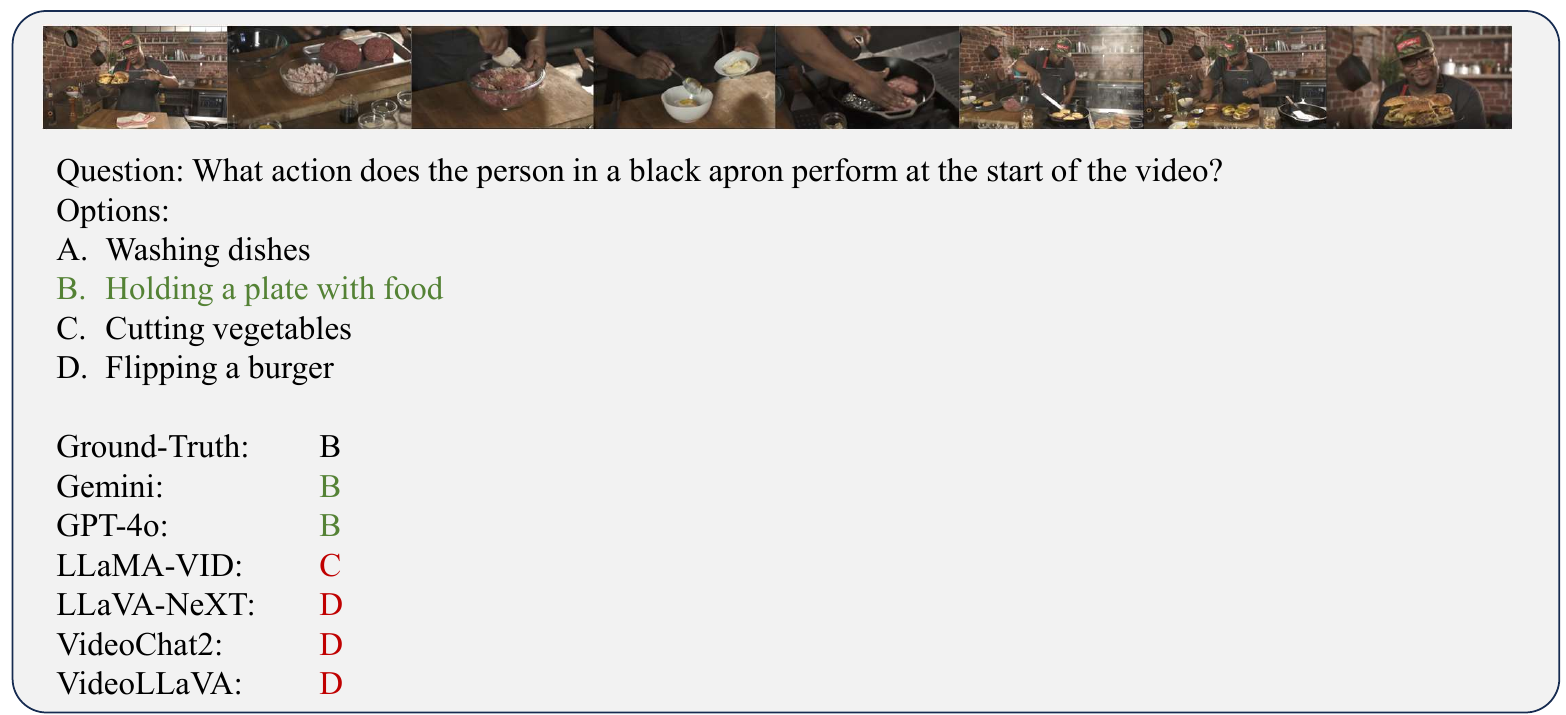}
    \caption{An Example of Video-LLMs responses and evaluation results of the Action Recognition. We use \textcolor{forestgreen}{Green} to indicate correct and \textcolor{red}{Red} to indicate incorrect.}
    \label{fig:example_ar}
\end{figure}

\begin{figure}
    \includegraphics[width=1\textwidth]{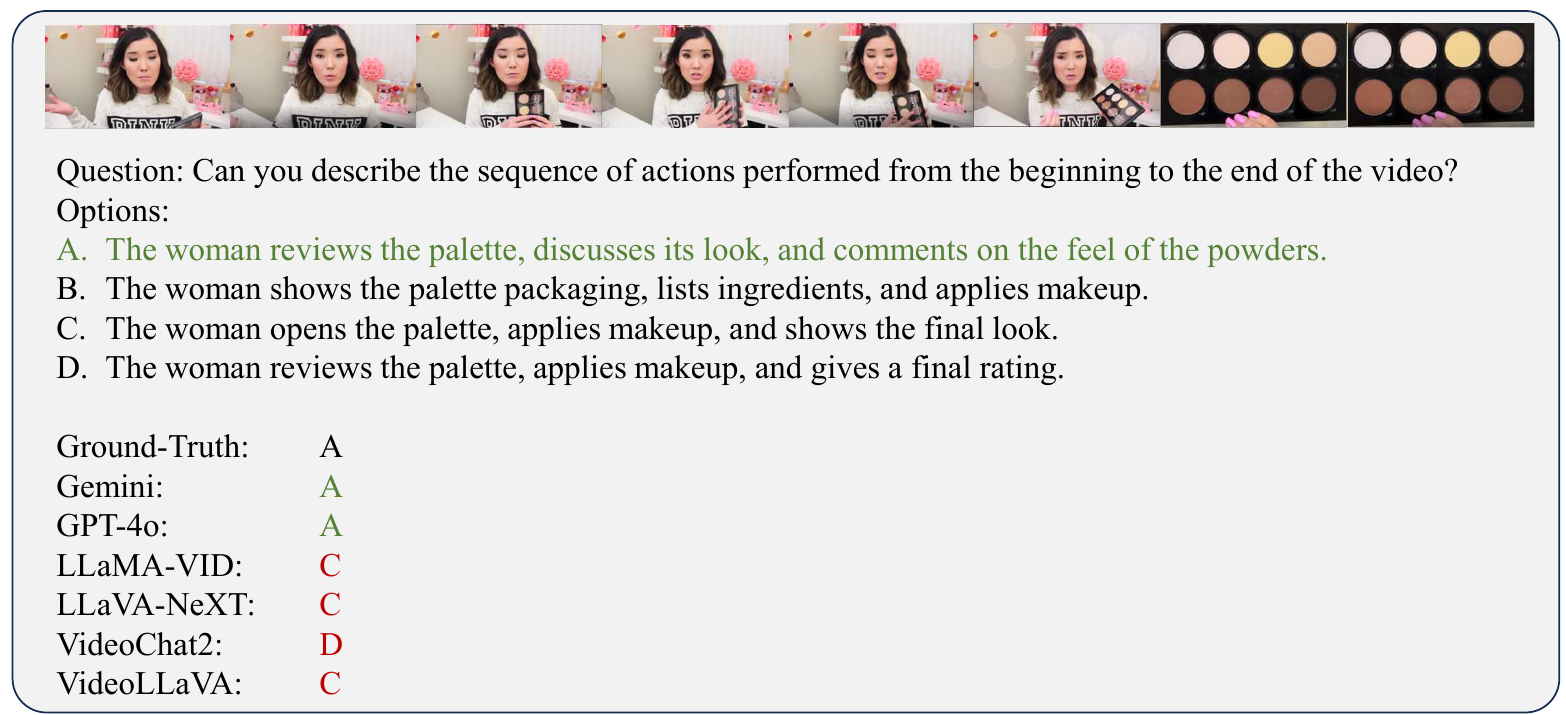}
    \caption{An Example of Video-LLMs responses and evaluation results of the Action Sequence. We use \textcolor{forestgreen}{Green} to indicate correct and \textcolor{red}{Red} to indicate incorrect.}
    \label{fig:example_as}
\end{figure}

\begin{figure}
    \includegraphics[width=1\textwidth]{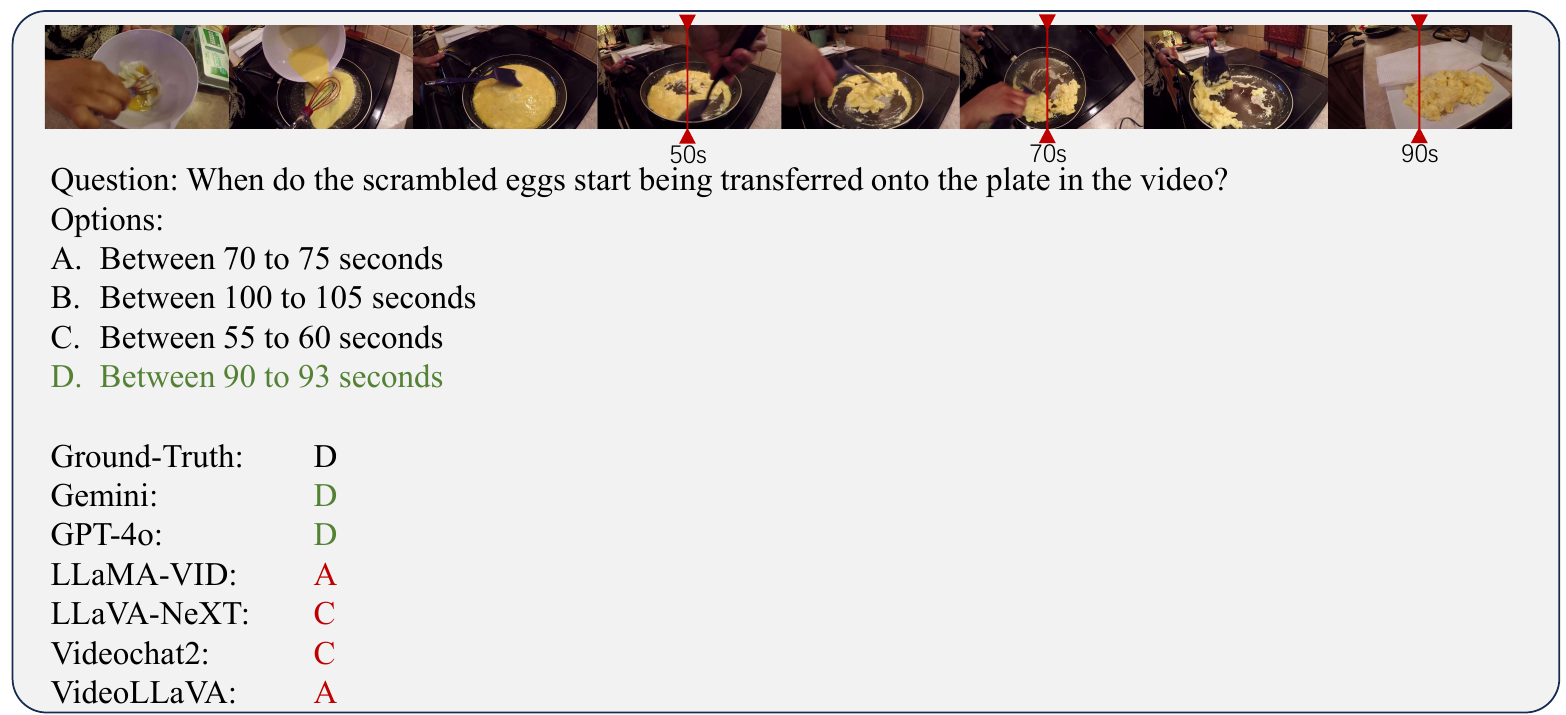}
    \caption{An Example of Video-LLMs responses and evaluation results of the Event Location. We use \textcolor{forestgreen}{Green} to indicate correct and \textcolor{red}{Red} to indicate incorrect.}
    \label{fig:example_el}
\end{figure}

\begin{figure}
    \includegraphics[width=1\textwidth]{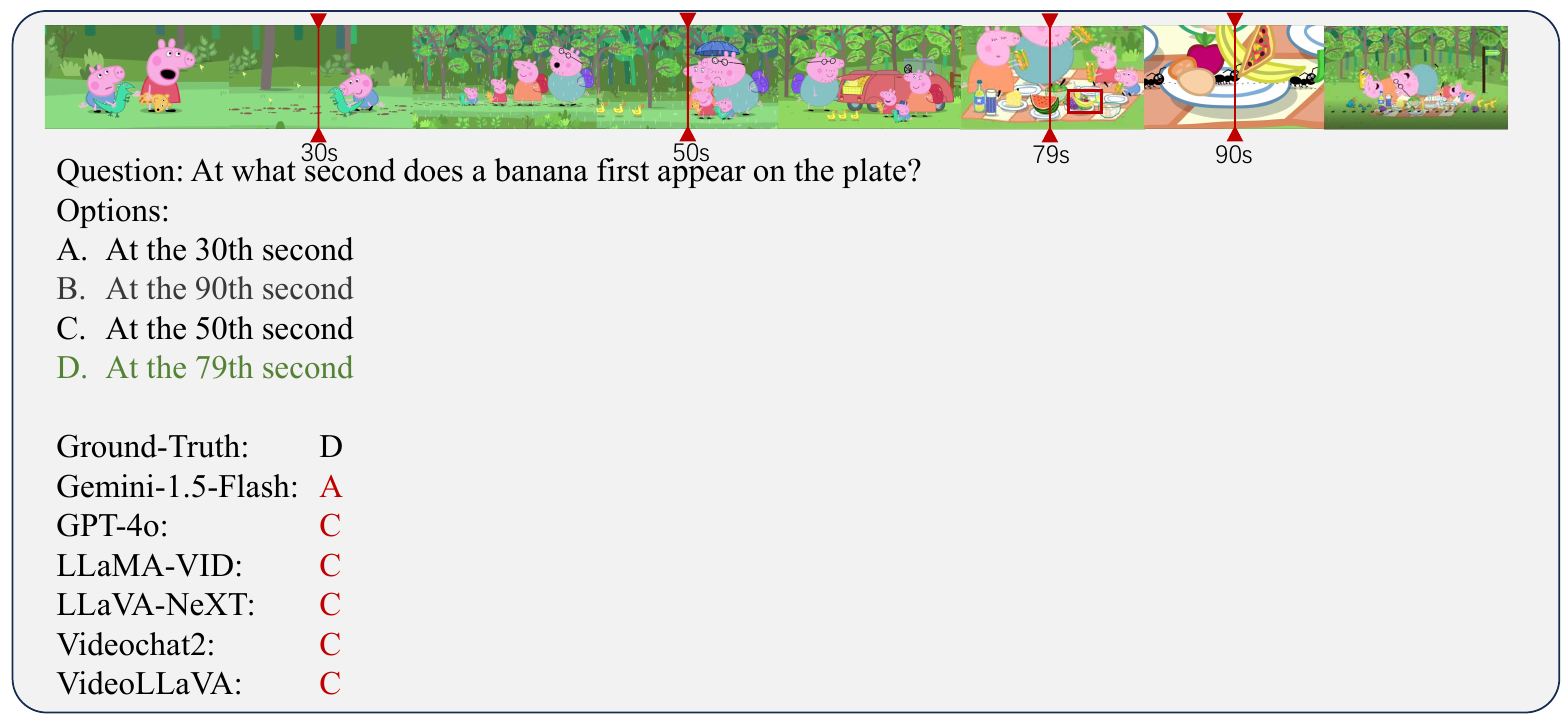}
    \caption{An Example of Video-LLMs responses and evaluation results of the Object Temporal Location. We use \textcolor{forestgreen}{Green} to indicate correct and \textcolor{red}{Red} to indicate incorrect.}
    \label{fig:example_otl}
\end{figure}

\begin{figure}
    \includegraphics[width=1\textwidth]{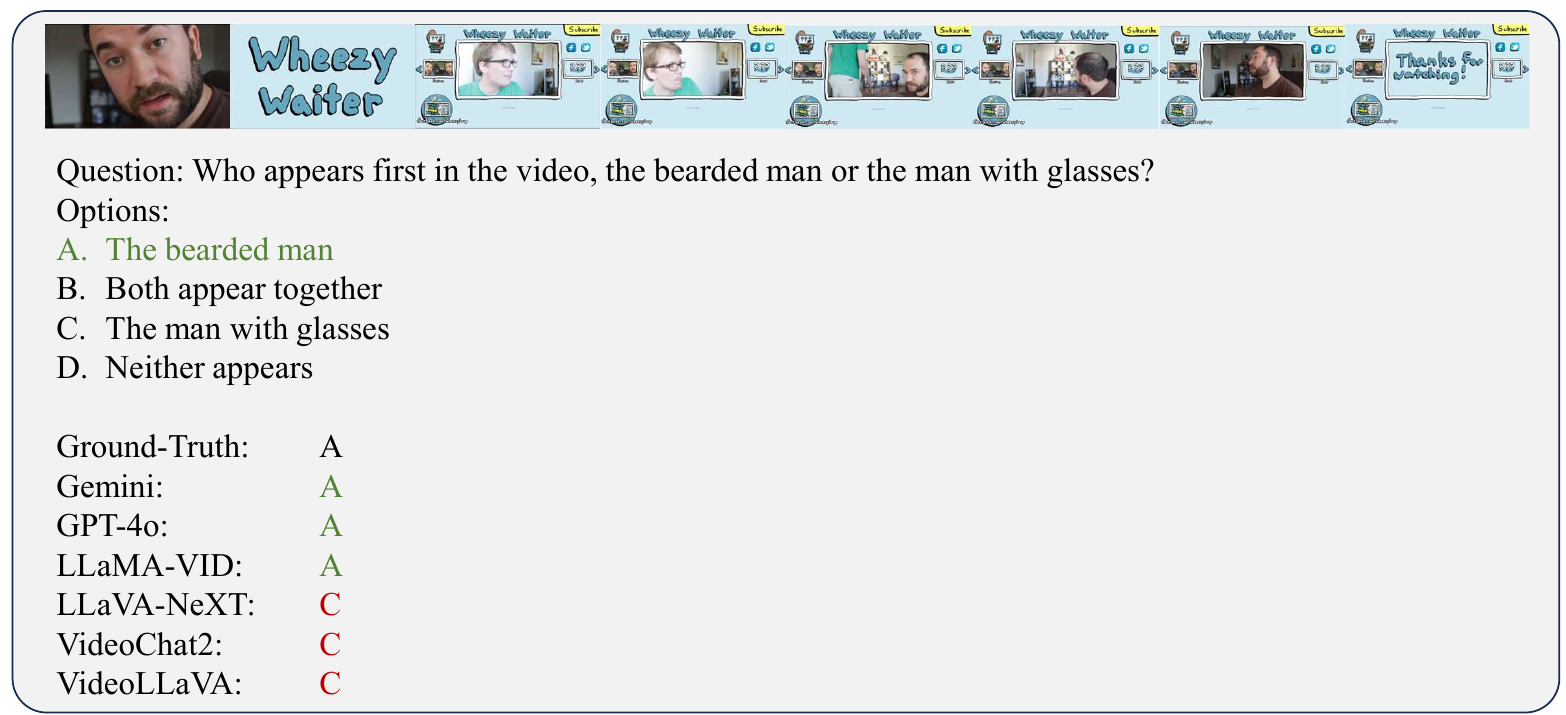}
    \caption{An Example of Video-LLMs responses and evaluation results of the Object Temporal Relation. We use \textcolor{forestgreen}{Green} to indicate correct and \textcolor{red}{Red} to indicate incorrect.}
    \label{fig:example_otr}
\end{figure}

\begin{figure}
    \includegraphics[width=1\textwidth]{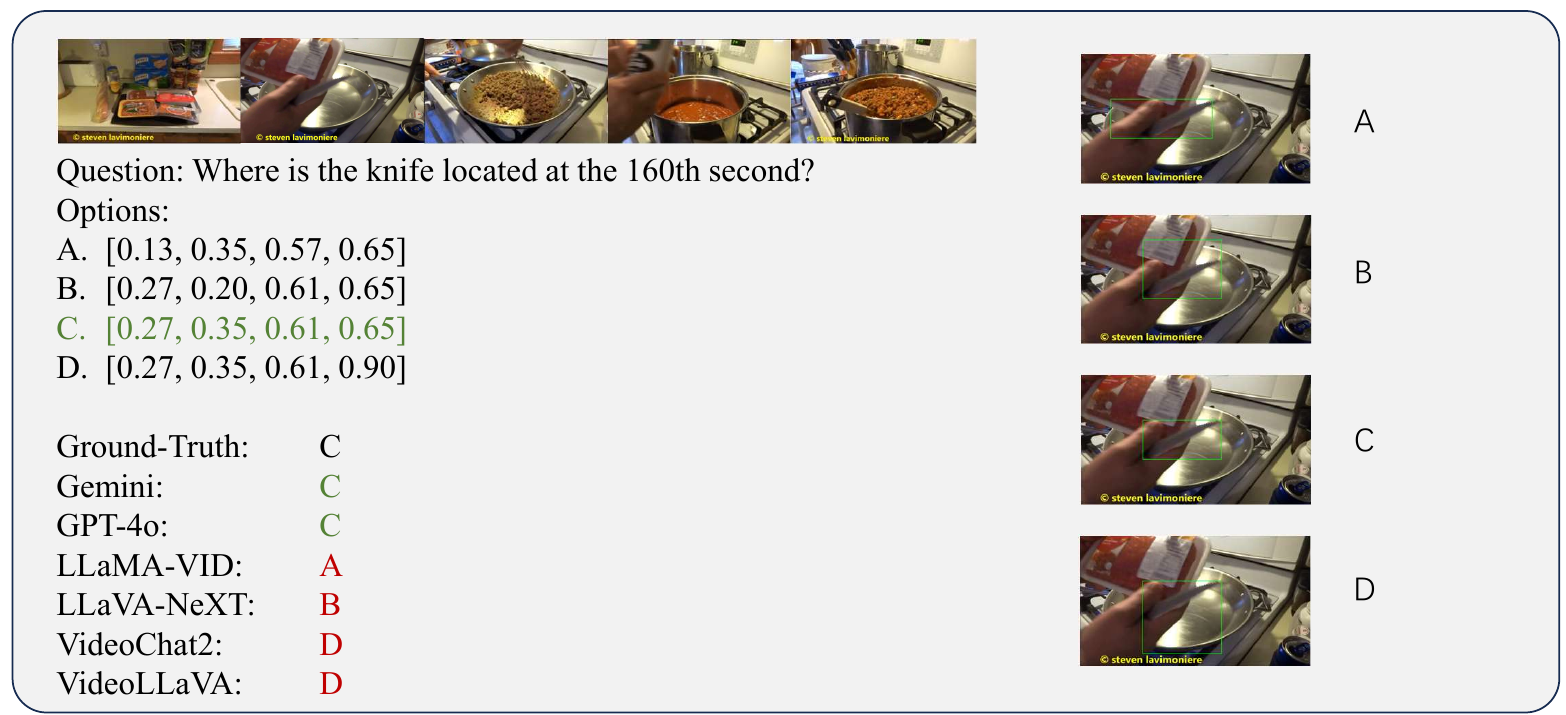}
    \caption{An Example of Video-LLMs responses and evaluation results of the Object Spatial Location. We use \textcolor{forestgreen}{Green} to indicate correct and \textcolor{red}{Red} to indicate incorrect.}
    \label{fig:example_osl}
\end{figure}

\begin{figure}
    \includegraphics[width=1\textwidth]{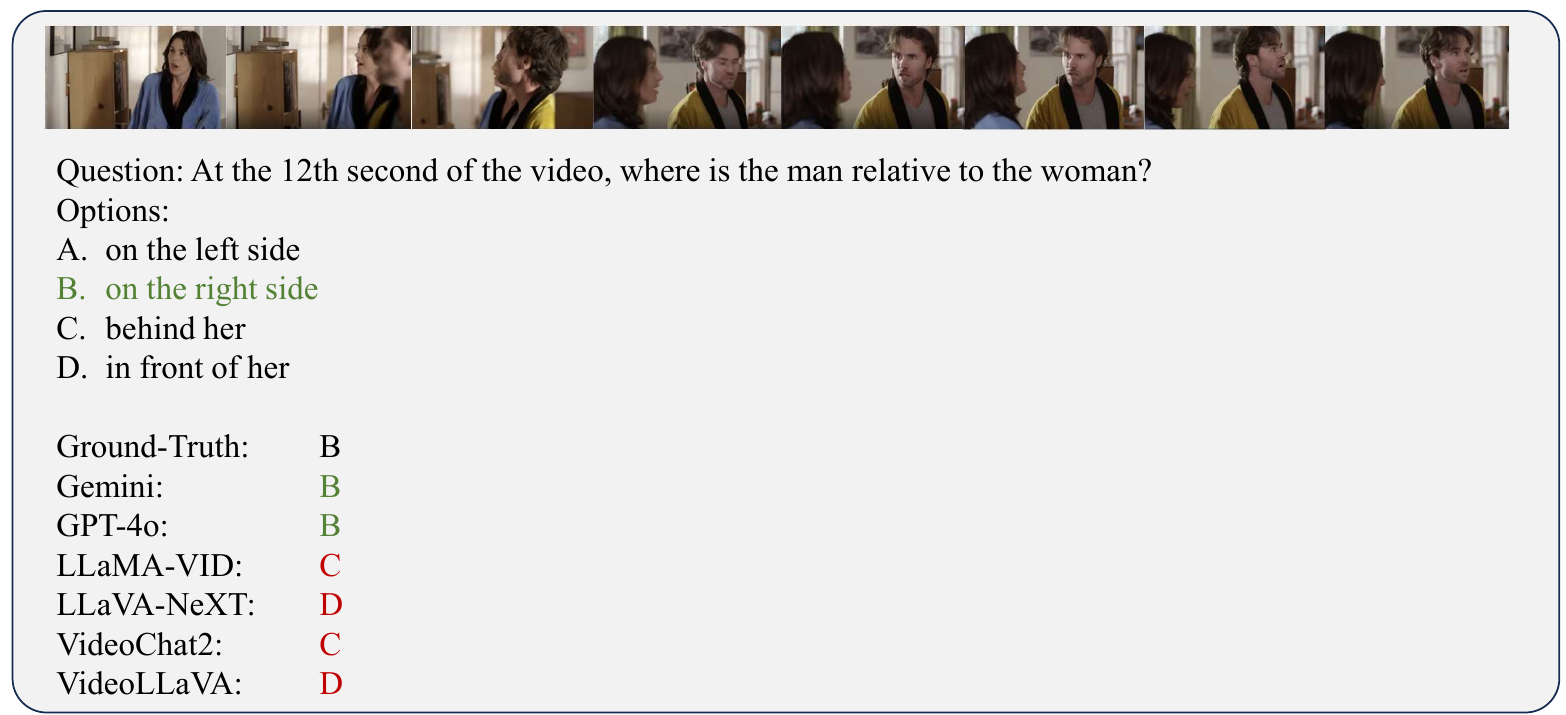}
    \caption{An Example of Video-LLMs responses and evaluation results of the Object Spatial Relation. We use \textcolor{forestgreen}{Green} to indicate correct and \textcolor{red}{Red} to indicate incorrect.}
    \label{fig:example_osr}
\end{figure}

\begin{figure}
    \includegraphics[width=1\textwidth]{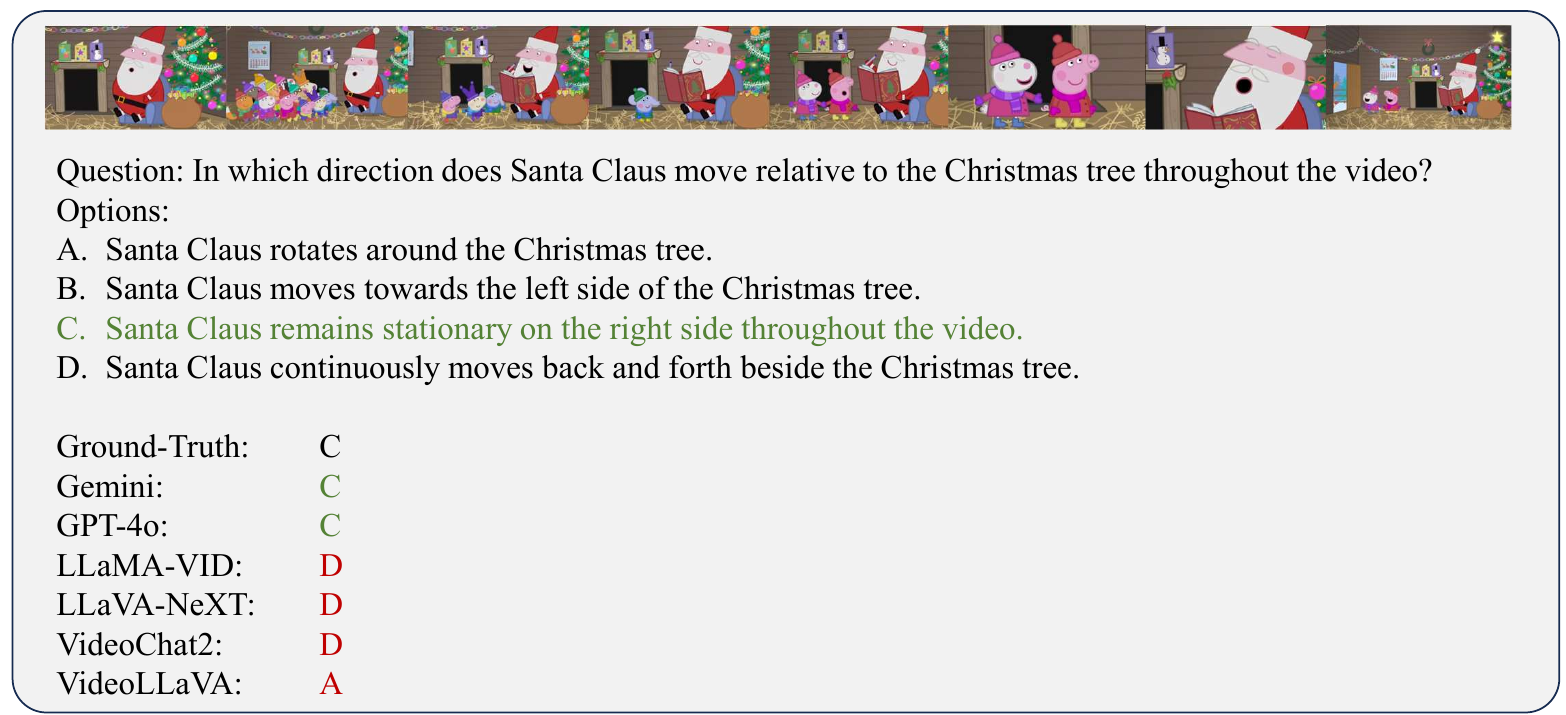}
    \caption{An Example of Video-LLMs responses and evaluation results of the Object Spatial Tracking. We use \textcolor{forestgreen}{Green} to indicate correct and \textcolor{red}{Red} to indicate incorrect.}
    \label{fig:example_ost}
\end{figure}

\begin{figure}
    \includegraphics[width=1\textwidth]{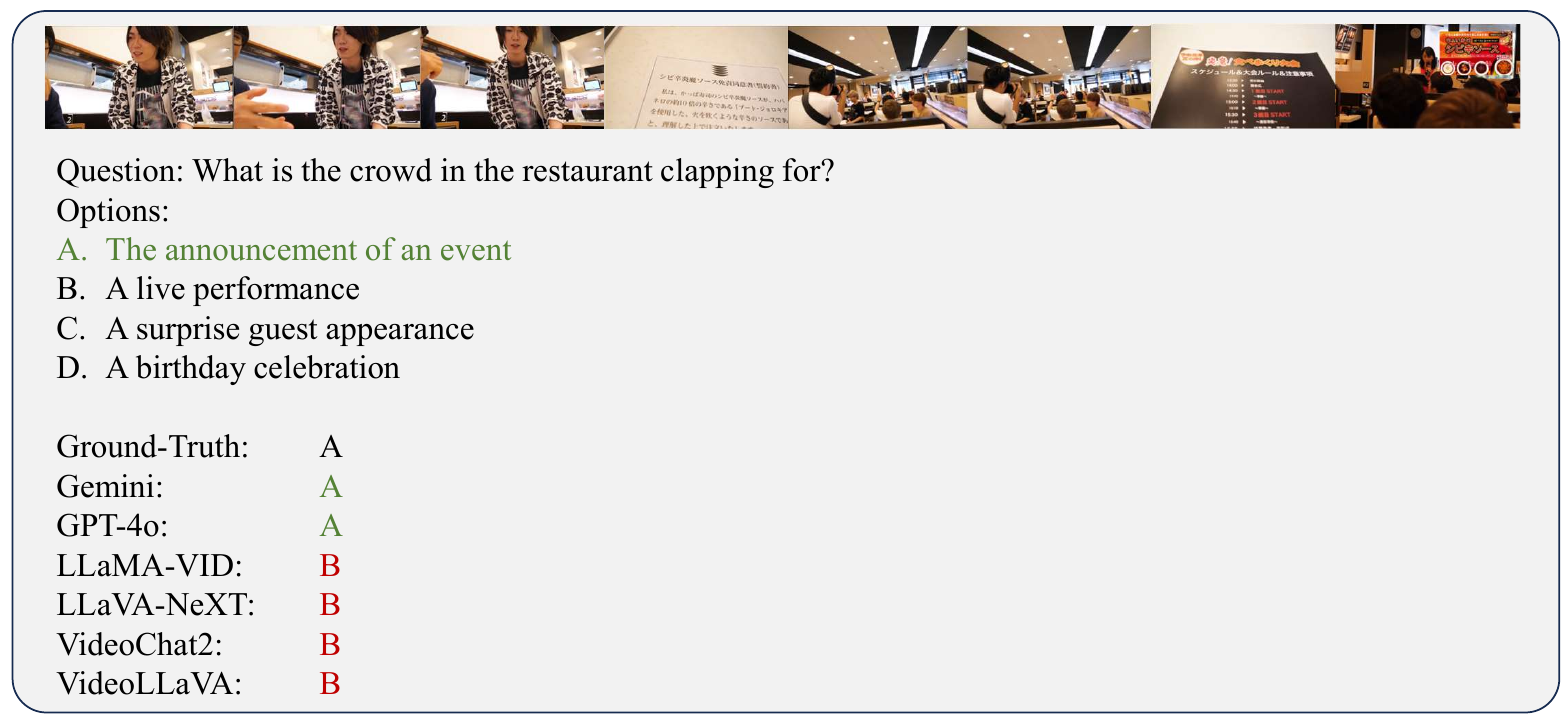}
    \caption{An Example of Video-LLMs responses and evaluation results of the Human Activity Analysis. We use \textcolor{forestgreen}{Green} to indicate correct and \textcolor{red}{Red} to indicate incorrect.}
    \label{fig:example_haa}
\end{figure}

\begin{figure}
    \includegraphics[width=1\textwidth]{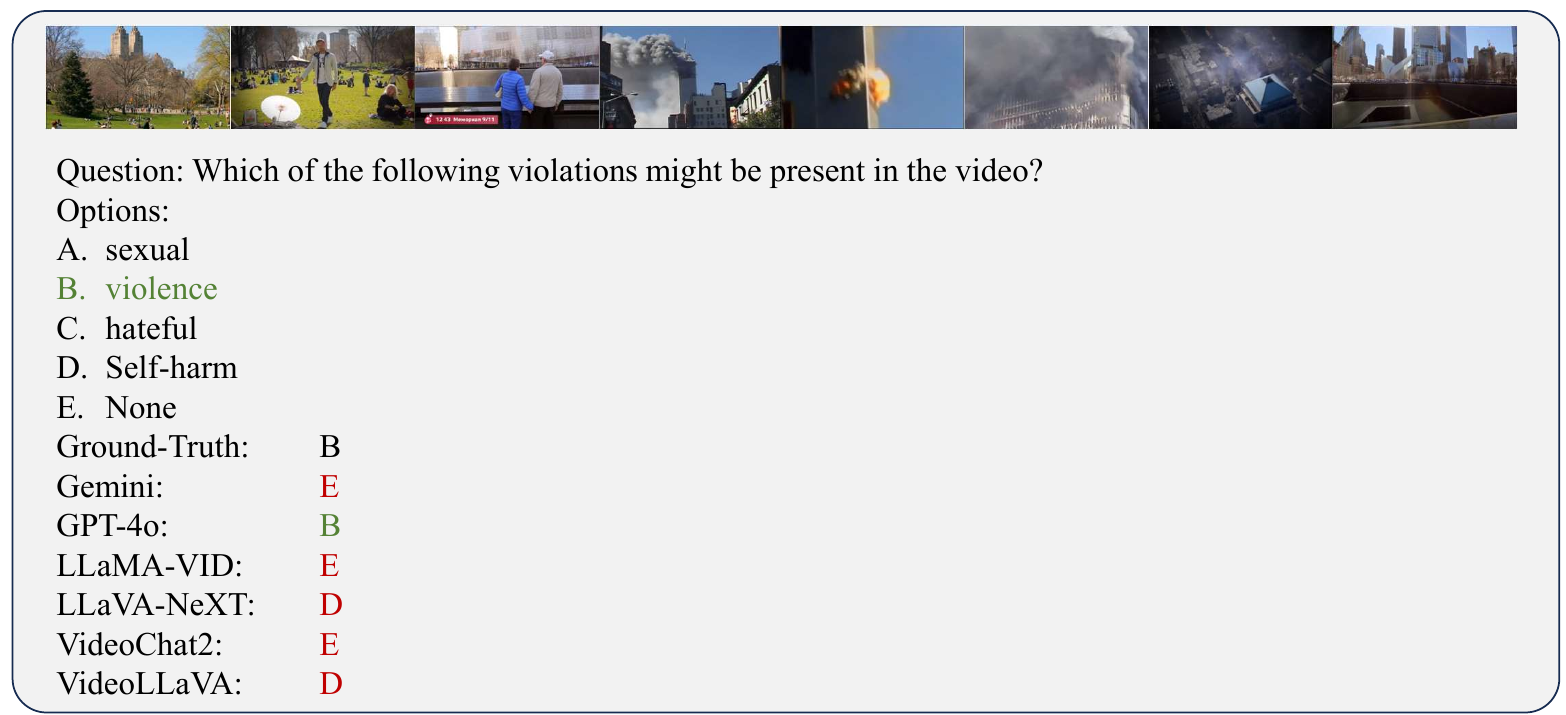}
    \caption{An Example of Video-LLMs Responses and evaluation results of the Anomaly Detection. We use \textcolor{forestgreen}{Green} to indicate correct and \textcolor{red}{Red} to indicate incorrect.}
    \label{fig:example_ad}
\end{figure}

\begin{figure}
    \includegraphics[width=1\textwidth]{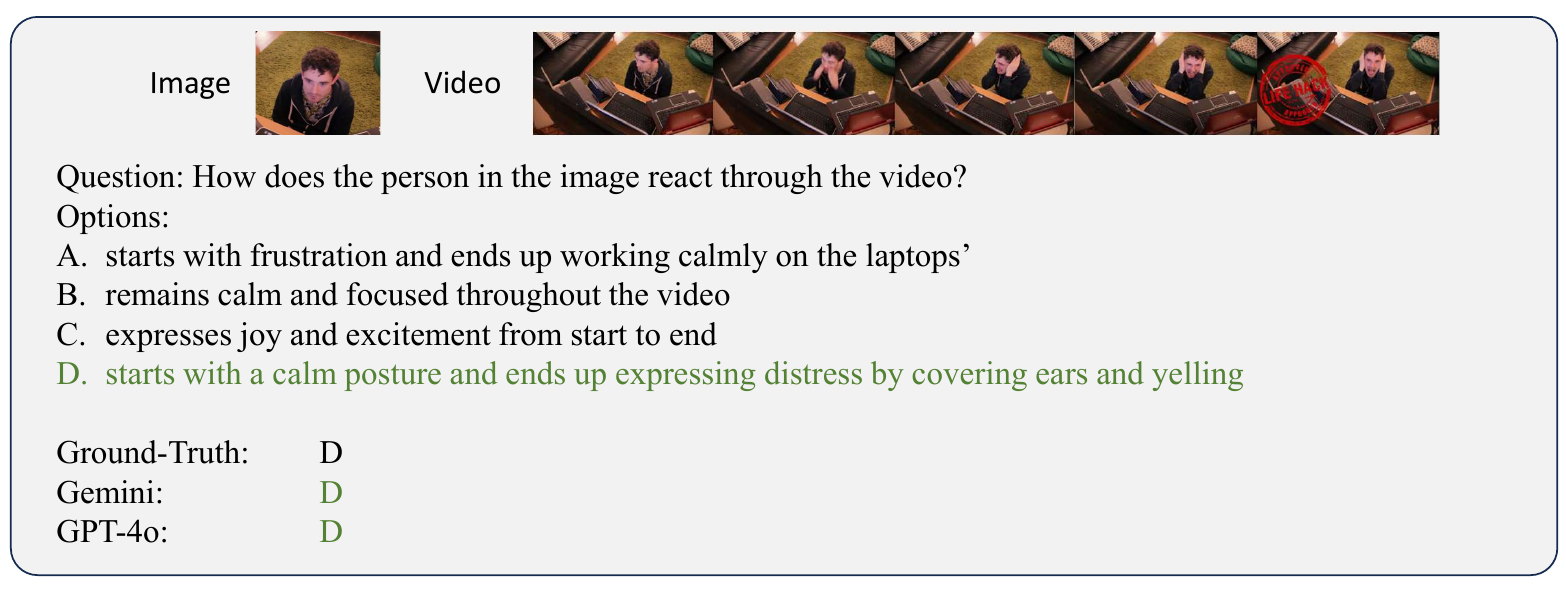}
    \caption{An Example of Video-LLMs responses and evaluation results of the Relation Reasoning(Image-Video). We use \textcolor{forestgreen}{Green} to indicate correct and \textcolor{red}{Red} to indicate incorrect.}
    \label{fig:example_rri}
\end{figure}

\begin{figure}
    \includegraphics[width=1\textwidth]{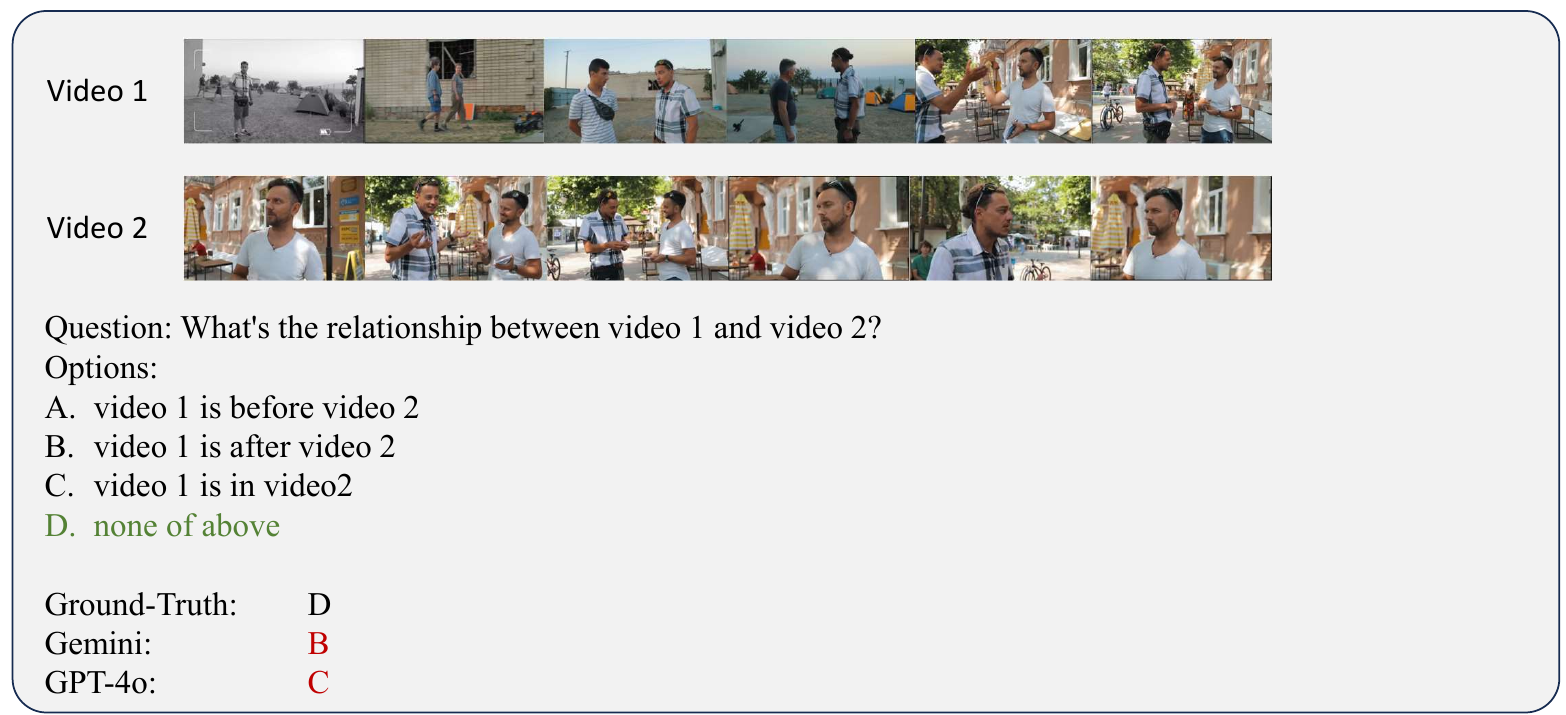}
    \caption{An Example of Video-LLMs responses and evaluation results of the Relation Reasoning(Video-Video). We use \textcolor{forestgreen}{Green} to indicate correct and \textcolor{red}{Red} to indicate incorrect.}
    \label{fig:example_rrv}
\end{figure}

\begin{figure}
    \includegraphics[width=1\textwidth]{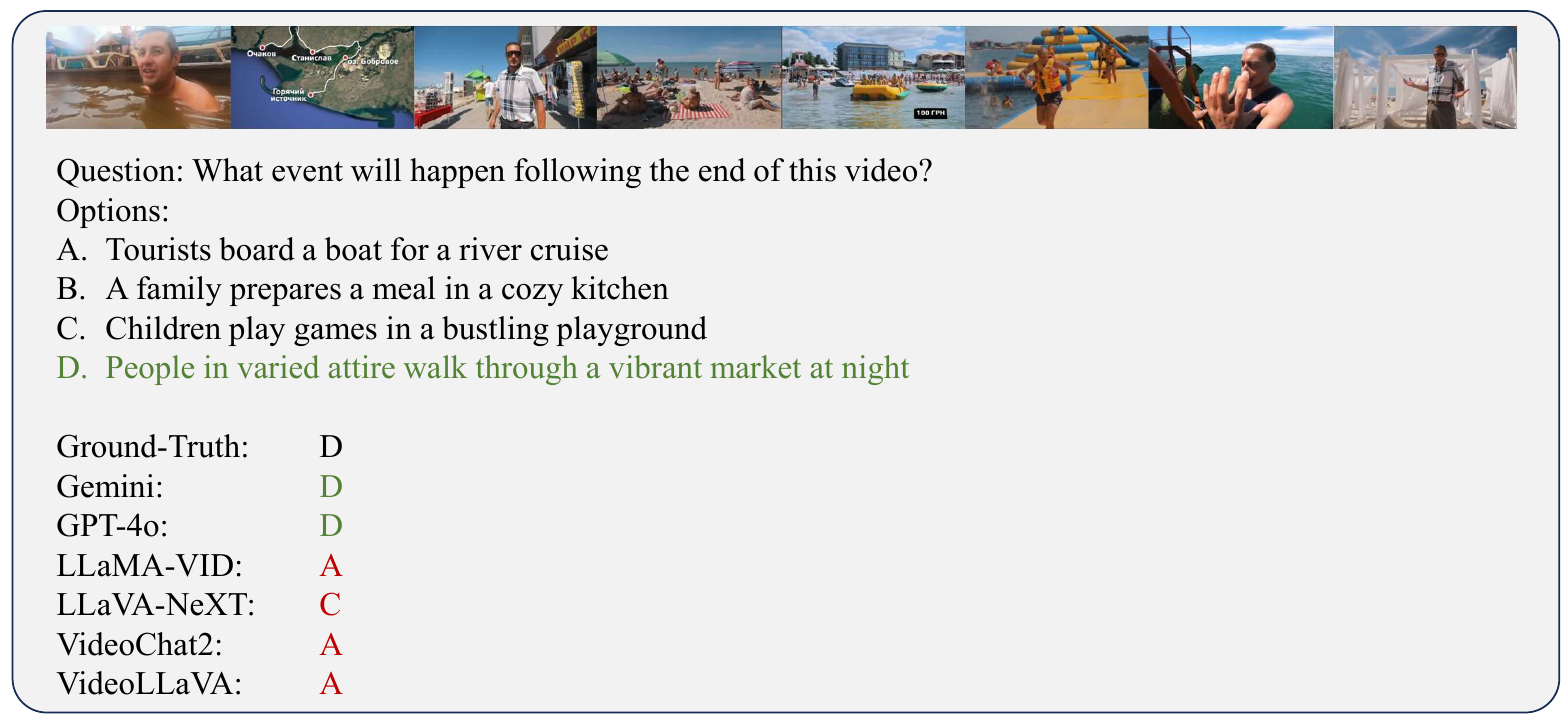}
    \caption{An Example of Video-LLMs responses and evaluation results of the Event Prediction. We use \textcolor{forestgreen}{Green} to indicate correct and \textcolor{red}{Red} to indicate incorrect.}
    \label{fig:example_ep}
\end{figure}

\begin{figure}
    \includegraphics[width=1\textwidth]{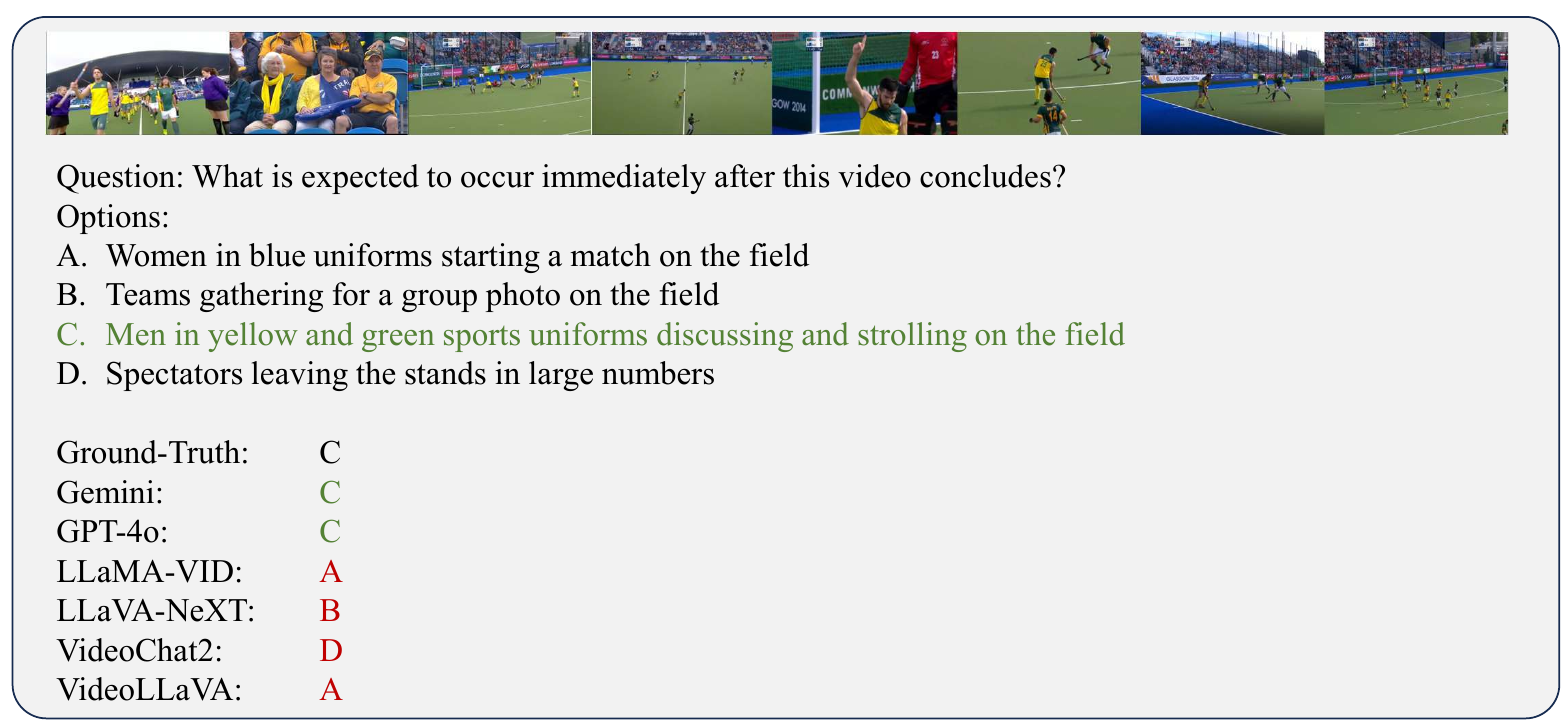}
    \caption{An Example of Video-LLMs responses and evaluation results of the Action Prediction. We use \textcolor{forestgreen}{Green} to indicate correct and \textcolor{red}{Red} to indicate incorrect.}
    \label{fig:example_ap}
\end{figure}

\begin{figure}
    \includegraphics[width=1\textwidth]{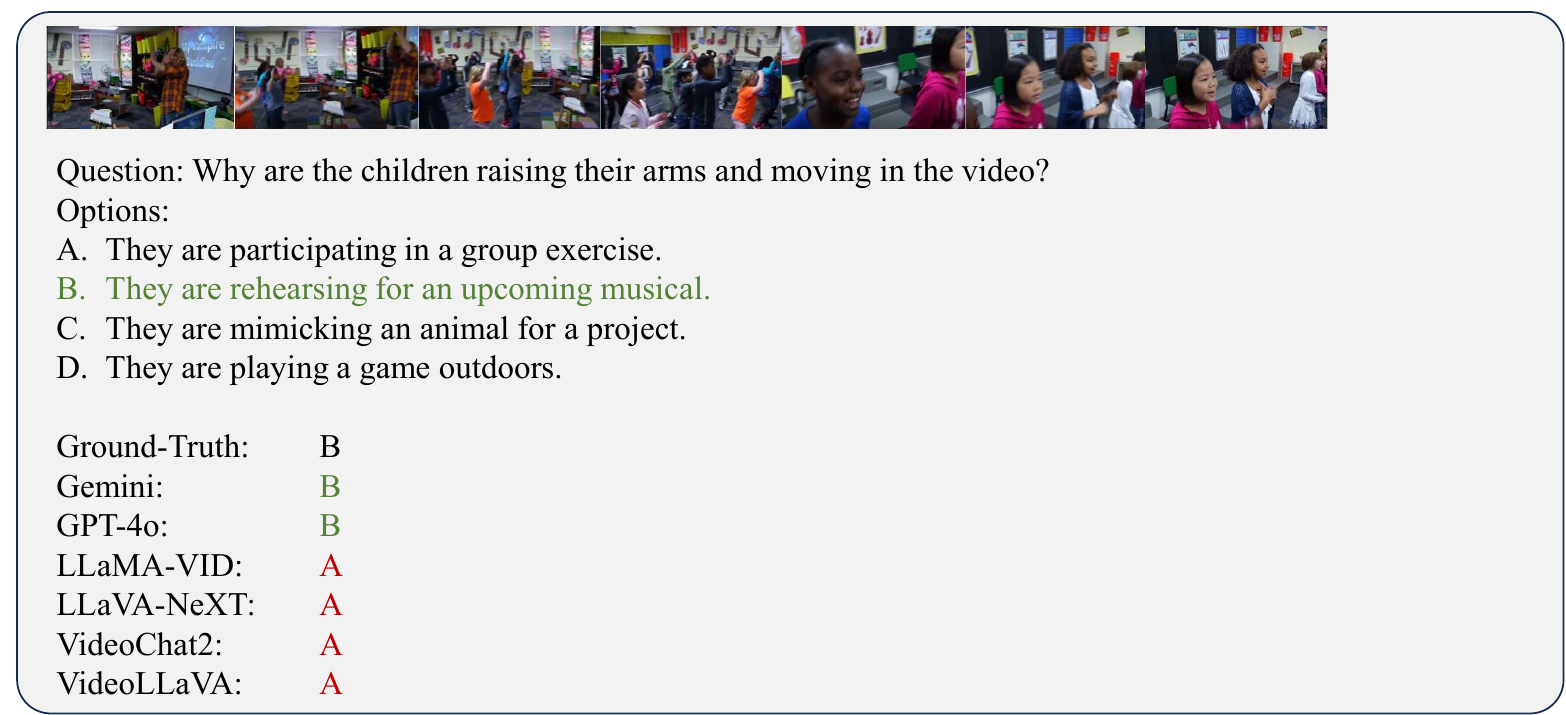}
    \caption{An Example of Video-LLMs responses and evaluation results of the Causal Reasoning. We use \textcolor{forestgreen}{Green} to indicate correct and \textcolor{red}{Red} to indicate incorrect.}
    \label{fig:example_car}
\end{figure}

\begin{figure}
    \includegraphics[width=1\textwidth]{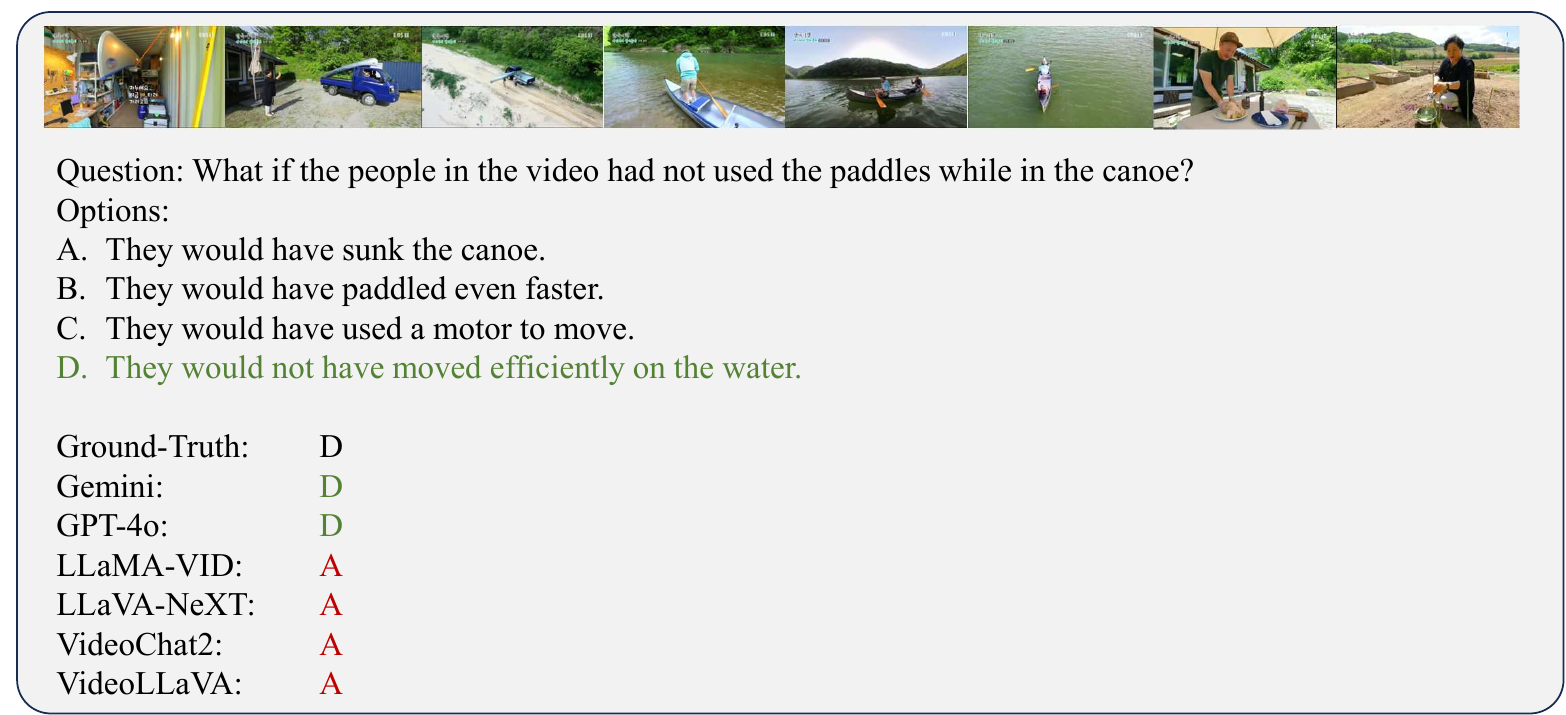}
    \caption{An Example of Video-LLMs responses and evaluation results of the  Counterfactual Reasoning. We use \textcolor{forestgreen}{Green} to indicate correct and \textcolor{red}{Red} to indicate incorrect.}
    \label{fig:example_cfr}
\end{figure}

\begin{figure}
\centering
    \includegraphics[width=0.85\textwidth]{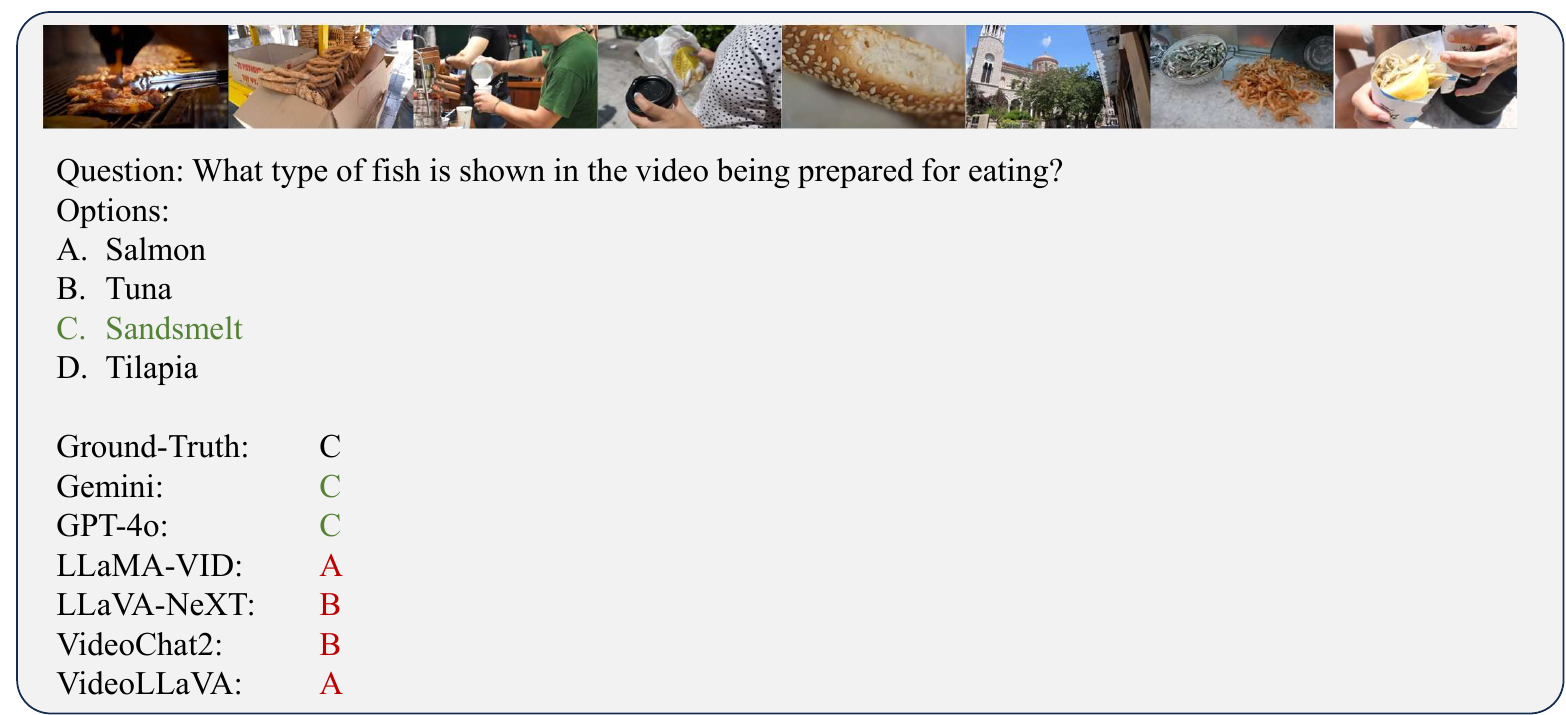}
    \caption{An Example of Video-LLMs responses and evaluation results of the  Commonsense Reasoning. We use \textcolor{forestgreen}{Green} to indicate correct and \textcolor{red}{Red} to indicate incorrect.}
    \label{fig:example_csr}
\end{figure}

\begin{figure}
\centering
    \includegraphics[width=0.85\textwidth]{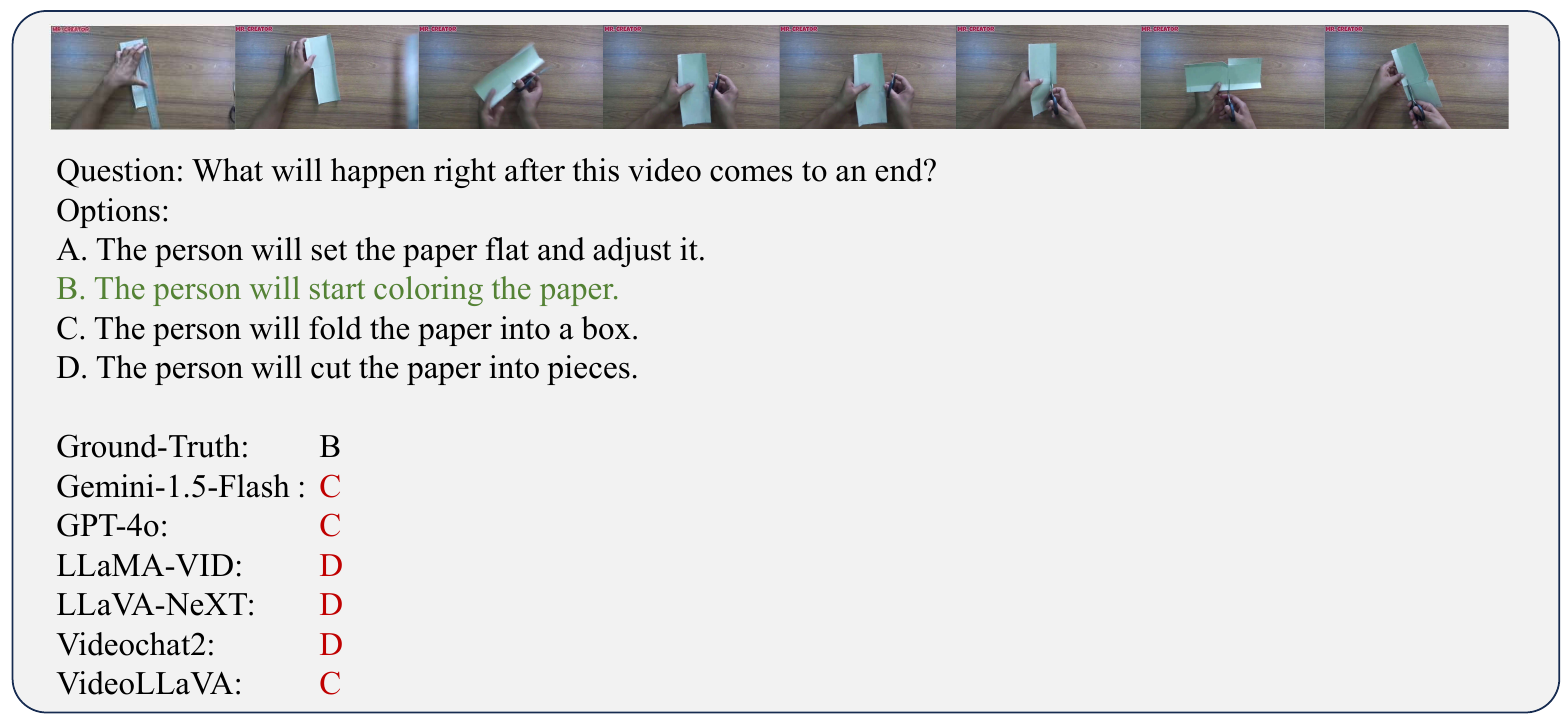}
    \caption{An Example of Video-LLMs responses and evaluation results of the  Logic Reasoning. We use \textcolor{forestgreen}{Green} to indicate correct and \textcolor{red}{Red} to indicate incorrect.}
    \label{fig:lr}
\end{figure}

\begin{figure}
\centering
    \includegraphics[width=0.85\textwidth]{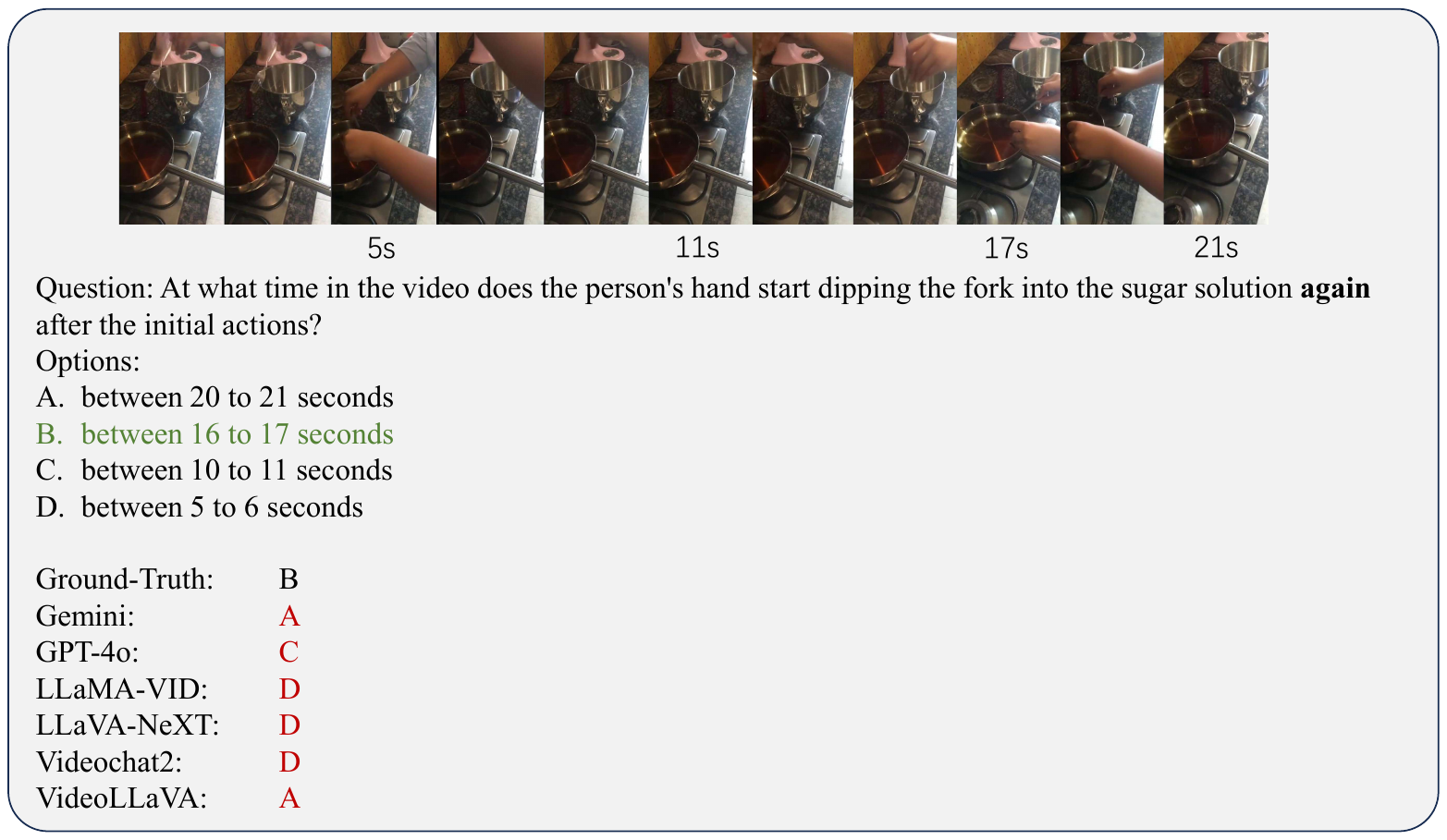}
    \caption{An Example of Video-LLMs responses and evaluation results of the  Action Location. We use \textcolor{forestgreen}{Green} to indicate correct and \textcolor{red}{Red} to indicate incorrect.}
    \label{fig:example_al}
\end{figure}

In Figures~\ref{fig:example_oe}- \ref{fig:example_al}, we provide a specific case for each proposed task type.
The case shows sampled frames from the video, along with the corresponding questions, options and ground truth. Besides, we included the responses to the questions from two commercial models, GPT-4o and Gemini, as well as four high-performing open-source Video-LLMs: LLaMA-VID, LLaVA-NeXT, Videochat2, and VideoLLaVA. From these cases, we can also observe the significant gap between open-source Video-LLMs and commercial models.

\end{document}